%% file: acl_latex.tex
\documentclass[11pt]{article}

\usepackage[preprint]{acl}

\usepackage{times}
\usepackage{latexsym}
\usepackage{enumitem}

\usepackage[T1]{fontenc}

\usepackage[utf8]{inputenc}

\usepackage{microtype}

\usepackage{inconsolata}

\usepackage{graphicx}

\usepackage{booktabs}
\usepackage{multirow}
\usepackage{algorithm}
\usepackage{algorithmic}
\usepackage{xspace}

\usepackage{amsmath}
\usepackage{amssymb}
\usepackage{booktabs,multirow,adjustbox,diagbox,threeparttable,tabularray}
\usepackage{subcaption}
\usepackage{tabularx}
\usepackage{array}
\usepackage{makecell}
\usepackage{longtable} 
\usepackage{paracol}
\usepackage{fontawesome5}
\usepackage[tableposition=top]{caption}
\usepackage{colortbl}
\usepackage[table]{xcolor}

\usepackage{xcolor}
\usepackage{tcolorbox}
\tcbuselibrary{breakable,skins}

\definecolor{insightbg}{RGB}{239,242,248}
\newcommand{\insighticon}{\textcolor{yellow!85!orange}{\faLightbulb}}

\usepackage{microtype}

\usepackage{appendix} %
\usepackage{pifont}
\usepackage{listings}

\title{GroundAct: Can LLM Agents Ground Actions in Environmental States?}
\newcommand{\framework}{GroundAct\xspace}
\newcommand{\simulator}{GA-Sim\xspace}
\newcommand{\benchmark}{GA-Bench\xspace}

\author{
  \textbf{Zixuan Wang}\textsuperscript{1*},
  \textbf{Dingming Li}\textsuperscript{1*},
  \textbf{Hongxing Li}\textsuperscript{1},
  \textbf{Yanrui Miao}\textsuperscript{1},
  \textbf{Shuo Chen}\textsuperscript{1},
  \textbf{Yuchen Yan}\textsuperscript{1},
\\
  \textbf{Wenqi Zhang}\textsuperscript{1},
  \textbf{Yongliang Shen}\textsuperscript{1\dag},
  \textbf{Weiming Lu}\textsuperscript{1},
  \textbf{Jun Xiao}\textsuperscript{1},
  \textbf{Yueting Zhuang}\textsuperscript{1}
\\
  \textsuperscript{1}Zhejiang University
\\
  \texttt{\{wang.zixuan, syl\}@zju.edu.cn}
\\
  \faGithub~\href{https://github.com/ZJU-REAL/OmniEmbodied}{github.com/ZJU-REAL/OmniEmbodied}
\\
  \faGlobe~\href{https://zju-real.github.io/OmniEmbodied}{zju-real.github.io/OmniEmbodied}
}

\begin{document}
\maketitle

\renewcommand{\thefootnote}{\fnsymbol{footnote}}
\footnotetext[1]{Equal contribution.}
\footnotetext[2]{Corresponding author.}
\renewcommand{\thefootnote}{\arabic{footnote}}

\begin{figure*}[t]
    \centering
    \includegraphics[width=1\linewidth]{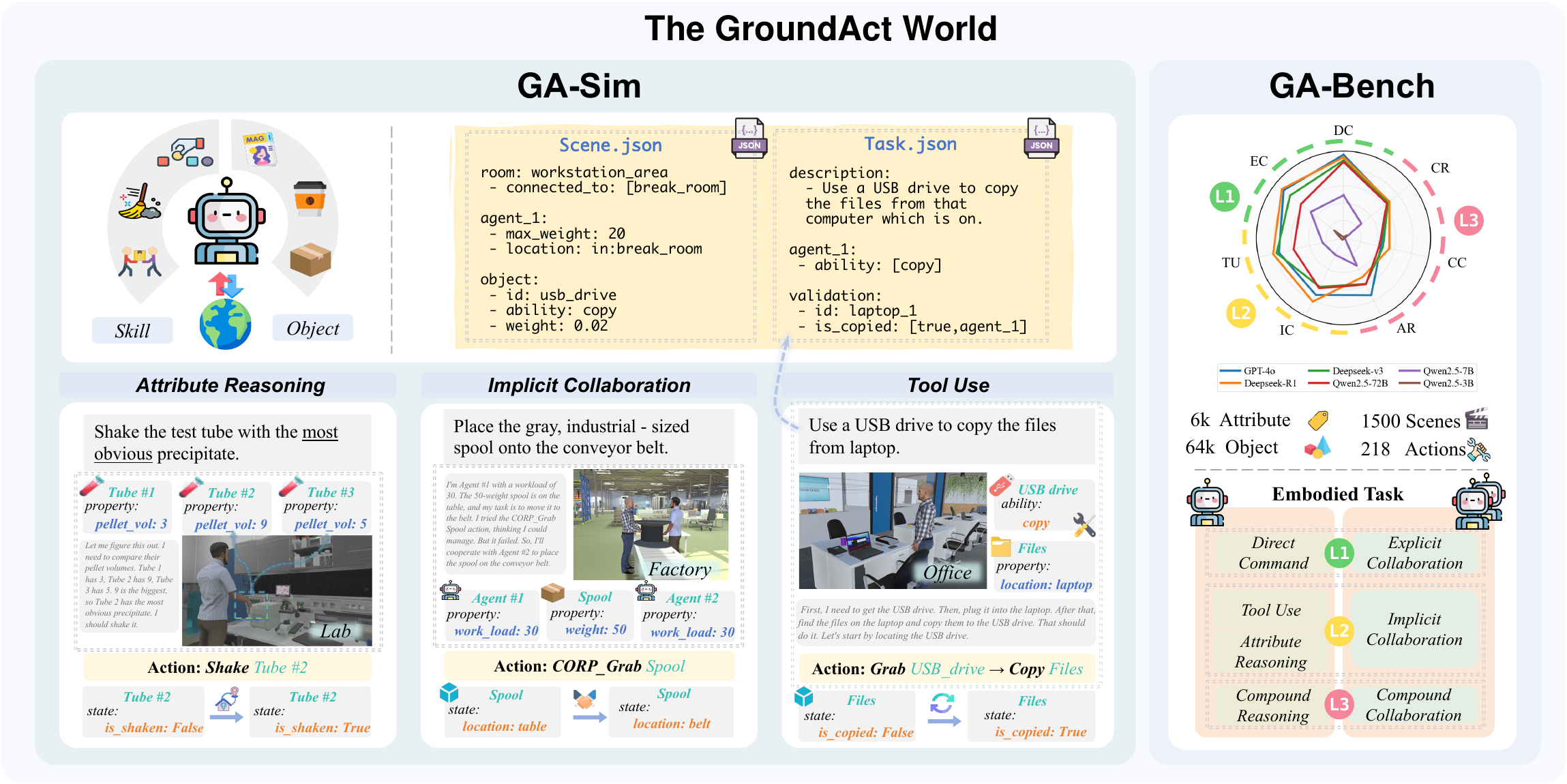}   \caption{Overview of the \framework framework comprising three integrated components: \textbf{\simulator} (left) uses structured text representation to model environments with objects, agents, and spatial relationships, enabling dynamic tool-capability binding and physics-constrained collaboration; \textbf{\benchmark} (right) presents our comprehensive evaluation matrix spanning single-agent and multi-agent tasks across increasing cognitive complexity levels.}
    \label{fig:main-overview}
    \vspace{-1em}
\end{figure*}

\begin{abstract}
    \input{sections/0_abstract}
\end{abstract}

\input{sections/1_introduction}
\input{sections/2_related_works}
\input{sections/3_method}

\input{sections/4_experiment}
\input{sections/5_conclusion}

\section*{Limitations}

Our framework employs text-based environment representation rather than continuous 3D simulation. While this design enables efficient large-scale evaluation and isolates reasoning capabilities from perceptual challenges, it abstracts away aspects of physical embodiment including continuous control, sensorimotor feedback, and real-time constraints. The performance gaps we identify may manifest differently in systems that must process raw sensory inputs and execute continuous motor commands, and our findings require validation in physically grounded robotic systems.

Our evaluation covers 15 representative models across three architectural paradigms using a fixed prompt format. The rapid pace of model development means newer architectures may address some identified limitations, and performance may vary with alternative prompting strategies not explored in this work. Additionally, although \benchmark spans 11 domains, coverage of household and outdoor scenarios is limited compared to professional environments. All task instructions are in English, and the reasoning patterns we identify may not generalize to multilingual settings.

\section*{Ethics Statements}

We adhere to the ACL Code of Ethics and Code of Conduct. Human annotators involved in scenario verification provided informed consent, were compensated above local minimum wage, and their privacy was protected throughout the process. GroundAct benchmark and code will be released under the MIT License. We used Claude for grammatical refinement; all scientific content and conclusions are the authors' own work.

\bibliography{custom}
\appendix
\input{sections/6_appendix}

\end{document}

%% file: sections/0_abstract.tex
LLM agents achieve 85--96\% success on tasks where instructions fully specify the action, but drop to 29--53\% when action feasibility depends on environmental state that the instruction does not mention. We argue that this gap reflects a missing capability: action grounding, the ability to infer from structured environmental state whether an action is feasible, what prerequisites it lacks, and whether it exceeds individual capacity. We introduce GroundAct, a benchmark of 1,500 scenarios and 16,592 task instances in text-based interactive environments spanning 11 domains, with tasks organized into seven categories along a cognitive complexity hierarchy. Evaluating 15 LLMs (3B--671B), we find three diagnostic patterns: (i) attribute reasoning is weakly correlated with tool and coordination reasoning, producing distinct model profiles; (ii) complete environment graphs yield up to $+27.6$/$-22.9$\% on tool use vs.\ implicit collaboration, separating search-bound from constraint-filtering bottlenecks; and (iii) supervised fine-tuning lifts Qwen2.5-3B from 0.6\% to 76.3\% on direct command but only 1.5\% to 5.5\% on implicit collaboration. These results establish action grounding as a multi-dimensional challenge irreducible to scaling.

%% file: sections/1_introduction.tex
\section{Introduction}

Consider an agent asked to ``clear the workbench.'' Three items remain: a coffee mug the agent can lift, a 50\,kg cable spool that exceeds the agent's load limit, and a coffee stain that requires a wiping tool. None of these constraints is named in the instruction; they live in the environment. Stronger models recognize each, recruit a partner for the spool, fetch a cloth for the stain, and complete the task. Weaker models attempt the spool alone, repeat the failed grab after an explicit error, and declare the task done. We call the ability to infer such constraints \emph{action grounding}.

We decompose action grounding into three operationalized dimensions. \emph{Attribute reasoning} grounds action targets to continuous object properties, requiring comparison across candidates rather than predicate lookup, such as identifying the heaviest of several boxes or the only object above a temperature threshold. \emph{Tool reasoning} grounds action prerequisites to dynamic capability availability: removing the stain requires grasping the cloth, which extends the agent's action set to include \texttt{clean}; releasing the cloth retracts this capability. \emph{Coordination reasoning} grounds action feasibility to individual capacity limits: when the spool's weight exceeds any single agent's load limit, the agent must recognize the mismatch and initiate collaboration without explicit direction. The structural property shared by all three: the constraint that determines what action to take, or whether the agent can take it alone, lives in the environment, not the instruction.

\begin{tcolorbox}[enhanced,colback=insightbg,colframe=gray!55,boxrule=1.5pt,arc=6pt,left=7pt,right=7pt,top=3pt,bottom=3pt,fontupper=\itshape]
\insighticon~Do benchmarks evaluate whether agents can ground actions in environmental states?
\end{tcolorbox}

Existing interactive-agent benchmarks implicitly perform action grounding on the agent's behalf. In ALFRED~\citep{shridhar2020alfredbenchmarkinterpretinggrounded} and ALFWorld~\citep{shridhar2021alfworldaligningtextembodied}, instructions name target objects by identifier (``put mug\_1 on desk\_1''), so attribute reasoning is pre-solved; BEHAVIOR-1K~\citep{li2024behavior} extends attribute coverage but retains discrete predicates (open/closed, clean/dirty) as primary state variables. In tool-use evaluations such as PARTNR~\citep{chang2024partnrbenchmarkplanningreasoning} and EmbodiedBench~\citep{yang2025embodiedbench}, tool-corresponding actions are placed in the action set at initialization, so tool reasoning is pre-loaded. In multi-agent benchmarks such as PARTNR, TDW-MAT, and C-WAH~\citep{zhang2024buildingcooperativeembodiedagents}, collaboration is triggered by explicit directives (``Agent A and Agent B cooperate to\ldots'') or by efficiency optimization, so coordination reasoning is pre-resolved. Across these settings, the action, tool, or collaboration pattern is supplied rather than inferred; existing benchmarks evaluate action execution under pre-grounded constraints, not action grounding from environmental state. Table~\ref{tab:dataset_comparison} summarizes the contrast.

We introduce \textbf{GroundAct}, a benchmark designed so each task category requires a form of action grounding that prior work bypasses. Three design choices instantiate the three reasoning dimensions: 6{,}381 continuous attribute types as first-class primitives, so attribute reasoning requires comparison rather than predicate lookup; dynamic tool-capability binding over 214 action types, where grasping a tool extends the agent's action set at runtime and releasing it retracts the extension, so tool reasoning requires capability-gap detection and acquisition; and constraint-triggered collaboration encoded in task preconditions, where collaboration emerges from physical infeasibility rather than instruction directives, so coordination reasoning requires comparing object properties against agent parameters. GroundAct contains 1{,}500 validated scenarios and 16{,}592 task instances across 11 domains, organized into seven categories along two axes: cognitive complexity and agent configuration. GroundAct uses text-based structured environments by design, isolating action grounding from perceptual challenges so every failure can be attributed to reasoning rather than perception.

Evaluating 15 representative LLMs (3B--671B parameters), we find three diagnostic patterns. First, attribute reasoning is weakly correlated with tool and coordination reasoning, producing distinct capability profiles across models: GPT-4o scores 77.8\% on attribute reasoning but 32.0\% on compound collaboration, while DeepSeek-R1 shows the opposite at 41.9\% and 48.5\%. Second, exposing the complete environment graph yields up to $+27.6$/$-22.9$\% on tool use vs.\ implicit collaboration, separating search-bound from constraint-filtering bottlenecks. Third, supervised fine-tuning lifts Qwen2.5-3B from 0.6\% to 76.3\% on direct command but only from 1.5\% to 5.5\% on implicit collaboration, exposing a graded learnability profile across dimensions.

Our contributions are as follows:
\begin{itemize}[nosep, leftmargin=*]
    \item We define \emph{action grounding} as inferring action-relevant constraints from environmental state, operationalized as attribute, tool, and coordination reasoning.
    \item We introduce \textbf{GroundAct}, a benchmark of 1{,}500 scenarios and 16{,}592 task instances across 11 domains in seven categories, each requiring a form of grounding bypassed by prior work.
    \item Evaluating 15 LLMs, we show GroundAct yields dimension-level diagnostics: separating instruction execution from grounding, exposing capability profiles that aggregate scores hide, and revealing uneven effects of scaling, reasoning training, and fine-tuning across grounding dimensions.
\end{itemize}

%% file: sections/2_related_works.tex
\section{Related Works}

Prior embodied benchmarks have made significant contributions to task evaluation, but differ fundamentally in their approach to physical reasoning and collaboration. While ALFRED \citep{shridhar2020alfredbenchmarkinterpretinggrounded} and BEHAVIOR-1K \citep{li2024behavior} provide extensive task coverage, they model physical states through discrete representations (e.g., binary door states, picked/placed objects) rather than continuous attributes necessary for reasoning about weight, temperature, or material properties. Tool usage evaluation spans from low-level manipulation in RoCo \citep{mandi2024roco} to high-level planning in PARTNR \citep{chang2024partnrbenchmarkplanningreasoning}, yet both maintain static action spaces determined at initialization, preventing assessment of dynamic capability acquisition. Recent multi-agent benchmarks including TDW-MAT \citep{zhang2024buildingcooperativeembodiedagents} and EmbodiedBench \citep{yang2025embodiedbench} advance collaboration evaluation through load constraints and task allocation optimization, but rely on explicit task division instructions or efficiency-driven participation rather than collaboration emerging from physical constraints. 

In contrast, \framework introduces continuous property reasoning with 6,381 distinct attributes, dynamic tool-capability binding that expands action spaces during execution, and implicit collaboration where agents must autonomously recognize when tasks exceed individual capacities. This fundamentally shifts evaluation from instruction compliance to constraint-based reasoning. A comprehensive comparison is in Appendix~\ref{sec:related_works}.

%% file: sections/3_method.tex
\section{Framework}
\label{sec:method}

We present \framework, a comprehensive framework for evaluating agent reasoning in embodied tasks. Our framework addresses the fundamental challenge of assessing whether language models understand embodied principles. We achieve this through three key design principles: (1) tasks must require reasoning about physical properties and constraints rather than following explicit instructions, (2) agent capabilities should dynamically evolve based on tool acquisition rather than remaining static, and (3) collaboration needs should emerge from task requirements rather than predetermined protocols.

\subsection{Task Design and Formalization}

\paragraph{Environment Representation.}
We formalize embodied environments as directed graphs $G_t = (V_t, E_t, A_t)$ that capture the essential structure of physical spaces~\citep{côté2019textworldlearningenvironmenttextbased, yuan2019countingexploregeneralizetextbased}. The node set $V_t$ encompasses three entity types: spatial nodes representing rooms and areas, object nodes for interactive items, and agent nodes for autonomous entities. Each node maintains an attribute dictionary $A_t$ storing continuous physical properties such as weight, temperature, material composition, and geometric dimensions. The edge set $E_t$ encodes spatial relationships through static containment relations (e.g., ``in'', ``on'') and dynamic proximity relations $E_{\text{near}}$ that track which objects fall within an agent's interaction range. This graph representation enables efficient reasoning about spatial constraints while avoiding the computational overhead of continuous 3D simulation.

\paragraph{Task Formalization.}
Each evaluation task is defined as a tuple $\mathcal{T} = (S_{\text{init}}, I, G_{\text{goal}}, \mathcal{A}_{\text{task}})$, where $S_{\text{init}}$ specifies the initial environment state, $I$ provides the natural language instruction, $G_{\text{goal}}$ defines success conditions through logical predicates, and $\mathcal{A}_{\text{task}}$ identifies participating agents. The evaluation objective is to assess whether agents can generate an action sequence $\Pi = (\pi_1, \ldots, \pi_T)$ that transforms the environment from $S_{\text{init}}$ to a terminal state $S_{\text{final}}$ satisfying all predicates in $G_{\text{goal}}$. This formalization captures both the planning and execution aspects of embodied reasoning.

\subsection{Hierarchical Task Taxonomy}

Our evaluation framework organizes tasks along cognitive complexity (L0: instruction following, L1: dimensional reasoning, L2: compound tasks), with agent configuration (single vs. multi-agent) as a secondary axis. This structure enables systematic assessment of how reasoning capabilities scale with task demands. Representative examples for each task category are provided in Appendix~\ref{sec:task_examples}.

\paragraph{L0: Instruction Following.}
\textbf{\textit{Direct Command (DC)}} tasks require straightforward instruction following, such as ``place cup\#1 on table\#1,'' establishing baseline comprehension abilities. \textbf{\textit{Explicit Collaboration (EC)}} tasks provide clear coordination directives, such as ``Agent A and Agent B cooperate to open the heavy cabinet,'' testing fundamental synchronization abilities.

\paragraph{L1: Dimensional Reasoning.}
\textbf{\textit{Tool Use (TU)}} tasks demand recognizing capability gaps and acquiring right tools. For instance, ``clean the table'' requires agents to identify that cleaning actions are unavailable in their base action set $\mathcal{A}_i$, locate cleaning tools, and execute $\texttt{grasp}(v_{\text{tool}})$ to dynamically expand their capabilities. \textbf{\textit{Attribute Reasoning (AR)}} tasks require comparing continuous properties to identify targets (e.g., ``move the heaviest cup'' requires solving $v^* = \arg\max_{v \in V_{\text{cups}}} A_t(v, \text{weight})$). \textbf{\textit{Implicit Collaboration (IC)}} removes explicit instructions, requiring agents to autonomously recognize when tasks exceed individual capabilities. For example, ``move the dining table to the storage room'' requires agents to infer that $A_t(v_{\text{table}}, \text{weight}) > C_{\max}(i)$ for any individual agent $i$, necessitating collaborative effort.

\paragraph{L2: Compound Tasks.}
\textbf{\textit{Compound Reasoning (CR)}} tasks integrate multiple challenges, such as ``clean the heaviest table,'' requiring simultaneous attribute comparison, tool acquisition, and multi-step planning. \textbf{\textit{Compound Collaboration (CC)}} combines all elements, such as ``cooperatively repair the malfunctioning television,'' demanding tool acquisition, capability assessment, and coordinated execution.

\subsection{\simulator: Efficient Environment Simulation}

\paragraph{State Representation and Updates.}
\simulator employs text-based environment modeling to achieve efficient simulation at scale. The graph structure $G_t$ maintains spatial relationships through topological connections rather than continuous coordinates, eliminating expensive collision detection while preserving essential spatial constraints. State updates follow an incremental approach where actions modify only directly affected nodes and edges. For instance, when an agent executes $\texttt{GOTO}(\text{table})$, the system updates only the relevant proximity relations in $E_{\text{near}}$ rather than recomputing global spatial relationships.

\paragraph{Dynamic Capability Management.}
A key innovation in \simulator is the dynamic tool-capability binding system. Agent actions are partitioned into basic actions (movement, grasping, opening) available to all agents, and tool-dependent actions (cleaning, heating, repairing) that require specific tools. Each tool object maintains a \texttt{capability} attribute specifying which actions it enables. When an agent grasps a tool, the system dynamically binds the associated capabilities to the agent's action set. Upon releasing the tool, these capabilities are automatically unbound. This mechanism enables realistic modeling of how agents extend their abilities through tool use, moving beyond the static action spaces of existing benchmarks.

\paragraph{Emergent Collaboration.}
\simulator supports collaboration that emerges from physical constraints rather than explicit programming. When agents attempt actions on objects whose properties exceed individual capabilities, the system enables collaboration request mechanisms. For instance, if an agent attempts to move an object where $A_t(v, \text{weight}) > C_{\max}(\text{agent})$, it can initiate collaboration by identifying suitable partners and coordinating joint actions. The system validates preconditions for all participating agents and maintains consistency throughout collaborative execution, ensuring realistic multi-agent interactions.

\subsection{Automated Benchmark Generation}

\begin{figure*}
    \centering
    \includegraphics[width=1.0\textwidth,trim=0 13 0 0pt, clip]{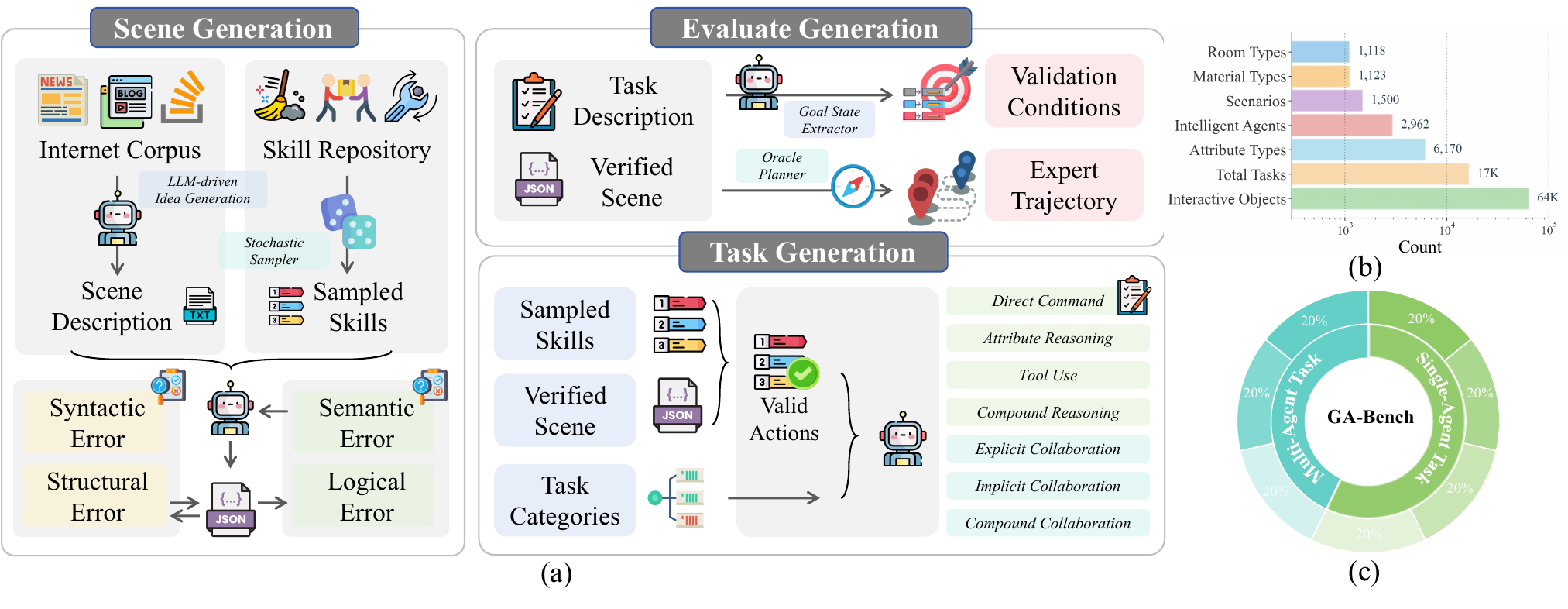}
    \caption{\framework automated benchmark generation and evaluation framework. (a) Four-stage generation pipeline combining LLMs with rule-based validation: scene generation from internet corpus, task generation with skill sampling, evaluation logic extraction, and expert trajectory generation with human validation. (b) \benchmark statistics: 1,500 scenarios, 64K objects, 6K attribute types, spanning diverse domains and material compositions. (c) Balanced task distribution across seven categories spanning single-agent (Direct Command, Tool Use, Attribute Reasoning, Compound Reasoning) and multi-agent tasks (Explicit/Implicit/Compound Collaboration).}
    \label{fig:data-generation-and-distribution}
    \vspace{-1em}
\end{figure*}

\paragraph{Generation Pipeline.}
Creating diverse, physically consistent scenarios at scale requires careful orchestration of neural generation and symbolic validation. As shown in Figure~\ref{fig:data-generation-and-distribution}, our pipeline operates in four stages, each combining the creative capabilities of large language models with rule-based consistency checking. This hybrid approach enables generating thousands of unique scenarios while maintaining physical realism and task solvability. A four-stage walkthrough for one scenario is in Appendix~\ref{sec:appendix_data_pipeline_case}.

\paragraph{Scene and Task Generation.}
Scene generation begins with semantic seeds extracted from diverse text sources~\citep{li2024datacomp, xu2025wizardlmempoweringlargepretrained, wang2023selfinstructaligninglanguagemodels}, which guide a neural generator $g_{\text{scene}}$ in creating structured environment descriptions. The generator, implemented using high-temperature language models for diversity, produces initial scenes $S_0$ containing objects, spatial layouts, and agent configurations. Task generation follows a two-stage process: first, an environment analyzer $C_{\text{env}}$ extracts feasible actions based on scene structure, then a task generator $g_{\text{task}}$ creates instructions anchored in physical possibilities. This grounding prevents generation of impossible tasks while maintaining creative diversity.

\paragraph{Evaluation Logic and Trajectories.}
For each task, we automatically derive evaluation criteria by parsing the instruction and scene to extract minimal state changes required for success. This produces a goal predicate set $G_{\text{goal}}$ that serves as an objective success measure. Expert trajectories are generated using oracle agents with complete environmental knowledge, creating high-quality demonstrations for each task. These trajectories undergo filtering to remove suboptimal sequences, providing ideal solutions for comparison and learning.

\begin{figure}[htbp]
    \centering
    \includegraphics[width=\linewidth]{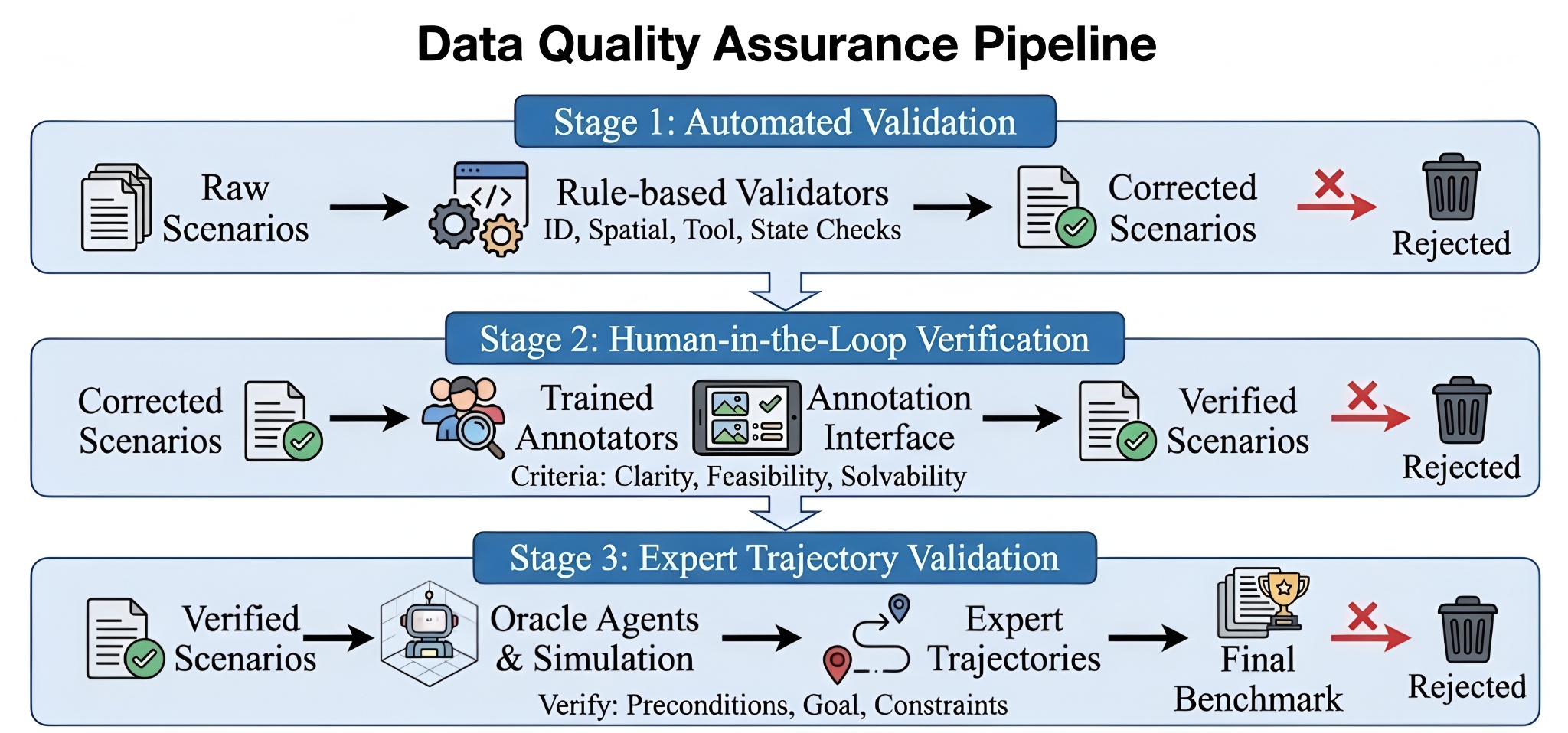}
    \caption{Overview of the Data Quality Assurance Pipeline. The pipeline consists of three sequential stages: (1) Automated Validation for rule-based error correction, (2) Human-in-the-Loop Verification for semantic and physical feasibility checks, and (3) Expert Trajectory Validation to ensure genuine solvability via oracle agents.}
    \label{fig:qa_pipeline}
    \vspace{-1em}
\end{figure}

\paragraph{Quality Assurance.}
As shown in Figure~\ref{fig:qa_pipeline}, generated scenarios undergo a three-stage validation pipeline. Automated validators check structural consistency and physical feasibility, correcting 87.6\% of detected errors. Trained human annotators then verify scenarios against explicit criteria for instruction clarity, feasibility, and solvability (Cohen's $\kappa = 0.84$). Finally, oracle agents validate expert trajectories through simulation replay. This rigorous process yielded 1,500 validated scenarios from 2,100 initial candidates (71.4\% acceptance rate). Detailed procedures, annotation interface, and quality threshold examples are provided in Appendix~\ref{sec:quality_assurance}.

\subsection{Benchmark Statistics and Coverage}

\benchmark encompasses 1,500 scenarios across 11 domains including laboratory (39\%), office (19\%), industrial (12\%), and medical environments, containing 64,057 interactive objects with rich physical properties. The dataset maintains careful balance across our task taxonomy: 65\% single-agent tasks spanning all complexity levels, and 35\% multi-agent tasks with emphasis on implicit collaboration scenarios requiring genuine coordination reasoning. With 6,381 distinct property types and 214 action types, \benchmark provides comprehensive coverage of embodied reasoning challenges while maintaining tractable evaluation scope. Detailed statistics are provided in Appendix~\ref{sec:dataset_statistics}.

%% file: sections/4_experiment.tex
\section{Experiments}

\begin{table*}[t]
\centering
\small
\setlength{\tabcolsep}{4pt}
\begin{tabular*}{\textwidth}
{@{}l@{\extracolsep{\fill}}*{14}{c}>{\columncolor{gray!15}}c@{}}
\toprule
\multirow{4}{*}{\textbf{Model}} &
\multicolumn{4}{c}{\textbf{L0: Instruction Following}} &
\multicolumn{6}{c}{\textbf{L1: Dimensional Reasoning}} &
\multicolumn{4}{c}{\textbf{L2: Compound Tasks}} &
\multicolumn{1}{c}{\multirow{3}{*}{\textbf{Overall}}} \\
\cmidrule(lr){2-5} \cmidrule(lr){6-11} \cmidrule(lr){12-15}
& \multicolumn{2}{c}{\textbf{Direct}} & \multicolumn{2}{c}{\textbf{Explicit}} & \multicolumn{2}{c}{\textbf{Tool}} & \multicolumn{2}{c}{\textbf{Attribute}} & \multicolumn{2}{c}{\textbf{Implicit}} & \multicolumn{2}{c}{\textbf{Compound}} & \multicolumn{2}{c}{\textbf{Compound}} & \multicolumn{1}{c}{} \\
& \multicolumn{2}{c}{\textbf{Command}} & \multicolumn{2}{c}{\textbf{Collab.}} & \multicolumn{2}{c}{\textbf{Use}} & \multicolumn{2}{c}{\textbf{Reasoning}} & \multicolumn{2}{c}{\textbf{Collab.}} & \multicolumn{2}{c}{\textbf{Reasoning}} & \multicolumn{2}{c}{\textbf{Collab.}} & \multicolumn{1}{c}{} \\
\cmidrule(lr){2-3} \cmidrule(lr){4-5} \cmidrule(lr){6-7} \cmidrule(lr){8-9} \cmidrule(lr){10-11} \cmidrule(lr){12-13} \cmidrule(lr){14-15}
& \textbf{SR} & \textbf{Step} & \textbf{SR} & \textbf{Step} & \textbf{SR} & \textbf{Step} & \textbf{SR} & \textbf{Step} & \textbf{SR} & \textbf{Step} & \textbf{SR} & \textbf{Step} & \textbf{SR} & \textbf{Step} & \multicolumn{1}{c}{\textbf{SR}} \\
\midrule
\rowcolor{gray!10}\multicolumn{16}{@{}l@{}}{\textit{Closed-source Models}} \\
\midrule
\centering Claude-Opus-4.6 & 95.3 & 9.4 & 92.5 & 5.8 & \underline{\textbf{89.9}} & 11.6 & 55.7 & 10.3 & 91.0 & 8.2 & \underline{\textbf{79.4}} & 12.4 & \underline{\textbf{53.0}} & 12.7 & 79.5 \\
\centering Claude-Sonnet-4.6 & 95.9 & 10.2 & 92.0 & 6.3 & 84.6 & 11.8 & 70.1 & 10.2 & 91.5 & 7.8 & 74.1 & 12.7 & 52.5 & 12.0 & \underline{\textbf{80.1}} \\
\centering GPT-4o & \underline{\textbf{96.6}} & 12.9 & 90.0 & 13.9 & 80.0 & 13.6 & \underline{\textbf{77.8}} & 12.3 & 77.5 & 14.4 & 69.2 & 14.5 & 32.0 & 22.9 & 74.7 \\
\centering Gemini-2.5-Flash & 90.5 & 11.0 & 88.5 & 8.4 & 82.3 & 16.5 & 56.3 & 17.5 & 85.5 & 7.1 & 59.4 & 20.0 & 40.5 & 16.2 & 71.9 \\
\centering GLM-5 & 93.5 & 9.8 & \underline{\textbf{93.0}} & 6.2 & 84.6 & 11.6 & 61.7 & 10.0 & \underline{\textbf{92.5}} & 8.3 & 72.9 & 12.9 & 48.0 & 11.7 & 78.0 \\
\centering GLM-5-Turbo & 94.1 & 10.2 & 92.5 & 6.7 & 85.8 & 12.0 & 49.1 & 10.7 & 88.5 & 8.6 & 74.1 & 13.5 & 46.0 & 12.8 & 75.7 \\
\centering Qwen3.6-Plus & 92.9 & 9.7 & 90.5 & 6.8 & 85.2 & 11.2 & 52.1 & 9.4 & 86.5 & 8.5 & 70.6 & 11.9 & 40.0 & 11.5 & 74.0 \\
\midrule
\rowcolor{gray!10}\multicolumn{16}{@{}l@{}}{\textit{Reasoning-specialized Models}} \\
\midrule
\centering Deepseek-R1 & \textbf{94.1} & 10.3 & \textbf{92.0} & 7.4 & \textbf{85.8} & 14.1 & 41.9 & 12.2 & \textbf{84.5} & 9.6 & \textbf{70.6} & 16.2 & \textbf{48.5} & 12.5 & \textbf{73.9} \\
\centering QwQ-32B & 85.2 & 10.3 & 88.0 & 8.5 & 73.4 & 13.0 & \textbf{44.9} & 11.0 & 84.0 & 8.3 & 54.1 & 13.6 & 36.5 & 19.0 & 66.6 \\
\midrule
\rowcolor{gray!10}\multicolumn{16}{@{}l@{}}{\textit{Open-source Foundation Models}} \\
\midrule
\centering Kimi-K2 & \textbf{94.7} & 10.4 & \textbf{85.0} & 5.8 & \textbf{86.4} & 13.7 & \textbf{63.5} & 10.0 & \textbf{87.0} & 7.8 & \textbf{74.7} & 14.6 & \textbf{37.0} & 11.4 & \textbf{75.5} \\
\centering Deepseek-V3 & 91.1 & 11.2 & 82.0 & 9.4 & 82.3 & 15.1 & 56.3 & 10.3 & 63.0 & 9.7 & 67.1 & 16.0 & 36.0 & 20.2 & 68.3 \\
\centering Qwen2.5-72B & 89.7 & 14.7 & 56.0 & 24.1 & 56.4 & 21.7 & 57.4 & 17.2 & 65.4 & 15.6 & 66.7 & 21.1 & 28.6 & 29.5 & 60.0 \\
\centering Llama3.1-8B & 24.9 & 34.4 & 4.0 & 3.5 & 8.3 & 34.6 & 9.9 & 34.8 & 1.5 & 2.1 & 12.4 & 34.3 & 0.0 & 3.4 & 8.7 \\
\centering Qwen2.5-7B & 40.2 & 24.1 & 38.5 & 25.0 & 15.4 & 31.7 & 22.2 & 26.6 & 13.5 & 24.1 & 16.5 & 30.5 & 1.0 & 27.2 & 21.0 \\
\centering Qwen2.5-3B & 0.6 & 30.5 & 8.5 & 20.4 & 1.8 & 31.3 & 0.6 & 34.0 & 1.5 & 16.3 & 2.9 & 32.9 & 0.5 & 16.8 & 2.3 \\
\quad + SFT & 76.3 & 15.4 & 22.5 & 29.2 & 45.0 & 24.7 & 33.5 & 22.8 & 5.5 & 28.3 & 36.5 & 24.7 & 1.0 & 27.1 & 31.5 \\
\bottomrule
\end{tabular*}
\caption{Performance across task categories grouped by cognitive complexity (L0: Instruction Following; L1: Dimensional Reasoning; L2: Compound Tasks). Success Rate (SR) measures task completion percentage, Step Count indicates average actions for successful completion. \textbf{Overall} reports the unweighted mean SR across all seven task categories. Bold indicates best in section, underline shows overall best.}
\label{tab:comprehensive_performance}
    \vspace{-1em}
\end{table*}

We systematically evaluate current LLMs on \benchmark to assess their physical reasoning capabilities in embodied tasks. 

\subsection{Experimental Setup}

\paragraph{Model Selection.}
We evaluate 15 representative models spanning three architectural paradigms. Closed-source models include GPT-4o \citep{hurst2024gpt}, Gemini-2.5-Flash \citep{comanici2025gemini}, Claude-Sonnet-4.6, Claude-Opus-4.6, GLM-5 \citep{glm5team2026glm5vibecodingagentic}, GLM-5-Turbo, and Qwen3.6-Plus \citep{yang2025qwen3technicalreport}, representing current commercial state-of-the-art. Open-source foundation models cover a wide parameter range: Deepseek-V3 \citep{liu2024deepseek} at 671B parameters, Kimi-K2 \citep{kimiteam2026kimik2openagentic}, the Qwen2.5 series \citep{team2024qwen2} at 3B, 7B, and 72B parameters, and Llama3.1-8B~\citep{touvron2023llama}. This selection enables analysis of how model scale and architecture affect embodied reasoning. We also include reasoning-specialized models: Deepseek-R1 \citep{guo2025deepseek} and QwQ-32B~\citep{qwq32b}, employing explicit chain-of-thought reasoning.

\paragraph{Evaluation Protocol.}
All models undergo identical evaluation to ensure fair comparison. We implement partial observability where agents must explore environments to discover object locations and properties, reflecting realistic deployment conditions. Each model completes 2,800 test scenarios across seven task categories with three independent runs for statistical reliability. We standardize prompts, environment descriptions, and action vocabularies across all models, with tool-dependent actions dynamically enabled based on context. This design ensures performance differences reflect reasoning capabilities rather than implementation artifacts. Detailed experimental configurations are provided in Appendix~\ref{sec:appendix_prompts}.

\paragraph{Implementation Details.}
To test whether supervised learning addresses reasoning limitations, we fine-tune Qwen2.5-3B on 20,346 instruction–action pairs distilled from 1,942 optimal Qwen2.5-72B trajectories collected with complete environmental access (hyperparameters in Appendix~\ref{sec:hyperparameters}). Evaluation uses two configurations: single-agent scenarios test individual reasoning, while multi-agent scenarios employ centralized coordination where one model controls all agents with full state visibility, isolating collaborative reasoning from communication artifacts.

\subsection{Main Results}
\label{sec:main_results}

\paragraph{Instruction following is saturated at frontier scale.}
At L0, all ten frontier models cluster narrowly: $85.2$--$96.6\%$ on
Direct Command and $85.0$--$93.0\%$ on Explicit Collaboration. The
same models spread sharply at L1: Tool Use ranges $73.4$--$89.9\%$,
Implicit Collaboration $77.5$--$92.5\%$, and Attribute Reasoning more
than twice as wide at $41.9$--$77.8\%$. Frontier-scale discrimination
comes from the action-grounding categories, not from L0.

\begin{figure}[t]
    \centering
    \includegraphics[width=\columnwidth]{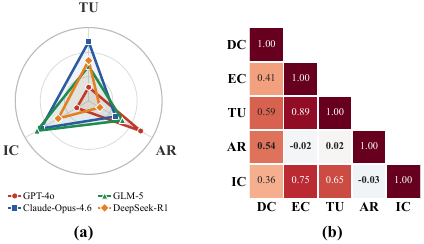}
    \caption{\textbf{(a)} Capability profiles over the three L1 primitives (TU, AR, IC). \textbf{(b)} Pairwise Pearson correlations across 12 frontier models on five categories (DC, EC, TU, AR, IC).}
    \label{fig:l1_radar}
    \label{fig:l1_correlation}
\end{figure}

\paragraph{Aggregate scores hide structurally different capability
profiles.}
Models with comparable overall performance can have inverted
category-level profiles (Figure~\ref{fig:l1_radar}a). GPT-4o and
DeepSeek-R1 sit within one point overall ($74.7\%$ vs.\ $73.9\%$),
but GPT-4o leads on AR ($77.8\%$, the highest in the set) while
DeepSeek-R1 drops to the lowest AR score ($41.9\%$). On IC, the
direction reverses: GPT-4o reaches $77.5\%$, whereas DeepSeek-R1
reaches $84.5\%$. The $35.9$-point AR gap and the reversed IC gap
describe two qualitatively different models that an aggregate score
would treat as nearly equivalent. Per-category breakdown is therefore
not a refinement but the primary diagnostic signal that GroundAct
provides.

\paragraph{L2 amplifies L1 gaps into composition cost.}
Composing grounding dimensions compounds the difficulty rather than
averaging it. Compound Reasoning spans $54.1$--$79.4\%$, while
Compound Collaboration drops to $32.0$--$53.0\%$, the lowest range
among all categories. Strong L1 performance therefore does not
guarantee robust composition: GPT-4o averages $78.4\%$ across the
three L1 categories but falls to $32.0\%$ on Compound Collaboration.
L2 is not merely a harder version of L1 along the same axis; it is a
regime where multiple grounding dimensions interact. This motivates
the bottleneck-level analyses in Section~\ref{sec:analysis}.

\subsection{Analysis}
\label{sec:analysis}

\paragraph{Task categories expose different action-grounding bottlenecks.}
GroundAct is designed not only to rank models but to localize where
action grounding fails. Pairwise Pearson correlations of success rates
across 12 frontier models (Figure~\ref{fig:l1_correlation}) show that
EC, TU, and IC form a correlated cluster ($r=0.65$--$0.89$): all three
involve reasoning over action-space feasibility or coordination signals,
whether a coordination directive (EC), a missing capability (TU), or a
partner-dependent precondition (IC). AR is nearly uncorrelated with this
cluster ($r\in[-0.03, 0.02]$), indicating that continuous-attribute
grounding is a distinct bottleneck; its moderate $r=0.54$ with DC
suggests a shared instruction-execution component rather than the same
feasibility reasoning. This separation explains how aggregate scores
hide qualitatively different failures: GPT-4o and DeepSeek-R1 land
within one point overall ($74.7\%$ vs.\ $73.9\%$), yet GPT-4o leads by
$35.9$ points on AR, while DeepSeek-R1 leads by $7.0$ points on IC.
GroundAct surfaces this structure by separating execution, attribute
grounding, and action-space feasibility rather than collapsing them into
a single success rate.

\begin{figure}[t]
    \centering
    \includegraphics[width=\columnwidth]{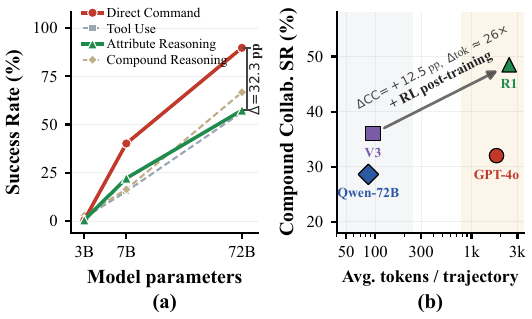}
    \caption{Scaling and reasoning specialization. \textbf{(a)} Qwen2.5 series success rate across task categories at 3B/7B/72B. \textbf{(b)} Per-trajectory token cost: DeepSeek-R1 vs.\ foundation models.}
    \label{fig:scaling}
\end{figure}

\paragraph{Scaling and reasoning training shift bottlenecks
differently.}
Per-category results show that scaling and reasoning post-training do
not uniformly improve action grounding; they shift different
bottlenecks in different directions. Across the Qwen2.5 series
(Figure~\ref{fig:scaling}a), Direct Command rises from $0.6\%$ at 3B
to $89.7\%$ at 72B while Attribute Reasoning reaches only $57.4\%$:
scaling raises the execution floor but leaves attribute grounding
$32.3$pp behind. Reasoning post-training shifts the profile in another
direction: relative to DeepSeek-V3, DeepSeek-R1 improves Implicit
Collaboration ($63.0\%\!\to\!84.5\%$), Compound Reasoning
($67.1\%\!\to\!70.6\%$), and Compound Collaboration
($36.0\%\!\to\!48.5\%$) but drops on Attribute Reasoning
($56.3\%\!\to\!41.9\%$). Its larger token budget
($1{,}400$--$2{,}457$ vs.\ $60$--$110$ tokens per trajectory;
Figure~\ref{fig:scaling}b) coincides with gains on collaboration and
composition but not with improved continuous-attribute grounding.
Aggregate scores would record only the net effect; GroundAct surfaces
the profile shift because it separates execution, attribute
grounding, collaboration inference, and composition rather than
reporting only a single success rate.

\begin{figure}[t]
    \centering
    \includegraphics[width=\columnwidth]{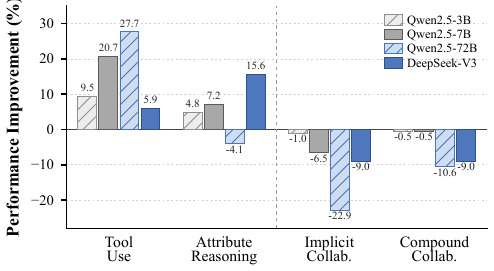}
    \caption{World Graph ablation. Complete-state observability helps Tool Use at every scale but harms Implicit Collaboration.}
    \label{fig:world_graph}
\end{figure}

\paragraph{Complete state helps search but hurts constraint filtering.}
We compare partial observability with a World Graph condition that
exposes the complete environment graph
(Figure~\ref{fig:world_graph}). Tool Use improves at every scale,
reaching $+27.6$\,pp on Qwen2.5-72B ($56.4\%\!\to\!84.0\%$),
indicating that tool reasoning is search-bound: providing object
locations removes the exploration step. Implicit Collaboration
degrades at every scale, with Qwen2.5-72B losing $22.9$\,pp
($65.4\%\!\to\!42.5\%$) and DeepSeek-V3 losing $9.0$\,pp
($63.0\%\!\to\!54.0\%$), indicating a filter-bound bottleneck:
complete state surfaces task-irrelevant attributes that obscure
whether collaboration is physically required. Attribute Reasoning
shows mixed effects, suggesting that property grounding is not
reducible to search or filtering alone. The same observability
intervention therefore moves different bottlenecks in opposite
directions, which GroundAct surfaces by evaluating each grounding
dimension as a separate category.

\paragraph{Imitation transfers single-agent execution but not 
coordination triggers.}
Supervised fine-tuning improves single-agent performance 
dramatically but transfers poorly to multi-agent coordination. 
Qwen2.5-3B SFT lifts Direct Command from $0.6\%$ to $76.3\%$ and 
Tool Use from $1.8\%$ to $45.0\%$, while Implicit Collaboration 
rises only from $1.5\%$ to $5.5\%$. This asymmetry suggests that 
imitation teaches the surface action patterns required for
instruction following and tool acquisition, but not the latent
trigger for when collaboration is physically required.

\begin{figure}[t]
    \centering
    \includegraphics[width=\columnwidth,clip]{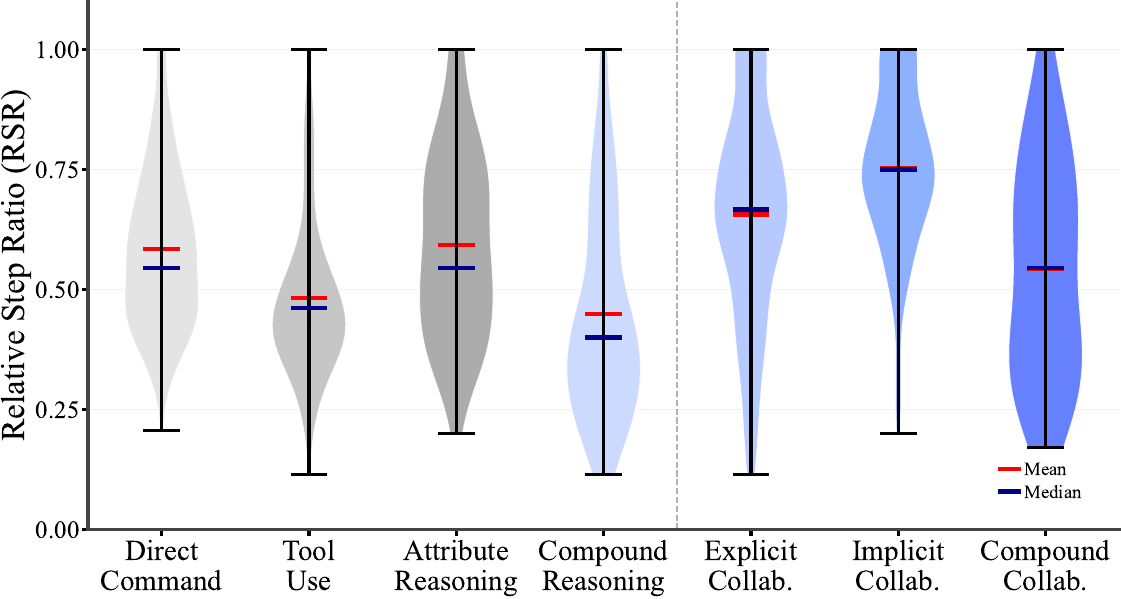}
    \caption{Relative Step Ratio (RSR) distributions across task categories. Multi-agent tasks show lower median and higher variance; Compound Collaboration is bimodal.}
    \label{fig:rsr}
\end{figure}

\paragraph{Multi-agent compound tasks expose strategic polarization.}
We measure execution efficiency as relative step ratio 
(RSR $= L_{\text{expert}} / L_{\text{model}}$). Single-agent tasks 
yield consistent moderate efficiency (median RSR 
$0.40$--$0.55$), while multi-agent tasks show both lower median
and higher variance (Figure~\ref{fig:rsr}). 
Compound Collaboration produces a striking bimodal distribution:
trajectories cluster at simple sequential execution or complex
parallel coordination, with the intermediate strategy region
nearly empty. Qualitative trajectories and category-level failure
modes are in Appendix~\ref{sec:case_study} and~\ref{sec:failure_mode_analysis}.

%% file: sections/5_conclusion.tex
\section{Conclusion}
We introduced GroundAct, a benchmark for evaluating whether LLM
agents can infer action feasibility from environmental state when
instructions are underspecified. Evaluating 15 models across its
seven categories, we find that action grounding dissociates into
distinct bottlenecks, that environmental information has
non-monotonic effects across these bottlenecks, and that supervised
fine-tuning transfers execution but not the capability-boundary
inference required for collaboration. These findings suggest that
progress in agent competence requires interventions targeted at
specific grounding dimensions rather than uniform scaling, and that
diagnosing such progress requires benchmarks that vary grounding
demands independently rather than collapsing them into a single
success rate. See Appendix~\ref{sec:discussion} for further discussion and future directions.

%% file: sections/6_appendix.tex
\clearpage
\section{Appendix}

\begin{table*}[htbp]
\small
\centering
\begin{tabular}{@{}lccccccc@{}}
\toprule
\textbf{Dataset} & \textbf{Scenes} & \textbf{Domain} & \textbf{Task Types} & \textbf{Actions} & \textbf{Action Space} & \textbf{Collab.} & \textbf{Auto Gen.} \\
\midrule
ALFRED & 120 & House & D & 7 & Static & --- & $\times$ \\
PARTNR & 60 & House & D & 11 & Static & Effic. & $\checkmark$ \\
BEHAVIOR-1K & 50 & Diverse & D,T & 6 & Static & --- & $\times$ \\
WAH & 7 & House & D & 10 & Static & Effic. & $\times$ \\
TDW-MAT & 6 & House & D,E & 7 & Static & Effic. & $\times$ \\
C-WAH & 6 & House & D,E & 7 & Static & Effic. & $\times$ \\
Overcooked & 5 & Kitchen & E,I,C & 6 & Static & Effic. & $\times$ \\
\midrule
\textbf{\framework} & \textbf{1.5K} & \textbf{Diverse} & \textbf{D,A,T,R,E,I,C} & \textbf{214} & \textbf{Dynamic} & \textbf{Phys.} & \textbf{\checkmark} \\
\bottomrule
\end{tabular}
\caption{Comparison of embodied AI datasets and benchmarks. Task types: D (Direct Command), A (Attribute Reasoning), T (Tool Use), R (Compound Reasoning), E (Explicit Collaboration), I (Implicit Collaboration), C (Compound Collaboration). Actions: number of available action types. Collab.: collaboration mechanism (Effic. = efficiency-based, Phys. = physical necessity-driven). Auto Gen.: automated task generation capability. Our framework uniquely combines comprehensive task coverage, dynamic action spaces, physical necessity-driven collaboration, and scalable automated generation.}
\label{tab:dataset_comparison}
\end{table*}

\subsection{Related Work}
\label{sec:related_works}

\paragraph{Embodied Intelligence Benchmarks}
The embodied intelligence evaluation landscape has established diverse benchmark frameworks spanning navigation to complex manipulation tasks~\citep{puig2023habitat,li2021igibson}. ALFRED \citep{shridhar2020alfredbenchmarkinterpretinggrounded} provides foundational standards for instruction-following task evaluation, while BEHAVIOR-1K \citep{li2024behavior} extends coverage to 1,000 daily activity scenarios. These benchmarks effectively assess task execution capabilities, yet physical property modeling predominantly employs discrete state representations, such as binary door operations and object pickup/placement, with limited requirements for reasoning about continuous attributes including weight, hardness, and temperature. Our framework addresses this limitation by introducing continuous physical property reasoning tasks that require agents to compare object attributes and make decisions based on physical constraints.

\paragraph{Embodied Tool Use}
Tool usage evaluation in embodied AI exhibits stratified characteristics across different complexity levels. RoCo \citep{mandi2024roco} focuses on low-level manipulation skills such as grasping precision, while high-level benchmarks like PARTNR \citep{chang2024partnrbenchmarkplanningreasoning} adopt predefined tool configurations with agent action spaces fixed at task initialization. This design effectively simplifies evaluation complexity but presents limitations in assessing dynamic tool reasoning capabilities based on task requirements. Current approaches typically provide static tool sets \citep{chowa2025languageactionreviewlarge, fung2025embodiedaiagentsmodeling}, preventing evaluation of how agents should reason about capability gaps and tool acquisition needs. Our framework introduces dynamic tool acquisition mechanisms, requiring agents to autonomously infer tool requirements and expand their action spaces based on task demands, thereby supplementing existing evaluation dimensions.

\paragraph{Multi-Agent Collaboration}
Multi-agent embodied intelligence evaluation has emerged as a significant research direction, with related work achieving valuable progress in collaboration modeling~\citep{sun2024aml,wang2024large}. PARTNR evaluates multi-agent planning capabilities through heterogeneous task design, TDW-MAT \citep{zhang2024buildingcooperativeembodiedagents} creates collaborative scenarios using load capacity constraints, and EmbodiedBench \citep{yang2025embodiedbench} focuses on task allocation and execution optimization. Existing approaches primarily model collaboration requirements through two pathways: explicit collaboration instructions that clearly specify inter-agent task division, and efficiency optimization that drives multi-agent participation to enhance task completion speed. However, real-world collaboration decisions often stem from physical constraints rather than external instructions or efficiency considerations. Our framework employs implicit collaboration design requiring agents to autonomously assess whether tasks exceed single-agent capability ranges based on physical constraints and determine collaboration strategies accordingly, transforming collaboration judgment from external instructions to constraint-driven internal reasoning processes.

\subsection{Task Category Examples}
\label{sec:task_examples}

\begin{figure*}[h]
    \centering
    \includegraphics[width=\textwidth]{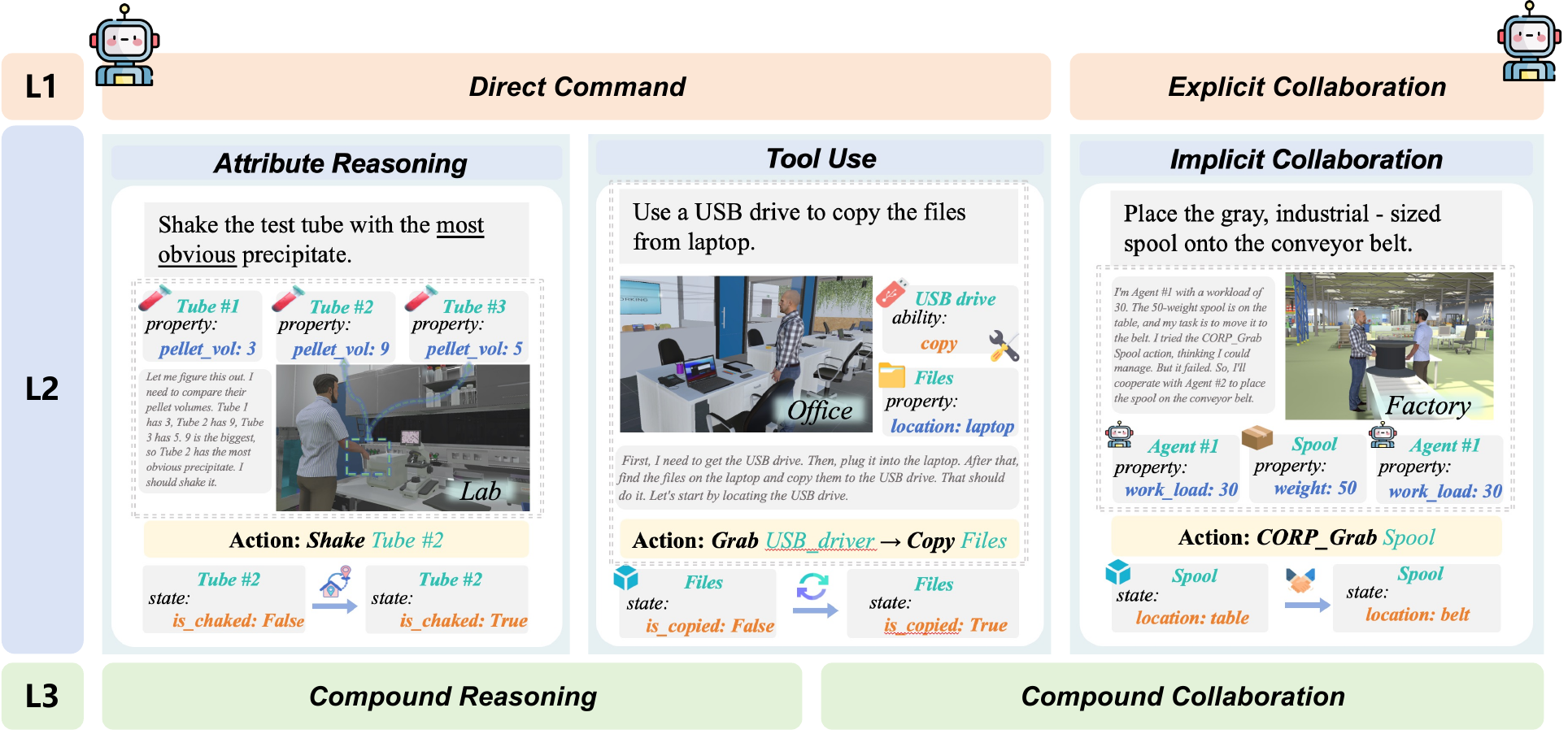}
    \caption{Representative examples for each task category in \benchmark. Left column shows single-agent tasks progressing from Direct Command (L0) through Attribute Reasoning and Tool Use (L1) to Compound Reasoning (L2). Right column shows multi-agent tasks from Explicit Collaboration (L0) through Implicit Collaboration (L1) to Compound Collaboration (L2). Each example includes the task instruction, relevant object properties, agent reasoning process, and resulting state transitions.}
    \label{fig:task_category}
\end{figure*}

Figure~\ref{fig:task_category} illustrates our hierarchical task taxonomy with representative examples, organized by agent configuration (single-agent vs. multi-agent) and cognitive complexity level (L0-L2). The figure demonstrates the progression from basic instruction following to advanced reasoning requiring dynamic capability acquisition and autonomous coordination. Table~\ref{tab:task_examples} provides detailed descriptions of the key reasoning requirements for each task category.

\begin{table*}[h]
\small
\centering
\small
\begin{tabular}{@{}p{0.7cm}p{2.8cm}p{4.1cm}p{7.0cm}@{}}
\toprule
\textbf{Level} & \textbf{Task Type} & \textbf{Example Task} & \textbf{Key Reasoning Requirements} \\
\midrule
\multicolumn{4}{l}{\textit{Single-Agent Tasks}} \\
\midrule
L0 & Direct Command & Place the red cup on the kitchen table. & Basic object manipulation and spatial understanding through straightforward instruction following. \\
\addlinespace
L1 & Tool Use & Clean the dirty table in the living room. & Recognizing capability gaps, locating appropriate tools, and dynamically expanding action capabilities through tool acquisition. \\
\addlinespace
L1 & Attribute Reasoning & Move the heaviest box to the storage room. & Comparing continuous physical properties across multiple objects to identify correct manipulation targets. \\
\addlinespace
L2 & Compound Reasoning & Clean the heaviest table in the room. & Integrating attribute comparison, tool acquisition, and multi-step planning simultaneously. \\
\midrule
\multicolumn{4}{l}{\textit{Multi-Agent Tasks}} \\
\midrule
L0 & Explicit Collab. & Agent A and Agent B cooperate to move the heavy dining table. & Fundamental multi-agent synchronization with explicit coordination directives provided. \\
\addlinespace
L1 & Implicit Collab. & Move the piano to the music room. & Autonomously recognizing when tasks exceed individual capabilities without explicit coordination instructions. \\
\addlinespace
L2 & Compound Collab. & Cooperatively repair the malfunctioning television. & Combining tool acquisition, capability assessment, and coordinated execution with autonomous collaboration recognition. \\
\bottomrule
\end{tabular}
\caption{Representative examples for each task category in \benchmark. Tasks span single-agent scenarios (Direct Command, Tool Use, Attribute Reasoning, Compound Reasoning) and multi-agent scenarios (Explicit, Implicit, and Compound Collaboration) across three complexity levels.}
\label{tab:task_examples}
\end{table*}

\input{pics/appendix_scenes/appendix_snippet_batch2.tex}

\subsection{Data Quality Assurance and Validation}
\label{sec:quality_assurance}

A key challenge in LLM-based benchmark generation is ensuring that generated scenarios are physically consistent, unambiguous, and genuinely solvable, as noted in prior work~\citep{li2025viewspatialbenchevaluatingmultiperspectivespatial, shen2023taskbench}. To address this challenge and ensure benchmark reliability at scale, we implement a rigorous three-stage validation pipeline that combines automated error correction, systematic human-in-the-loop verification, and simulation-based trajectory validation, as illustrated in Figure~\ref{fig:qa_pipeline}. This hybrid approach balances scalability with quality control, achieving validation standards comparable to manually curated benchmarks while enabling generation at scale.

\subsubsection{Stage 1: Automated Validation}

The first stage addresses systematic errors that can be detected and corrected programmatically, reducing the burden on human annotators while ensuring basic structural integrity.

\paragraph{Error Analysis and Correction.} Systematic analysis of 2,100 initially generated scenarios identified five primary error categories, summarized in Table~\ref{tab:error_dist}. We developed specialized rule-based validators to automatically detect and correct each error type: ID mismatches are resolved via semantic similarity matching (cosine threshold 0.7), spatial violations are corrected via dimension checking and constraint propagation, tool-capability conflicts are resolved via attribute cross-referencing against our capability ontology, and state inconsistencies are handled via priority-based resolution that preserves task-critical states. This automated pipeline successfully corrected 87.6\% of detected errors. Scenarios with complex logical inconsistencies requiring semantic understanding that could not be automatically corrected (14.1\%) were rejected, yielding 1,804 candidates for human review.

\begin{table}[h]
\centering
\small
\begin{tabular}{p{3.4cm}p{1.2cm}p{1.5cm}}
\toprule
\textbf{Error Type} & \textbf{Frequency} & \textbf{Auto-Correctable} \\
\midrule
ID Mismatch & 32.4\% & 91\% \\
Invalid Spatial Relations & 28.7\% & 85\% \\
Tool-Capability Conflict & 23.1\% & 89\% \\
State Inconsistency & 10.3\% & 94\% \\
Other Logical Errors & 5.5\% & 62\% \\
\midrule
\textbf{Overall} & \textbf{100\%} & \textbf{87.6\%} \\
\bottomrule
\end{tabular}
\caption{Distribution of LLM generation errors and automatic correction rates by category.}
\label{tab:error_dist}
\end{table}

\subsubsection{Stage 2: Human-in-the-Loop Verification}
\label{sec:hitl_verification}

While automated validation effectively handles systematic errors, subtle semantic ambiguities, non-obvious unsolvability, and nuanced physical inconsistencies require human expertise. We therefore implement rigorous human-in-the-loop verification with carefully selected and trained annotators, standardized evaluation criteria, and quantitative quality control measures.

\paragraph{Annotator Selection and Training.} Five annotators were recruited from graduate programs in robotics and embodied AI, each with at least two years of research experience in related domains. All annotators completed a standardized 4-hour training program comprising three components: conceptual training on embodied reasoning principles and physical constraints (1.5 hours), hands-on system familiarization with environment representation and the validation interface (1 hour), and calibration exercises on 20 expert-validated scenarios with group discussion and feedback (1.5 hours). The calibration scenarios were independently validated by two senior researchers to establish reliable ground truth labels. Annotators were required to achieve Cohen's $\kappa \geq 0.75$ against these gold labels before beginning independent annotation; those falling below this threshold received additional training until reaching the required level. During the main annotation phase, each annotator reviewed approximately 720 scenarios (with systematic overlap for agreement computation), averaging 3-4 minutes per scenario to ensure careful evaluation.

\begin{figure*}[h]
    \centering
    \includegraphics[width=0.9\linewidth]{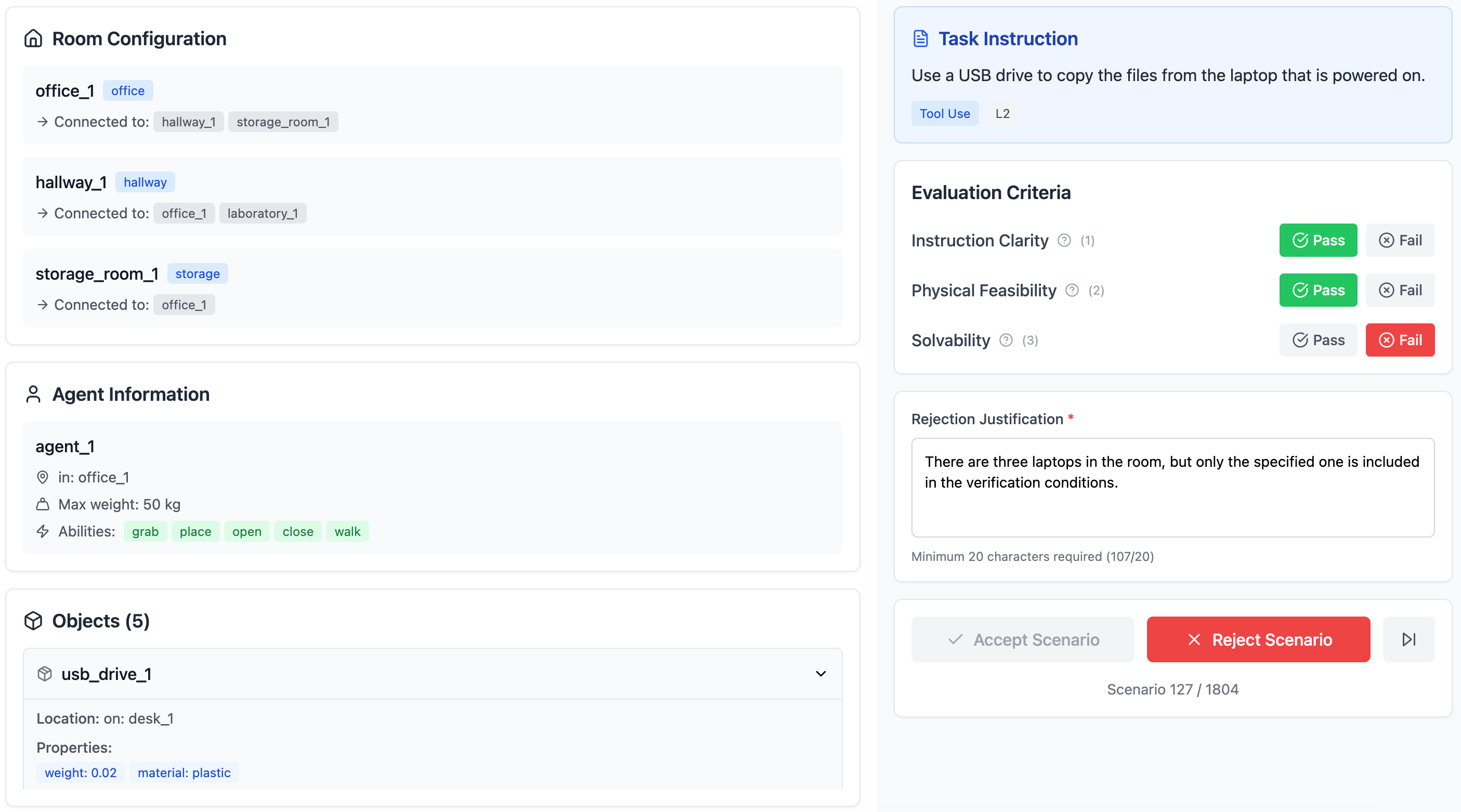}
    \caption{Human annotation interface for scenario verification. The left panel displays structured scene representation including room configuration, agent properties, and interactive objects with their attributes. The right panel shows the task instruction with highlighted key elements (top), three evaluation criteria with binary pass/fail selections (middle), and a rejection justification field requiring written rationale for any failed criterion (bottom).}
    \label{fig:annotation_interface}
\end{figure*}

\paragraph{Annotation Interface.} To ensure consistent and reliable human evaluation, we developed a standardized annotation interface illustrated in Figure~\ref{fig:annotation_interface}. The interface presents three integrated components: (1) a structured scene representation panel displaying room layouts, object properties (weight, material, dimensions), spatial relationships, and agent configurations in a clear hierarchical format; (2) a task instruction panel with automatically highlighted key elements including target objects, required actions, and relevant constraints; and (3) an evaluation panel presenting the three assessment criteria as binary selections along with a mandatory text field for rejection justification. This standardized interface ensures consistent evaluation conditions across all annotators and scenarios, minimizing variance arising from interface interpretation differences and enabling systematic collection of rejection rationales for error analysis.

\paragraph{Annotation Guidelines.} Annotators evaluated each scenario against three binary criteria with explicit operational definitions designed to ensure consistent judgment:
\begin{itemize}[nosep]
    \item \textbf{Instruction Clarity:} The task instruction unambiguously specifies the goal state, target objects, and required actions without relying on implicit assumptions or domain-specific knowledge not provided in the scene description.
    \item \textbf{Physical Feasibility:} All actions required to complete the task respect the physical constraints encoded in the environment, including weight limits, spatial accessibility, tool requirements, and material properties.
    \item \textbf{Solvability:} A valid action sequence exists that transforms the initial environment state to a goal state satisfying all success criteria within the maximum step limit of 25 steps.
\end{itemize}
Each scenario received independent review from two randomly assigned annotators to eliminate individual bias. Acceptance required unanimous approval on all three criteria; any rejection required written justification specifying the failed criterion and the specific issue identified, enabling systematic failure analysis and iterative improvement of the generation pipeline.

\paragraph{Quality Threshold Examples.} To provide concrete grounding for our annotation guidelines and demonstrate the practical application of our quality criteria, Table~\ref{tab:acceptance_examples} presents representative examples of accepted and rejected scenarios across different task categories. These contrastive examples illustrate the key distinctions between scenarios meeting our quality standards and those rejected for ambiguity, physical infeasibility, or unclear collaboration requirements. The examples were selected to highlight common failure modes and the reasoning behind acceptance decisions.

\begin{table*}[t]
\centering
\small
\setlength{\tabcolsep}{5pt}
\renewcommand{\arraystretch}{1.5}
\begin{tabular}{@{} 
>{\raggedright\arraybackslash}p{1.8cm}
>{\raggedright\arraybackslash}p{5.2cm}
>{\centering\arraybackslash}p{1.0cm}
>{\raggedright\arraybackslash}p{6.5cm}
@{}}
\toprule
\textbf{Task Type} & \textbf{Instruction} & \textbf{Decision} & \textbf{Rationale} \\
\midrule
Tool Use & 
Clean the dusty shelf using supplies from the cabinet which is located in the corner of the room next to the window. & 
\textcolor{green!60!black}{\ding{51}} & 
Clear goal specification, explicit tool source, verifiably solvable path exists, and the operation steps are clear and easy to execute. \\

Tool Use & 
Clean everything in the room including the floor, the ceiling, the furniture and all the small ornaments on the shelf. & 
\textcolor{red!70!black}{\ding{55}} & 
Ambiguous task scope; ``everything'' lacks clear termination criteria and target enumeration, which may lead to endless work and cannot be completed effectively. \\

Attribute Reasoning & 
Move the heaviest box which is marked with red label to storage room B on the second floor of the warehouse. & 
\textcolor{green!60!black}{\ding{51}} & 
Unambiguous quantitative comparison criterion (weight), clearly specified destination, and the marking of the box avoids confusion. \\

Attribute Reasoning & 
Move the large box to storage which is not clearly designated with a specific room number or location. & 
\textcolor{red!70!black}{\ding{55}} & 
``Large'' is subjectively interpretable; multiple boxes may satisfy this criterion, and the storage location is not clear, leading to operational ambiguity. \\

Implicit Collab. & 
Move the 150kg industrial cabinet which is made of solid steel to warehouse located 500 meters away from the current location. & 
\textcolor{green!60!black}{\ding{51}} & 
Object weight (150kg) clearly exceeds single-agent capacity (50kg); collaboration requirement is inferable from physical constraints, and the material and location information are clear. \\

Implicit Collab. & 
Move the cabinet which is of unknown material and weight to warehouse without any additional information. & 
\textcolor{red!70!black}{\ding{55}} & 
Cabinet weight (30kg) is within single-agent capacity; collaboration necessity unclear and potentially unnecessary, and the lack of detailed information increases operational uncertainty. \\
\bottomrule
\end{tabular}
\caption{Representative examples of accepted (\textcolor{green!60!black}{\ding{51}}) and rejected (\textcolor{red!70!black}{\ding{55}}) scenarios with detailed rationale illustrating our quality threshold criteria.}
\label{tab:acceptance_examples}
\end{table*}

\paragraph{Inter-Annotator Agreement.} We assessed annotation reliability by measuring pairwise agreement using Cohen's $\kappa$ across all 1,804 scenarios evaluated in Stage 2. The results demonstrate strong overall agreement: $\kappa = 0.84$ (95\% CI: [0.82, 0.86], computed via bootstrap resampling with 10,000 iterations). Agreement varied by criterion: Instruction Clarity achieved the highest agreement ($\kappa = 0.88$), followed by Physical Feasibility ($\kappa = 0.82$) and Solvability ($\kappa = 0.79$). The relatively lower agreement on Solvability reflects the inherent cognitive difficulty of determining solution existence through mental simulation alone, which validates our design decision to include simulation-based verification in Stage 3. Disagreements between annotators (204 cases, 11.3\% of scenarios) were systematically adjudicated by a senior researcher through interactive simulation-based verification, with final decisions documented for quality tracking.

\paragraph{Acceptance Results.} Human verification accepted 1,518 of 1,804 candidate scenarios (84.1\%), demonstrating that automated validation alone is insufficient for ensuring benchmark quality—15.9\% of automatically validated scenarios contained subtle issues detectable only through human expertise. Analysis of rejection rationales revealed the following distribution: ambiguous or underspecified instructions (37.2\%), unsolvable tasks due to missing preconditions or unreachable goals (31.6\%), physical inconsistencies between object properties and task requirements (22.4\%), and other issues including edge cases and annotation uncertainties (8.8\%).

Acceptance rates varied systematically by task complexity, as shown in Table~\ref{tab:acceptance_by_type}. The observed inverse relationship between task complexity and acceptance rate provides indirect validation of our hierarchical task taxonomy: more complex tasks involving multiple reasoning dimensions naturally present greater opportunities for subtle generation errors. This pattern also confirms the importance of human verification, particularly for advanced task categories where automated checks are less effective.

\begin{table}[h]
\centering
\small
\begin{tabular}{lc}
\toprule
\textbf{Task Category} & \textbf{Acceptance Rate} \\
\midrule
Direct Command (L0) & 91.2\% \\
Tool Use (L1) & 85.7\% \\
Attribute Reasoning (L1) & 82.3\% \\
Compound Reasoning (L2) & 78.9\% \\
Explicit Collaboration (L0) & 84.1\% \\
Implicit Collaboration (L1) & 79.6\% \\
Compound Collaboration (L2) & 76.4\% \\
\bottomrule
\end{tabular}
\caption{Human verification acceptance rates by task category, showing systematic variation with task complexity.}
\label{tab:acceptance_by_type}
\end{table}

\subsubsection{Stage 3: Expert Trajectory Validation}

The final validation stage ensures that all accepted scenarios are not only well-formed but genuinely solvable through executable action sequences. Oracle agents with complete environmental knowledge generated expert demonstration trajectories for all 1,518 human-verified scenarios.

\paragraph{Trajectory Generation and Verification.} Each generated trajectory underwent rigorous simulation replay to verify three critical properties: (1) precondition satisfaction at each action step, ensuring no invalid state transitions; (2) goal state achievement upon trajectory completion, confirming task success; and (3) physical constraint compliance throughout execution, validating that all intermediate states respect environmental limitations. Trajectories failing any verification criterion were regenerated using alternative planning strategies, with a maximum of three attempts per scenario. Scenarios with persistent trajectory failures were removed from the final dataset (18 of 1{,}518, 1.2\%; yielding the final 1{,}500 scenarios), as such failures indicate potential issues not detected in earlier stages.

\paragraph{Oracle Quality Validation.} To verify that oracle trajectories represent high-quality reference solutions rather than merely valid but suboptimal ones, we conducted an additional validation study on a randomly sampled subset of 100 scenarios stratified across task categories. Two human experts with extensive experience in embodied AI independently solved each scenario without access to oracle solutions, and we compared their solutions with the corresponding oracle trajectories using step count as the primary efficiency metric. Oracle solutions matched or outperformed human expert solutions in 94\% of evaluated cases, with the remaining 6\% representing alternative but equally valid strategies (typically involving different but equivalent tool choices or spatial paths). This validation confirms that oracle trajectories provide appropriate reference solutions for both evaluation and potential use in supervised learning approaches.

\paragraph{Final Dataset Statistics.} The validated dataset comprises 1,500 scenarios instantiated into 16,592 task instances (multiple task variations per scenario). Expert trajectory statistics: mean length 9.1 steps (SD 4.2), median 9.0 steps, range [3, 25] steps. The high trajectory generation success rate (98.8\% of human-verified scenarios, averaging 1.3 generation attempts per scenario) provides empirical validation that our human verification process effectively identifies genuinely solvable scenarios.

\paragraph{Ethics Statement.} All human annotators provided written informed consent prior to participation and were compensated at standard institutional rates (\$20 USD per hour). The annotation task involved evaluation of synthetic embodied AI scenarios and did not involve deceptive practices, collection of personal information, or exposure to sensitive or harmful content. This study protocol was reviewed and approved by our institutional review board (IRB) prior to data collection.

\paragraph{Summary.} Our three-stage validation pipeline systematically addresses the quality challenges inherent in LLM-based benchmark generation. Automated validation handles 87.6\% of systematic errors efficiently, human verification with trained annotators (Cohen's $\kappa = 0.84$) catches subtle semantic and physical issues, and simulation-based trajectory validation confirms genuine solvability. The resulting benchmark of 1,500 validated scenarios provides a reliable foundation for evaluating embodied reasoning capabilities.

\subsection{Benchmark Statistics and Coverage}

\label{sec:dataset_statistics}

\benchmark encompasses 1,500 scenarios with 64,057 interactive objects, providing comprehensive coverage across diverse domains and task complexities. Tables~\ref{tab:dataset_overview} through \ref{tab:domain_distribution} present detailed statistics demonstrating the scale and diversity of our benchmark.

\begin{table}[t]
\centering
\small
\begin{tabular}{lr}
\toprule
\textbf{Metric} & \textbf{Count} \\
\midrule
Total Scenarios & 1,500 \\
Total Task Files & 1,500 \\
Total Task Instances & 16,592 \\
Interactive Objects & 64,057 \\
Spatial Nodes (Rooms) & 6,634 \\
Average Objects per Scene & 42.7 \\
Average Rooms per Scene & 4.4 \\
Collaborative Agent Pairs & 1,481 \\
\bottomrule
\end{tabular}
\caption{Dataset Overview and Scale}
\label{tab:dataset_overview}
\end{table}

\begin{table}[t]
\centering
\small
\begin{tabular}{lrr}
\toprule
\textbf{Task Category} & \textbf{Count} & \textbf{Percentage} \\
\midrule
\multicolumn{3}{l}{\textit{Single-Agent Tasks (65\%)}} \\
Direct Command & 2,684 & 16.2\% \\
Attribute Reasoning & 2,669 & 16.1\% \\
Tool Use & 2,190 & 13.2\% \\
Compound Reasoning & 2,214 & 13.3\% \\
\midrule
\multicolumn{3}{l}{\textit{Multi-Agent Tasks (35\%)}} \\
Explicit Collaboration & 2,160 & 13.0\% \\
Implicit Collaboration & 2,582 & 15.6\% \\
Compound Collaboration & 2,093 & 12.6\% \\
\midrule
\textbf{Total} & \textbf{16,592} & \textbf{100\%} \\
\bottomrule
\end{tabular}
\caption{Hierarchical Task Distribution}
\label{tab:task_distribution}
\end{table}

\paragraph{Physical Property Modeling.}
The benchmark features exceptional attribute diversity with 6,381 distinct property types. Core physical properties are comprehensively modeled: weight (64,047 objects), material composition (35,411 objects), size dimensions (22,820 objects), color (28,034 objects), and dynamic states (17,547 objects). This rich attribute space enables sophisticated reasoning about physical constraints and object affordances.

\begin{table}[t]
\centering
\small
\begin{tabular}{lrr}
\toprule
\textbf{Category/Material} & \textbf{Count} & \textbf{Percentage} \\
\midrule
\multicolumn{3}{l}{\textit{Object Categories}} \\
Container & 17,632 & 27.5\% \\
Tool & 15,134 & 23.6\% \\
Appliance & 8,963 & 14.0\% \\
Furniture & 6,234 & 9.7\% \\
Consumable & 4,890 & 7.6\% \\
Others & 11,204 & 17.6\% \\
\midrule
\multicolumn{3}{l}{\textit{Material Types (Top 10 of 1,123)}} \\
Plastic & 13,767 & 21.5\% \\
Metal & 11,274 & 17.6\% \\
Wood & 8,263 & 12.9\% \\
Glass & 6,277 & 9.8\% \\
Fabric & 5,060 & 7.9\% \\
Ceramic & 3,843 & 6.0\% \\
Silicon & 1,794 & 2.8\% \\
Aluminum & 1,601 & 2.5\% \\
Steel & 1,153 & 1.8\% \\
Others & 11,025 & 17.2\% \\
\bottomrule
\end{tabular}
\caption{Object Categories and Material Distribution}
\label{tab:object_materials}
\end{table}

\paragraph{Action Space and Tool Ecosystem.}
The framework supports 214 distinct action types, partitioned into basic actions (60\%) available to all agents and tool-dependent actions (40\%) requiring specific capabilities. Among the 64,057 objects, 15,134 are classified as tools (23.6\%), with 13,482 objects possessing the \texttt{provides\_abilities} attribute that enables dynamic capability extension. This design enables realistic modeling of how agents acquire new abilities through tool use.

\begin{table}[t]
\centering
\small
\begin{tabular}{lrr}
\toprule
\textbf{Domain/Room Type} & \textbf{Count} & \textbf{Percentage} \\
\midrule
\multicolumn{3}{l}{\textit{Application Domains}} \\
Laboratory & 585 & 39.0\% \\
Office & 282 & 18.8\% \\
Industrial & 173 & 11.5\% \\
Medical & 93 & 6.2\% \\
Household & 93 & 6.2\% \\
Educational & 63 & 4.2\% \\
Retail & 48 & 3.2\% \\
Service & 30 & 2.0\% \\
Entertainment & 27 & 1.8\% \\
Transportation & 23 & 1.5\% \\
Others & 83 & 5.6\% \\
\midrule
\multicolumn{3}{l}{\textit{Room Types (Top 5)}} \\
Laboratory & 1,876 & 28.3\% \\
Storage & 1,234 & 18.6\% \\
Workspace & 987 & 14.9\% \\
Office & 765 & 11.5\% \\
Workshop & 543 & 8.2\% \\
\bottomrule
\end{tabular}
\caption{Domain and Spatial Distribution}
\label{tab:domain_distribution}
\end{table}

\paragraph{Cross-Domain Coverage.}
The benchmark spans diverse application domains, with laboratory environments comprising 39.0\% of scenarios, followed by office (18.8\%), industrial (11.5\%), and medical (6.2\%) settings. This distribution reflects our emphasis on professional environments where embodied reasoning is particularly critical. Each domain presents unique challenges: laboratory settings require precise tool usage and material handling, office environments emphasize multi-agent coordination, and industrial scenarios demand reasoning about heavy equipment and safety constraints.

\paragraph{Quality Assurance and Expert Trajectories.}
All 16,592 task instances include expert demonstration trajectories averaging 8.7 steps, providing optimal solutions for comparison and learning. Each trajectory undergoes validation to ensure physical feasibility and task completion. The evaluation framework supports multi-level verification including spatial relationships (1,300 location checks), state transitions (open/closed, on/off states), and compound conditions for complex task assessment. This comprehensive validation ensures that all tasks are both challenging and solvable, maintaining benchmark integrity while achieving unprecedented scale.

\subsection{Failure Mode Analysis}
\label{sec:failure_mode_analysis}

Systematic failure analysis reveals task-specific performance bottlenecks that vary distinctly across model scales. Tool Use failures are dominated by exploration deficits (31.2\%), where models fail to locate required tools while maintaining spatial representations. Models below 7B parameters exhibit 2.7-fold higher failure rates (84.2\% vs. 31.2\%), confirming critical scale thresholds for embodied reasoning. Compound Reasoning failures stem primarily from planning degradation (28.7\%), with models losing track of intermediate subgoals during execution.

Implicit Collaboration shows distinct timing failures (35.8\%)---models either initiate collaboration prematurely or miss coordination opportunities. This failure mode exhibits no scale correlation, indicating that collaboration timing demands reasoning mechanisms absent from current architectures. These failure patterns demonstrate that task categories stress fundamentally different cognitive capabilities, necessitating targeted architectural solutions beyond universal parameter scaling.

\subsection{Hyperparameters}

\label{sec:hyperparameters}

\paragraph{Supervised Fine-Tuning.}
The 1{,}942 source scenarios used for SFT distillation are drawn from a held-out partition disjoint from the 2{,}800 evaluation scenarios; no scenario, object instance, or task instance appears in both sets. We verify disjointness by scenario ID at preprocessing time. We performed full-parameter supervised fine-tuning on the \texttt{Qwen2.5-3B-Instruct} model using LLaMA-Factory~\citep{zheng2024llamafactoryunifiedefficientfinetuning} to adapt it to our dataset. The training was conducted on 4x NVIDIA A100 GPUs. The effective batch size was 64, achieved through a per-device batch size of 1 and 16 gradient accumulation steps across 4 devices. Key hyperparameters for the SFT stage are summarized in Table~\ref{tab:sft_hyperparameters}.

\begin{table}[!h]
\centering
\small
\begin{tabular}{@{}ll@{}}
\toprule
\textbf{Hyperparameter} & \textbf{Value} \\
\midrule
Base Model & \texttt{Qwen2.5-3B-Instruct} \\
Fine-tuning Method & Full-parameter \\
Effective Batch Size & 64 \\
Learning Rate & 1.0e-5 \\
LR Scheduler & Cosine Decay \\
Warmup Ratio & 0.1 \\
Training Epochs & 3 \\
Max Sequence Length & 15,360 \\
Precision & BF16 \\
\bottomrule
\end{tabular}
\caption{Hyperparameters for Supervised Fine-Tuning.}
\label{tab:sft_hyperparameters}
\end{table}

\paragraph{Model Inference.}
To ensure a fair and consistent comparison, all models were evaluated using the same set of inference parameters. We utilized the vLLM engine for efficient serving, with a tensor parallel size of 4. The decoding strategy was configured to balance response quality and exploration in complex reasoning tasks. The inference settings are detailed in Table~\ref{tab:inference_hyperparameters}.

\begin{table}[!h]
\centering
\small
\begin{tabular}{@{}ll@{}}
\toprule
\textbf{Hyperparameter} & \textbf{Value} \\
\midrule
Inference Engine & vLLM \\
Tensor Parallel Size & 4 \\
Decoding Strategy & Nucleus Sampling \\
Temperature & 0.3 \\
Top-p & 1.0 (Default) \\
Max Generation Tokens & 4096 \\
Max Model Length & 15,360 \\
\bottomrule
\end{tabular}
\caption{Hyperparameters for Model Inference.}
\label{tab:inference_hyperparameters}
\end{table}

\input{sections/7_discussion}
\input{sections/9_case_study_appendix}
\input{sections/8_prompts_appendix}
\input{sections/10_data_pipeline_case}

%% file: pics/appendix_scenes/appendix_snippet_batch2.tex
\subsection{Additional Scene Cases}
\label{appendix:scene-cases-extra}

\begin{figure*}[!t]
  \centering
  \begin{subfigure}[t]{0.32\textwidth}
    \includegraphics[width=\linewidth]{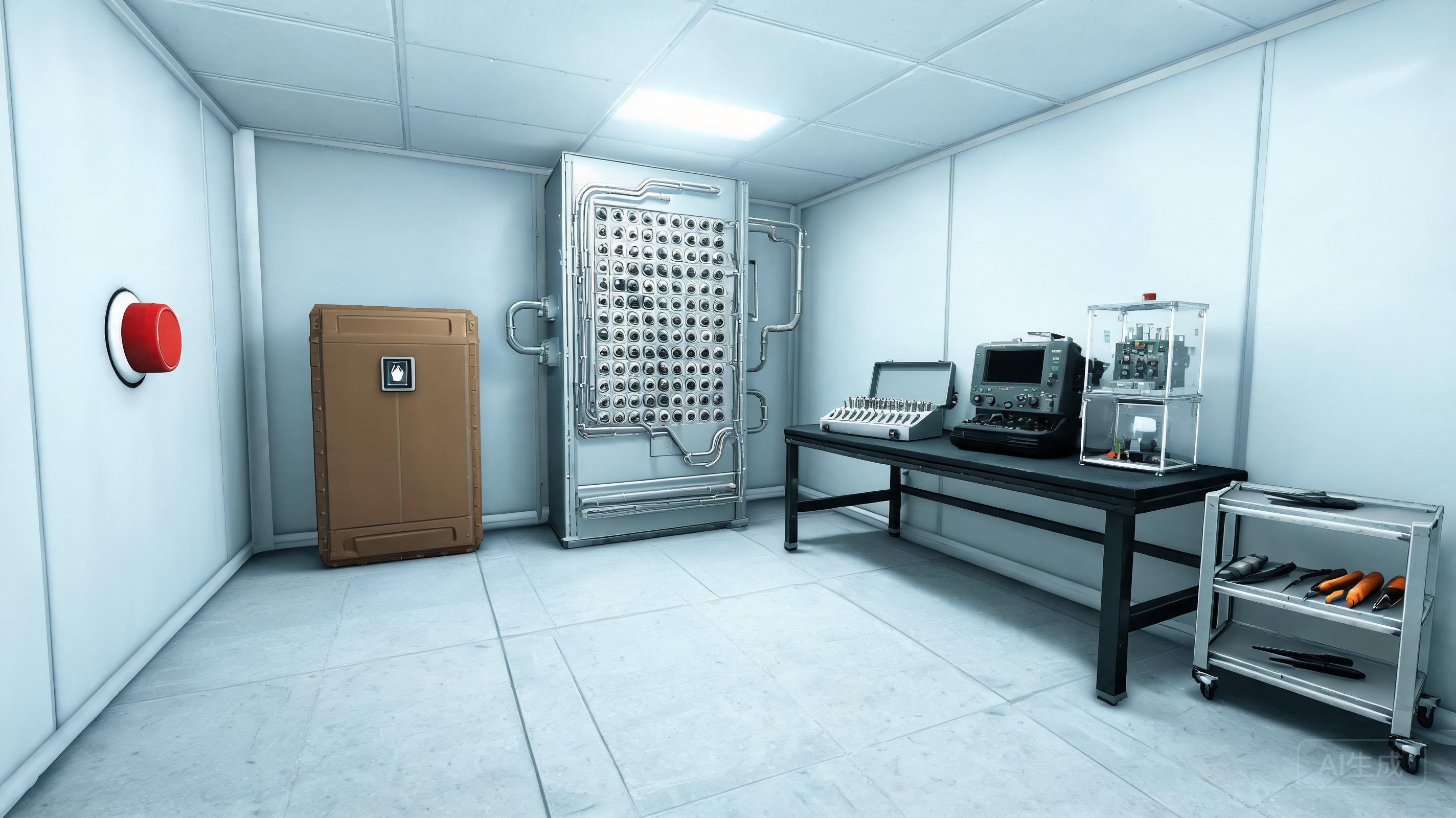}
    \caption{\texttt{single/00001}: quantum calibration lab (\textit{direct\_command}).}
  \end{subfigure}\hfill
  \begin{subfigure}[t]{0.32\textwidth}
    \includegraphics[width=\linewidth]{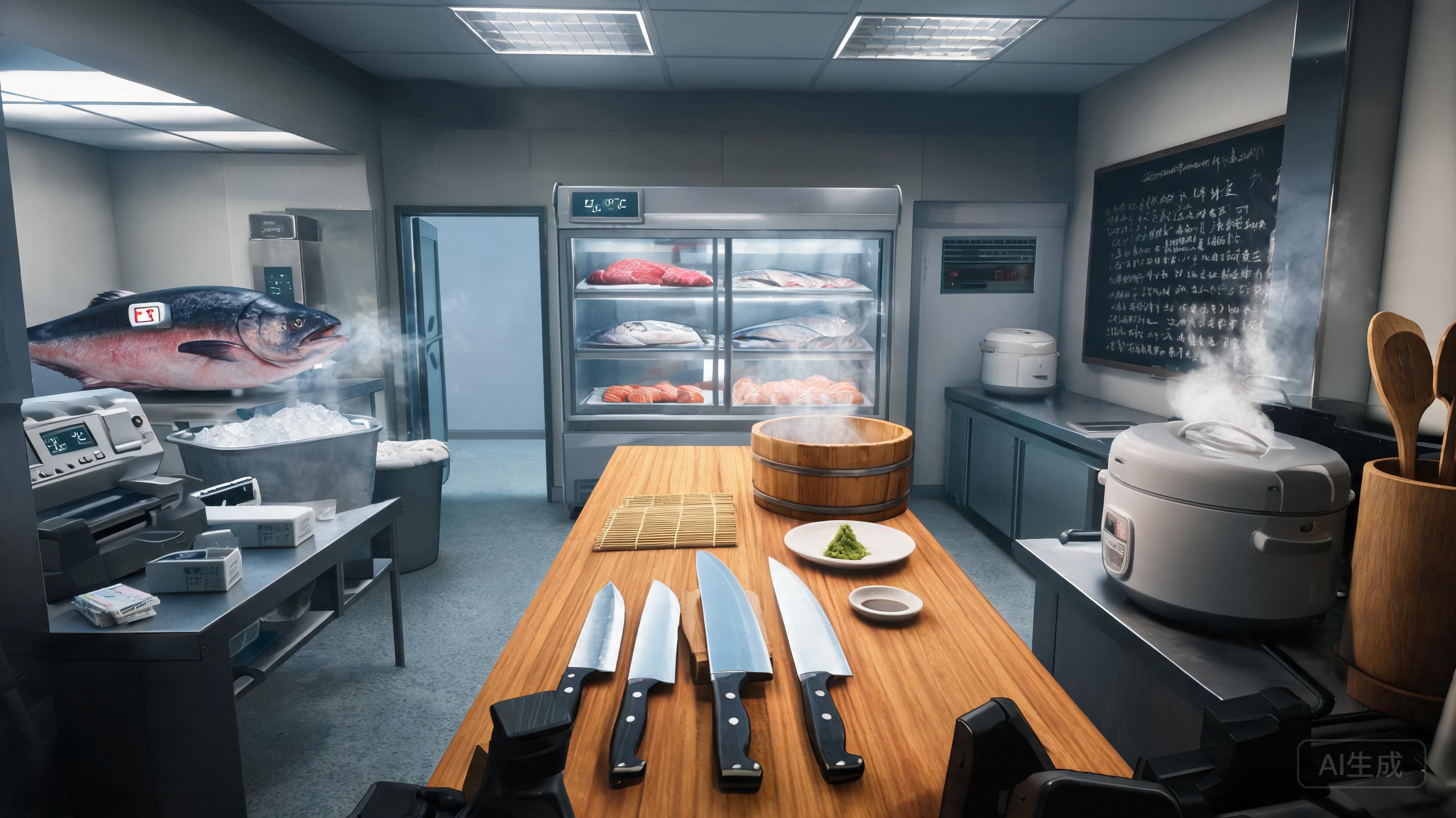}
    \caption{\texttt{single/00370}: sushi kitchen (\textit{direct\_command}).}
  \end{subfigure}\hfill
  \begin{subfigure}[t]{0.32\textwidth}
    \includegraphics[width=\linewidth]{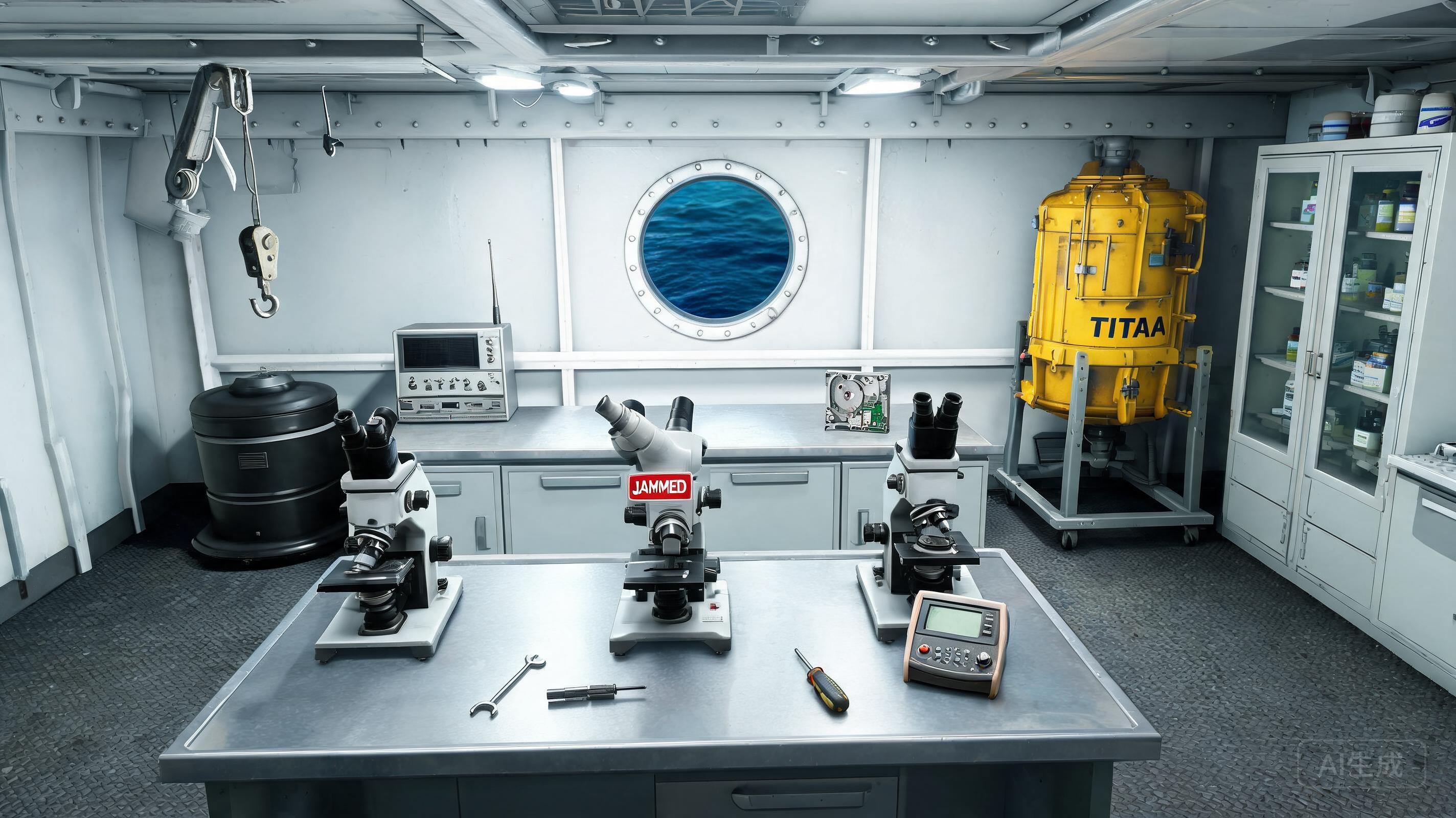}
    \caption{\texttt{single/00350}: research vessel main lab (\textit{tool\_use}).}
  \end{subfigure}

  \vspace{0.6em}
  \begin{subfigure}[t]{0.32\textwidth}
    \includegraphics[width=\linewidth]{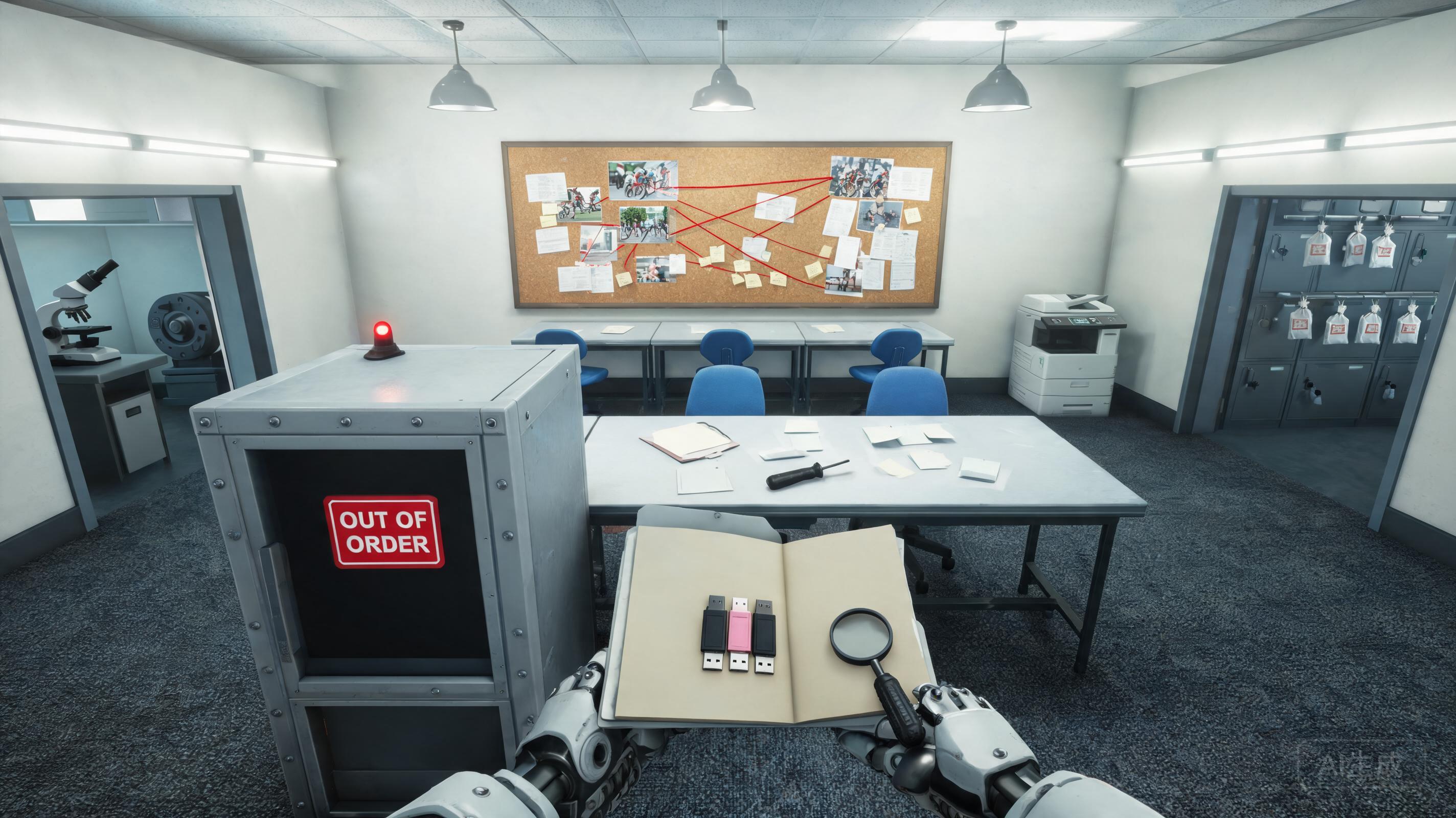}
    \caption{\texttt{single/00007}: anti-doping forensics (\textit{tool\_use}).}
  \end{subfigure}\hfill
  \begin{subfigure}[t]{0.32\textwidth}
    \includegraphics[width=\linewidth]{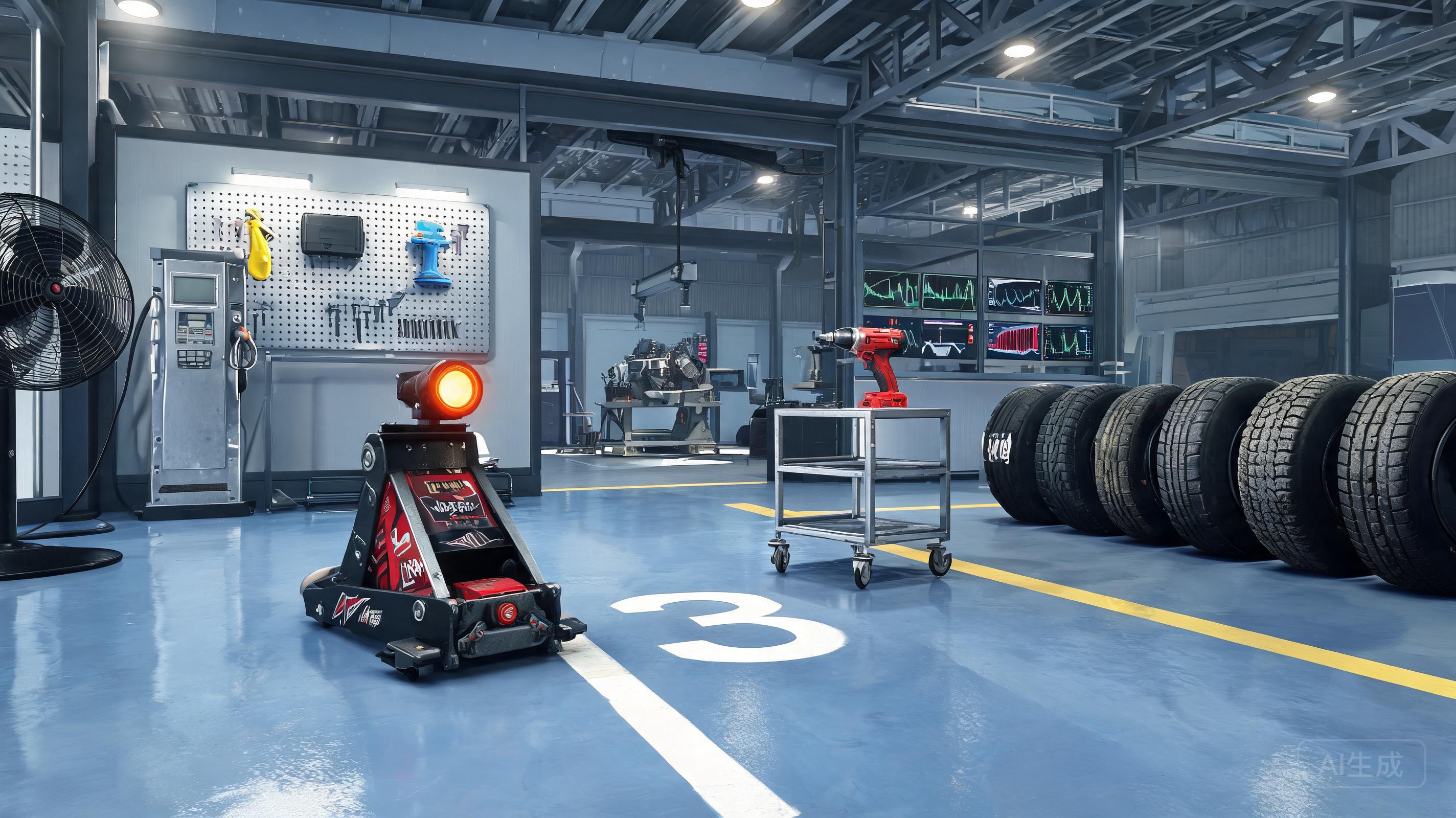}
    \caption{\texttt{single/00035}: NASCAR pit garage (\textit{tool\_use}).}
  \end{subfigure}\hfill
  \begin{subfigure}[t]{0.32\textwidth}
    \includegraphics[width=\linewidth]{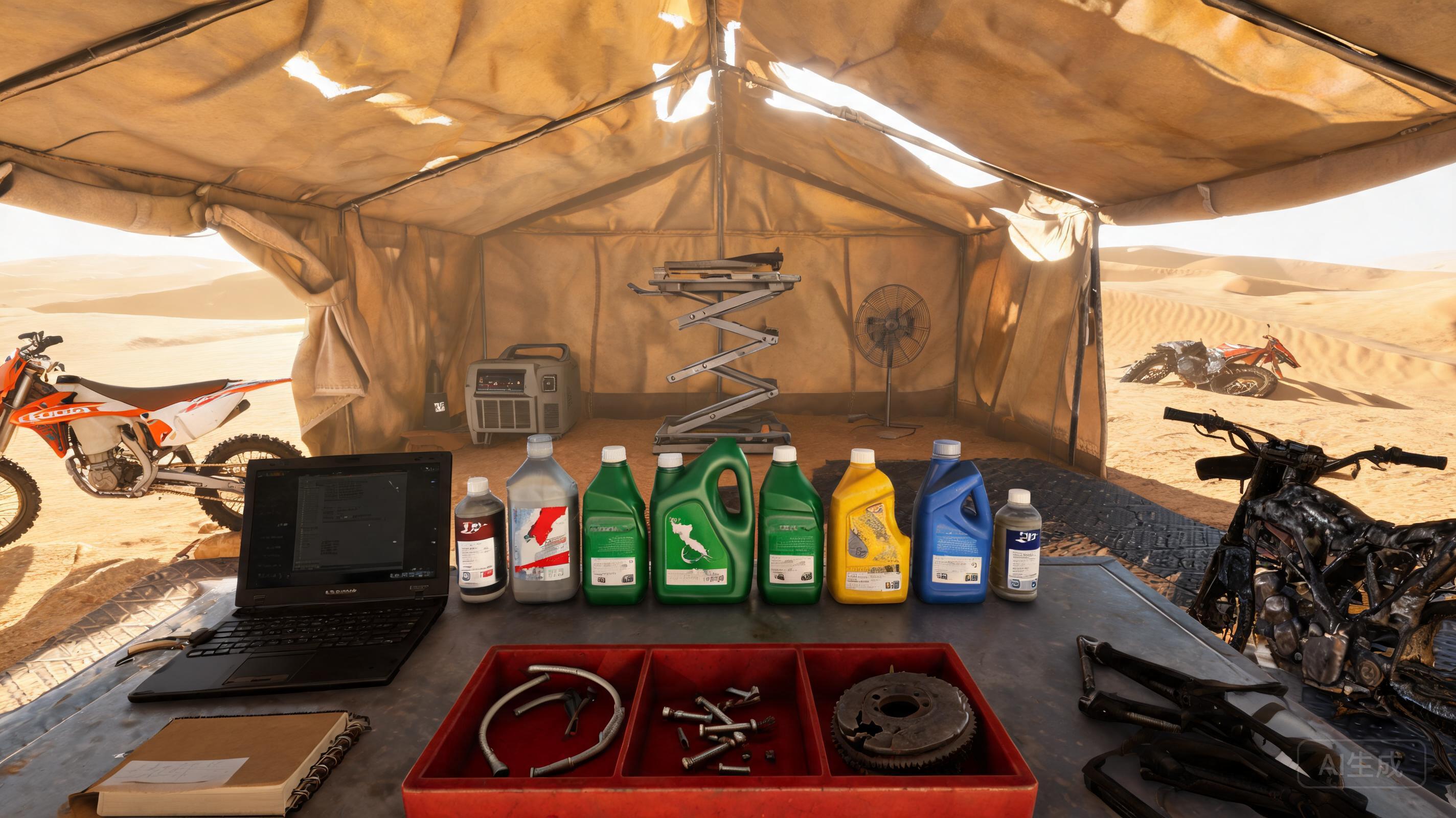}
    \caption{\texttt{single/00038}: Dakar rally bivouac (\textit{attribute\_reasoning}).}
  \end{subfigure}

  \vspace{0.6em}
  \begin{subfigure}[t]{0.32\textwidth}
    \includegraphics[width=\linewidth]{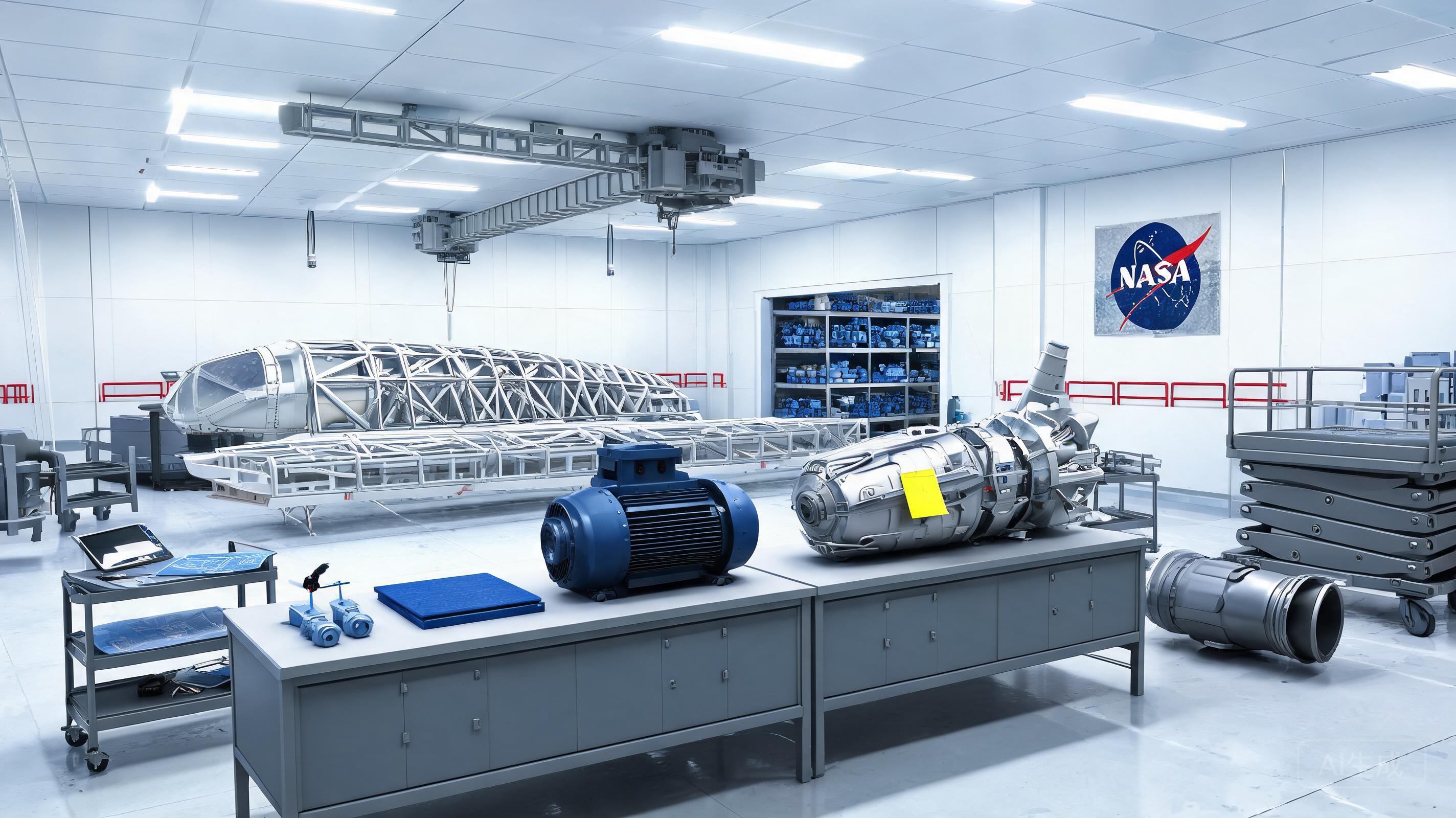}
    \caption{\texttt{single/00094}: aerospace assembly bay (\textit{attribute\_reasoning}).}
  \end{subfigure}\hfill
  \begin{subfigure}[t]{0.32\textwidth}
    \includegraphics[width=\linewidth]{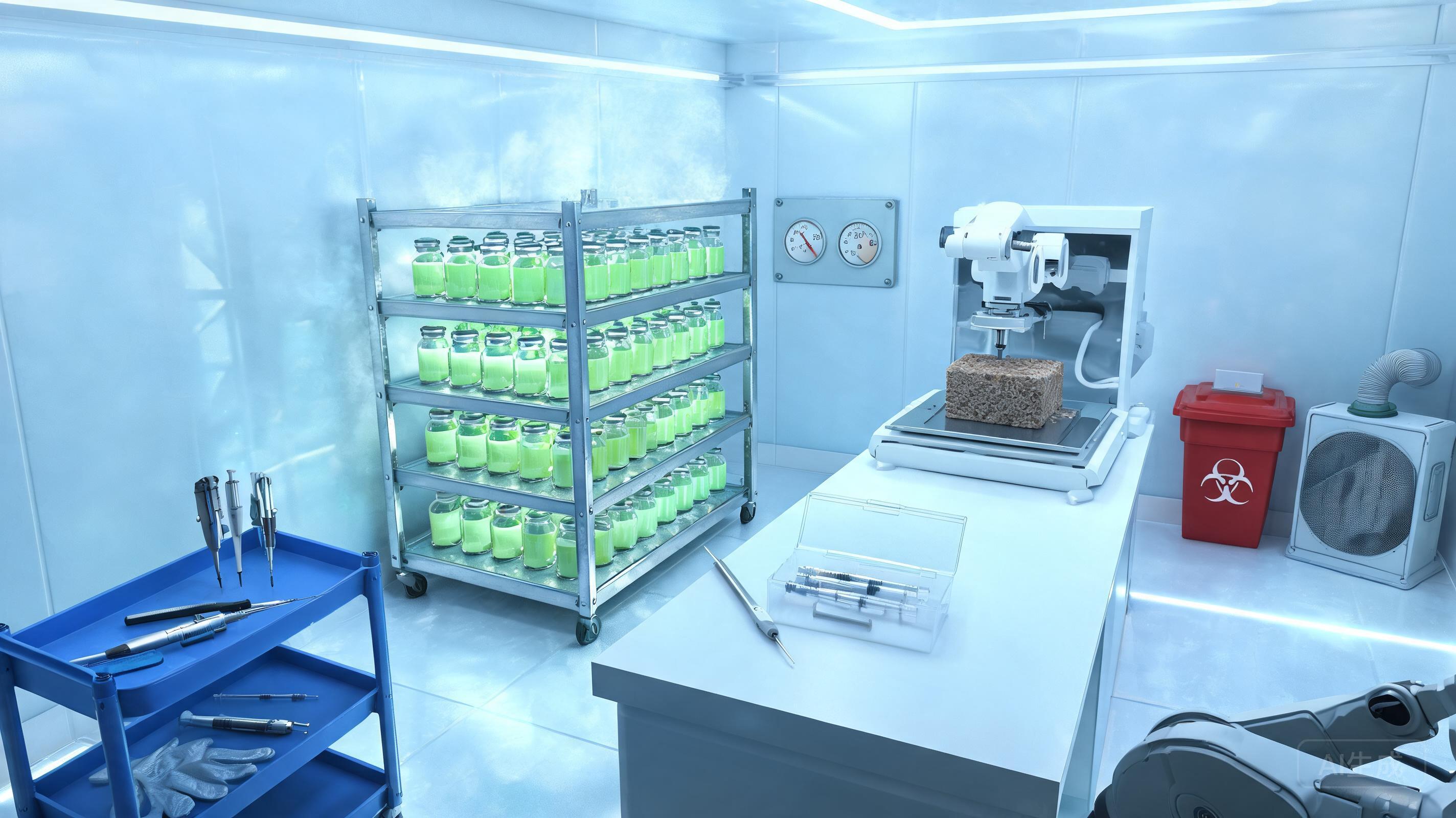}
    \caption{\texttt{single/00153}: living architecture bio-chamber (\textit{compound\_reasoning}).}
  \end{subfigure}\hfill
  \begin{subfigure}[t]{0.32\textwidth}
    \includegraphics[width=\linewidth]{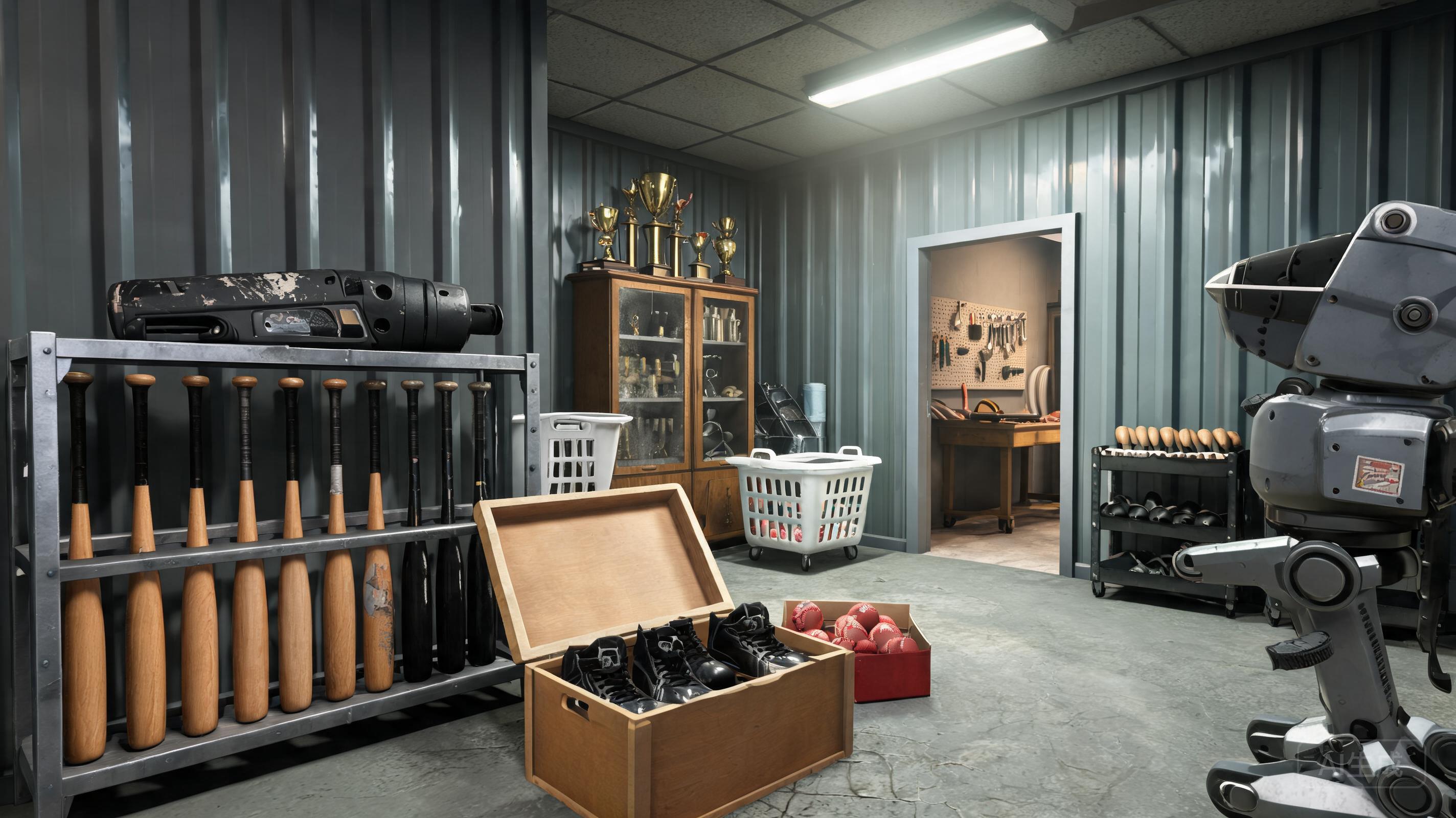}
    \caption{\texttt{multi/00117}: baseball equipment storage (\textit{explicit\_collaboration}).}
  \end{subfigure}

  \vspace{0.6em}
  \begin{subfigure}[t]{0.32\textwidth}
    \includegraphics[width=\linewidth]{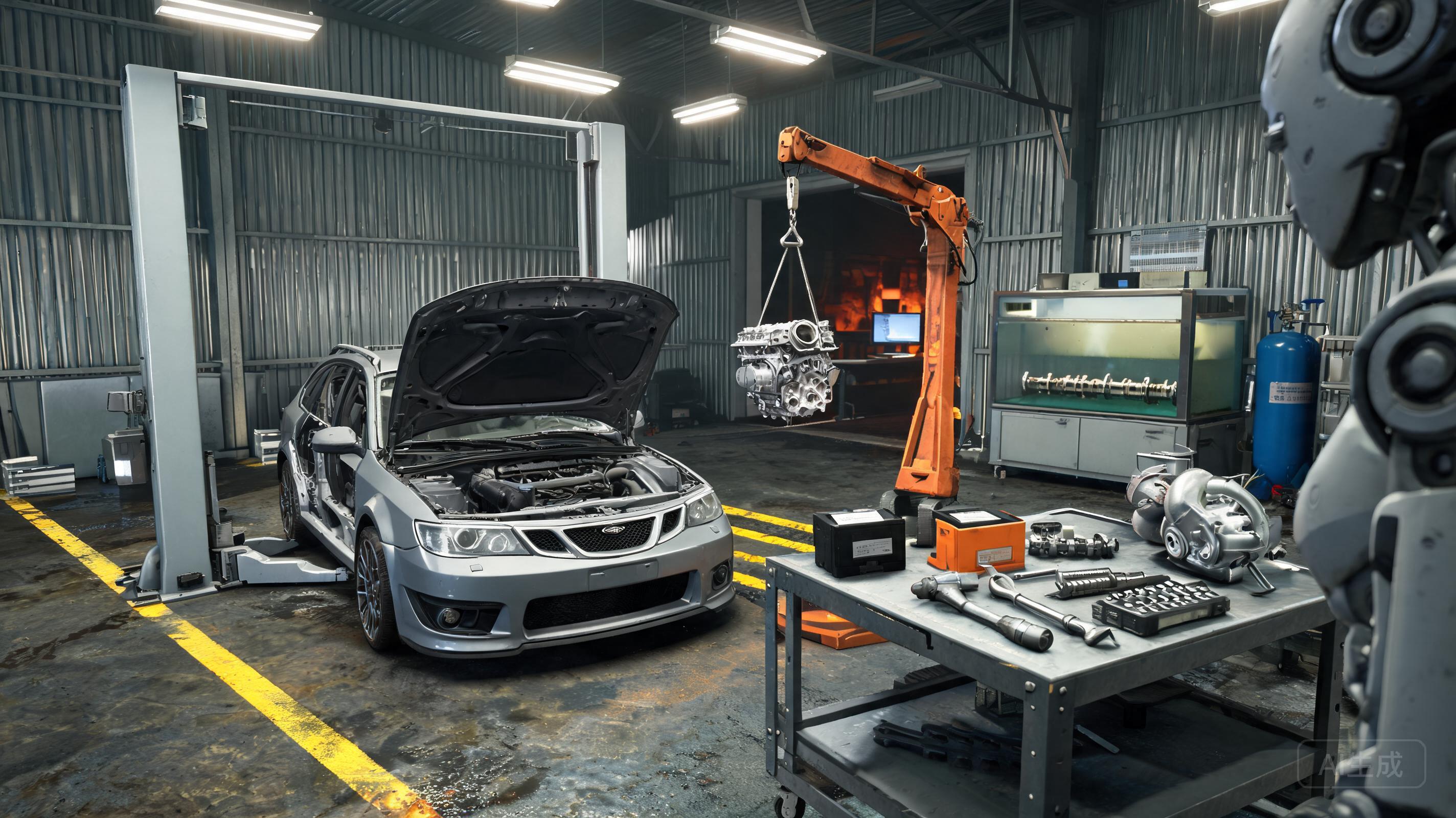}
    \caption{\texttt{multi/00138}: automotive prototype workshop (\textit{implicit\_collaboration}).}
  \end{subfigure}\hfill
  \begin{subfigure}[t]{0.32\textwidth}
    \includegraphics[width=\linewidth]{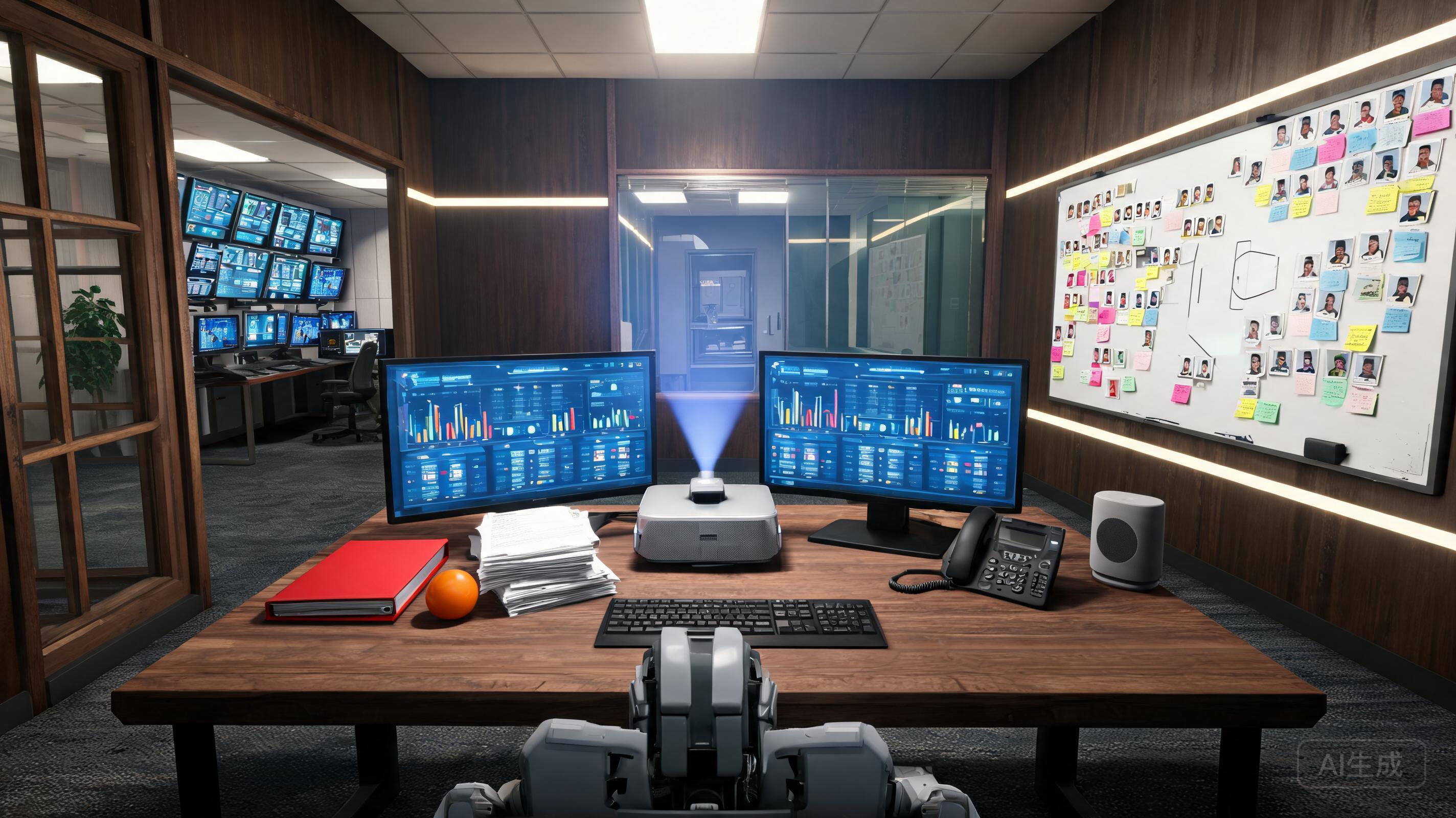}
    \caption{\texttt{multi/00259}: NBA trade-deadline war room (\textit{compound\_collaboration}).}
  \end{subfigure}\hfill
  \begin{subfigure}[t]{0.32\textwidth}
    \includegraphics[width=\linewidth]{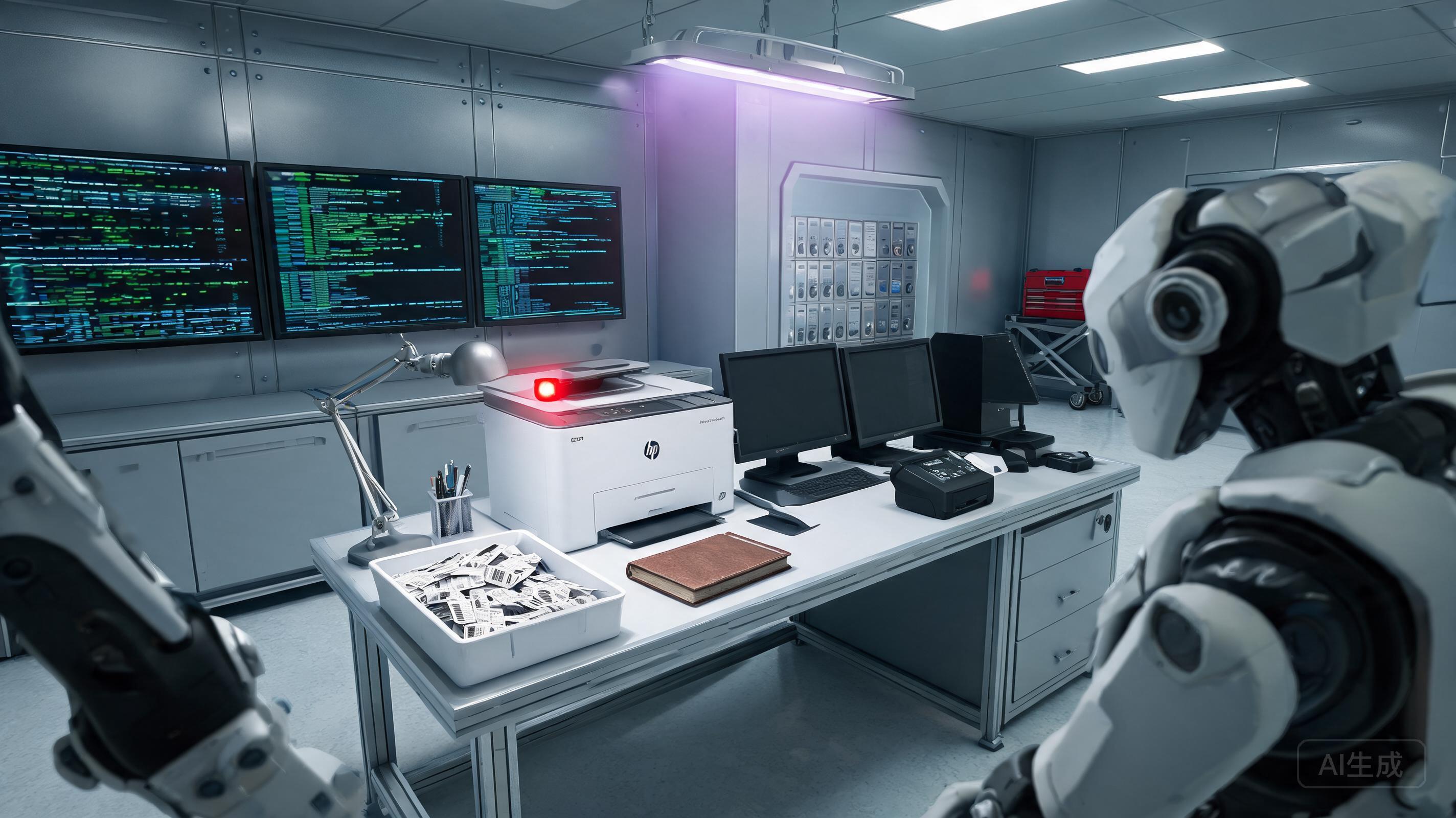}
    \caption{\texttt{multi/00598}: data verification lab (\textit{compound\_collaboration}).}
  \end{subfigure}

  \caption{Representative scene cases retained for visualization. Renderings are produced from the scene JSON descriptions with Doubao-Seedream-5.0-lite~\citep{gu2026seed3d20advancinghighfidelity} and are illustrative only.}
  \label{fig:appendix-scene-cases-extra}
\end{figure*}

Figure~\ref{fig:appendix-scene-cases-extra} visualizes twelve representative scenes retained after curation, covering all seven task categories. These renderings are illustrative only and are not used during evaluation; the underlying scene state consumed by agents is the structured JSON description.

%% file: sections/7_discussion.tex
\subsection{Discussion}
\label{sec:discussion}

\paragraph{Action Grounding as a Distinct Capability.}
Our results suggest that action grounding is not reducible to a single
reasoning skill. The L1 categories expose different bottlenecks:
Attribute Reasoning is nearly uncorrelated with Tool Use and Implicit
Collaboration ($r\in[-0.03,0.02]$; Figure~\ref{fig:l1_correlation}),
whereas Tool Use and Implicit Collaboration respond oppositely to the
same observability intervention (Figure~\ref{fig:world_graph}). Complete
environment graphs help Tool Use, indicating a search-bound bottleneck,
but hurt Implicit Collaboration, indicating a constraint-filtering
bottleneck. Supervised fine-tuning further separates execution from
collaboration triggering: Qwen2.5-3B improves from $0.6\%$ to $76.3\%$
on Direct Command but only from $1.5\%$ to $5.5\%$ on Implicit
Collaboration (Table~\ref{tab:comprehensive_performance}). Finally,
reasoning-oriented models do not automatically solve continuous-attribute
grounding: DeepSeek-R1 reaches only $41.9\%$ on Attribute Reasoning
despite a $26\times$ larger token budget per trajectory
(Figure~\ref{fig:scaling}). These asymmetries suggest that action
grounding requires bottleneck-specific mechanisms for attribute
comparison, tool search, state filtering, and capacity-boundary
detection, rather than longer reasoning traces or uniform scaling alone.

\paragraph{Implications for Model Architecture.}
Unlike abstract reasoning tasks where all provided information typically bears relevance, embodied scenarios require selective attention over spatially and temporally distributed constraint sets.

\paragraph{Future Research Directions.}
Several promising directions emerge from our findings. First, developing attention mechanisms specifically designed for filtering task-relevant physical constraints from environmental context could improve performance on implicit collaboration tasks. Second, exploring hybrid symbolic-neural architectures that explicitly reason about physical laws while maintaining learned flexibility represents a promising approach~\citep{rabinowitz2018machine}. Third, investigating training methods that better transfer single-agent reasoning capabilities to multi-agent scenarios remains an important open challenge: prior work~\citep{wang2026milestoneguidedpolicylearninglonghorizon}, together with related efforts~\citep{feng2025groupingrouppolicyoptimizationllm}, points to possible avenues. Finally, extending evaluation frameworks to incorporate continuous control and sensorimotor feedback would provide a more complete understanding of embodied reasoning requirements, building on prior work~\citep{li2025viewspatialbenchevaluatingmultiperspectivespatial, li2026spatialevoselfevolvingspatialintelligence}.

%% file: sections/9_case_study_appendix.tex
\onecolumn

\subsection{Case Studies}
\label{sec:case_study}

Two qualitative cases. Each pairs a strong model with Qwen2.5-7B on \emph{the same} scenario; all steps are reproduced verbatim from the logged \texttt{action\_sequence}. Cooperative actions issued jointly by both agents are abbreviated as \texttt{(a1,a2): CORP\_X \dots}.

\subsubsection{Case A: Implicit Collaboration}
\label{sec:case_a}

\begin{tcolorbox}[
    width=\linewidth,
    colback=gray!8!white,
    colframe=black!75,
    boxrule=0.4pt,
    arc=1pt,
    left=3mm, right=3mm, top=1.5mm, bottom=1.5mm,
    fontupper=\small,
    before skip=4pt, after skip=4pt
]
{\raggedright
\textbf{Task.} ``Move the \texttt{genome\_sequencing\_array\_1} to the \texttt{cold\_storage\_vault}.''\par}
\smallskip
\begin{center}
\renewcommand{\arraystretch}{1.1}
\begin{tabular}{@{}lr@{}}
\toprule
Object / agent & Property \\
\midrule
\texttt{genome\_sequencing\_array\_1} (target) & weight $= 100$\,kg \\
\texttt{agent\_1}, \texttt{agent\_2}            & max load $= 9$\,kg each \\
\midrule
Single-agent feasible? & \textcolor{red!70!black}{\textbf{No}} ($100 > 9$) \\
\texttt{CORP\_GRAB} required? & \textcolor{green!55!black}{\textbf{Yes}} \\
\bottomrule
\end{tabular}
\end{center}
\end{tcolorbox}

\begin{tcolorbox}[
    width=\linewidth,
    colback=green!4!white,
    colframe=green!45!black,
    boxrule=0.6pt,
    arc=1pt,
    title={\faCheckCircle~\textbf{Claude-Sonnet-4.6} \textnormal{--- task completed in 9 steps}},
    fonttitle=\small,
    coltitle=white,
    colbacktitle=green!45!black,
    left=3mm, right=3mm, top=1.5mm, bottom=1.5mm,
    fontupper=\footnotesize\ttfamily,
    before skip=4pt, after skip=4pt,
    breakable
]
\setlength{\parindent}{0pt}\setlength{\parskip}{1.5pt}
01.~a1: EXPLORE\quad\textbar\quad a2: GOTO main\_lab\_floor \hfill [SUCCESS]\\
02.~a1: GOTO data\_hub\quad\textbar\quad a2: EXPLORE \hfill [SUCCESS]\\
03.~(a1,a2): GOTO genome\_sequencing\_array\_1 \hfill [INVALID]\\
\hspace*{1.4em}{\normalfont\itshape env:}~``Must go to Main Lab Floor first, then approach Genome Sequencing Array.''\\
04.~a1: GOTO main\_lab\_floor\quad\textbar\quad a2: GOTO genome\_sequencing\_array\_1 \hfill [SUCCESS]\\
05.~(a1,a2): GOTO genome\_sequencing\_array\_1 \hfill [SUCCESS]\\
06.~\textbf{(a1,a2): CORP\_GRAB genome\_sequencing\_array\_1} \hfill [SUCCESS]\\
07.~\textbf{(a1,a2): CORP\_GOTO cold\_storage\_vault} \hfill [SUCCESS]\\
08.~\textbf{(a1,a2): CORP\_PLACE genome\_sequencing\_array\_1 in cold\_storage\_vault} \hfill [SUCCESS]\\
09.~(a1,a2): DONE\quad{\normalfont(validation: 1/1 task completed)} \hfill [SUCCESS]
\end{tcolorbox}

\begin{tcolorbox}[
    width=\linewidth,
    colback=red!4!white,
    colframe=red!55!black,
    boxrule=0.6pt,
    arc=1pt,
    title={\faTimesCircle~\textbf{Qwen2.5-7B} \textnormal{--- task failed; 19 steps total, non-critical turns collapsed}},
    fonttitle=\small,
    coltitle=white,
    colbacktitle=red!55!black,
    left=3mm, right=3mm, top=1.5mm, bottom=1.5mm,
    fontupper=\footnotesize\ttfamily,
    before skip=4pt, after skip=4pt,
    breakable
]
\setlength{\parindent}{0pt}\setlength{\parskip}{1.5pt}
01.~(a1,a2): EXPLORE \hfill [SUCCESS]\\
02.~a1: GRAB lubricant\_1\quad\textbar\quad a2: GOTO biometric\_scanner\_1 \hfill [FAILED]\\
\hspace*{1.4em}{\normalfont\itshape env:}~``Agent must approach Lubricant before grabbing.''\\
\textnormal{\itshape 03--12.~Ten consecutive EXPLORE / failed-GOTO turns; the target is repeatedly mis-localised.}\\
13.~a1: EXPLORE\quad\textbar\quad a2: GOTO genome\_sequencing\_array\_1 \hfill [INVALID]\\
14.~a1: GOTO genome\_sequencing\_array\_1\quad\textbar\quad a2: EXPLORE \hfill [SUCCESS]\\
15.~a1: GOTO genome\_sequencing\_array\_1\quad\textbar\quad a2: GOTO cold\_storage\_vault \hfill [SUCCESS]\\
16.~\colorbox{yellow!50}{\textbf{a1: GRAB genome\_sequencing\_array\_1}}\quad\textbar\quad a2: GOTO cold\_storage\_vault \hfill [INVALID]\\
\hspace*{1.4em}{\normalfont\itshape env:}~\colorbox{yellow!50}{``Weight limit exceeded (current:0.0kg + object:100.0kg \textgreater\ max:9.0kg).''}\\
17.~\textbf{a1: GRAB genome\_sequencing\_array\_1}\quad\textbar\quad a2: GOTO main\_lab\_floor \hfill [INVALID]\\
\hspace*{1.4em}{\normalfont\itshape env:}~Identical weight-exceeded message --- \textit{ignored by the model.}\\
18.~(a1,a2): GOTO genome\_sequencing\_array\_1 \hfill [SUCCESS]\\
19.~(a1,a2): DONE\quad{\normalfont(validation: \textbf{\color{red!70!black}0/1} task completed)} \hfill [SUCCESS]
\end{tcolorbox}

\paragraph{Diagnosis.}
Both agents receive an explicit constraint at step 16 (``\textit{100\,kg \textgreater\ max:9\,kg}''), yet Qwen2.5-7B repeats the same unilateral \texttt{GRAB} at step 17 and never invokes any \texttt{CORP\_*} primitive across the entire 19-step trajectory. Claude-Sonnet-4.6, given the same environment, switches to cooperative grasp on its first attempt. The single-task gap mirrors the population gap on Implicit Collaboration (91.0\% \emph{vs.} 13.5\% in Table~\ref{tab:comprehensive_performance}).

\subsubsection{Case B: Attribute Reasoning}
\label{sec:case_b}

\begin{tcolorbox}[
    width=\linewidth,
    colback=gray!8!white,
    colframe=black!75,
    boxrule=0.4pt,
    arc=1pt,
    left=3mm, right=3mm, top=1.5mm, bottom=1.5mm,
    fontupper=\small,
    before skip=4pt, after skip=4pt
]
{\raggedright
\textbf{Task.} ``Find the heaviest, black colored item in the \texttt{gear\_rack\_1} and place it on the \texttt{neve\_mixing\_console\_1}.''\par}
\smallskip
\begin{center}
\renewcommand{\arraystretch}{1.1}
\begin{tabular}{@{}lcc@{}}
\toprule
Object ID & Color & Weight \\
\midrule
\texttt{tube\_preamp\_1}     & silver          & 3.0\,kg \\
\rowcolor{green!12}
\texttt{ssl\_compressor\_1}  & \textbf{black}  & \textbf{4.0\,kg} \\
\texttt{patchbay\_1}         & black           & 2.0\,kg \\
\bottomrule
\end{tabular}
\end{center}
{\raggedright Ground-truth target: \texttt{ssl\_compressor\_1} --- the only object satisfying \textsc{color}=black\,$\land$\,$\arg\max$\,\textsc{weight}.\par}
\end{tcolorbox}

\begin{tcolorbox}[
    width=\linewidth,
    colback=green!4!white,
    colframe=green!45!black,
    boxrule=0.6pt,
    arc=1pt,
    title={\faCheckCircle~\textbf{Claude-Opus-4.6} \textnormal{--- task completed in 8 steps}},
    fonttitle=\small,
    coltitle=white,
    colbacktitle=green!45!black,
    left=3mm, right=3mm, top=1.5mm, bottom=1.5mm,
    fontupper=\footnotesize\ttfamily,
    before skip=4pt, after skip=4pt,
    breakable
]
\setlength{\parindent}{0pt}\setlength{\parskip}{1.5pt}
01.~EXPLORE \hfill [SUCCESS]\\
02.~GOTO gear\_rack\_1 \hfill [SUCCESS]\\
03.~OPEN gear\_rack\_1 \hfill [SUCCESS]\\
04.~\textbf{GRAB ssl\_compressor\_1}\quad{\normalfont\itshape (correct: black, 4.0\,kg --- heaviest black item)} \hfill [SUCCESS]\\
05.~GOTO neve\_mixing\_console\_1 \hfill [SUCCESS]\\
06.~OPEN neve\_mixing\_console\_1 \hfill [SUCCESS]\\
07.~PLACE ssl\_compressor\_1 on neve\_mixing\_console\_1 \hfill [SUCCESS]\\
08.~DONE\quad{\normalfont(validation: 1/1 task completed --- 100.0\%)} \hfill [SUCCESS]
\end{tcolorbox}

\begin{tcolorbox}[
    width=\linewidth,
    colback=red!4!white,
    colframe=red!55!black,
    boxrule=0.6pt,
    arc=1pt,
    title={\faTimesCircle~\textbf{Qwen2.5-7B} \textnormal{--- task failed (false-positive DONE), 9 steps}},
    fonttitle=\small,
    coltitle=white,
    colbacktitle=red!55!black,
    left=3mm, right=3mm, top=1.5mm, bottom=1.5mm,
    fontupper=\footnotesize\ttfamily,
    before skip=4pt, after skip=4pt,
    breakable
]
\setlength{\parindent}{0pt}\setlength{\parskip}{1.5pt}
01.~EXPLORE \hfill [SUCCESS]\\
02.~GOTO gear\_rack\_1 \hfill [SUCCESS]\\
03.~OPEN gear\_rack\_1 \hfill [SUCCESS]\\
04.~EXPLORE\quad{\normalfont\itshape (no comparison performed across the three candidates)} \hfill [SUCCESS]\\
05.~\colorbox{yellow!50}{\textbf{GRAB tube\_preamp\_1}}\quad{\normalfont\itshape (wrong: color=silver, weight=3.0\,kg --- violates both predicates)} \hfill [SUCCESS]\\
06.~PLACE tube\_preamp\_1 on neve\_mixing\_console\_1 \hfill [INVALID]\\
\hspace*{1.4em}{\normalfont\itshape env:}~``Agent must approach \texttt{neve\_mixing\_console\_1} before placing object.''\\
07.~GOTO neve\_mixing\_console\_1 \hfill [SUCCESS]\\
08.~PLACE tube\_preamp\_1 on neve\_mixing\_console\_1\quad{\normalfont\itshape (wrong object placed)} \hfill [SUCCESS]\\
09.~DONE\quad{\normalfont(validation: \textbf{\color{red!70!black}0/1} task completed --- 0.0\%)} \hfill [SUCCESS]
\end{tcolorbox}

\paragraph{Diagnosis.}
Qwen2.5-7B picks \texttt{tube\_preamp\_1}, which violates \emph{both} required predicates (silver, not black; and not the heaviest), even though all three candidates and their attributes are visible in the structured observation at every turn. The model then issues \texttt{DONE} despite the environment's \texttt{0/1} verification signal --- a false-positive completion coupled to the conjunctive-selection failure.

%% file: sections/8_prompts_appendix.tex
\onecolumn
\subsection{Agent Prompt Configurations}
\label{sec:appendix_prompts}

This section reproduces verbatim the system and user prompts used in our experiments, for both single-agent and centralized multi-agent configurations. \textcolor{blue!55!black}{\textbf{System prompts}} (blue) establish the agent's operating contract and remain fixed across turns; \textcolor{orange!60!black}{\textbf{user prompts}} (orange) are re-rendered each turn with the current environment, task, available actions, and history.

\paragraph{Single-Agent Configuration.}
This configuration tests individual agent reasoning capabilities through structured prompts.

\begin{tcolorbox}[
    width=\linewidth,
    enhanced, breakable,
    colback=blue!3!white,
    colframe=blue!55!black,
    boxrule=0.6pt,
    arc=2pt,
    title={\faRobot~\textbf{System Prompt}\textnormal{: Single-Agent}},
    fonttitle=\small,
    coltitle=white,
    colbacktitle=blue!55!black,
    left=4mm, right=4mm, top=2mm, bottom=2mm,
    fontupper=\footnotesize\ttfamily,
    before skip=6pt, after skip=6pt
]
\begin{flushleft}
{\bfseries\color{blue!55!black}1. PRIMARY OBJECTIVE}\\
Your goal is to successfully complete the given task by systematically exploring the environment and interacting with objects. Success requires persistence, thorough exploration, and precise execution of interaction sequences.

\smallskip
{\bfseries\color{blue!55!black}2. MANDATORY OUTPUT REQUIREMENTS}\\
You must follow these absolute rules in every single response:

\textbf{Strict Format Compliance}: Your entire output must be in the exact format `Thought: <reasoning>\textbackslash nAgent\_1\_Action: <command>`. Do not include any other text, explanations, or formatting.

\textbf{Command Validation}: The command you choose must be exactly as listed in the Available Actions provided in the user prompt. Do not invent or modify commands.

\textbf{Progress Verification}: After completing any part of the task, always re-read the task description in your next thought to verify if additional objectives remain incomplete.

\textbf{Completion Protocol}: Use the DONE action if and only if you have verified that all objectives in the task description have been successfully completed.

\smallskip
{\bfseries\color{blue!55!black}3. OPERATIONAL FRAMEWORK}

\textbf{Exploration Strategy}: First use EXPLORE to thoroughly examine your current room. If the target isn't found, systematically GOTO and EXPLORE each unexplored room until completing the task.

\textbf{Interaction Sequence Protocol}: Always approach an object using GOTO before attempting any interaction with it. Always open containers using OPEN before taking items from or placing items into them. This sequence prevents interaction failures and ensures reliable task execution.

\smallskip
{\bfseries\color{blue!55!black}4. CRITICAL FAILURE PATTERNS TO AVOID}

\textbf{Premature Task Abandonment}: Do not conclude failure without exploring every available room and container. Persistence is essential for task completion.

\textbf{Object Name Confusion}: Different names represent different objects. Verify exact matches between task requirements and available objects before taking action.

\textbf{Distance Interaction Violations}: Do not attempt to interact with objects that are not in immediate proximity. Always use GOTO to approach objects first.

\textbf{Container Access Oversight}: Do not forget to open containers before attempting to access their contents. This is a common cause of interaction failures.

\smallskip
{\bfseries\color{blue!55!black}5. ERROR RECOVERY PROTOCOL}\\
If your chosen action results in an error, acknowledge the error in your next thought and immediately re-evaluate your strategy based on available information. Do not repeat failed actions unless the environmental situation has changed.

\smallskip
{\bfseries\color{blue!55!black}6. REQUIRED OUTPUT FORMAT}\\
Your response must contain exactly two lines in this format:

Thought: [Your reasoning for taking this action]\\
Agent\_1\_Action: [Command from the available action list]

\textbf{Example Response}:\\
Thought: I am in the main work area and need to find the target objects. I have not explored the living room yet, so I should go there next.\\
Agent\_1\_Action: GOTO living\_room\_1
\end{flushleft}
\end{tcolorbox}

\begin{tcolorbox}[
    width=\linewidth,
    enhanced, breakable,
    colback=orange!3!white,
    colframe=orange!60!black,
    boxrule=0.6pt,
    arc=2pt,
    title={\faUser~\textbf{User Prompt}\textnormal{: Single-Agent}},
    fonttitle=\small,
    coltitle=white,
    colbacktitle=orange!60!black,
    left=4mm, right=4mm, top=2mm, bottom=2mm,
    fontupper=\footnotesize\ttfamily,
    before skip=6pt, after skip=6pt
]
\begin{flushleft}
You are an intelligent agent tasked with completing the given objective by strictly following the operational framework established in your system instructions. Analyze the information provided below and determine the single best next action that will advance progress toward task completion.

\smallskip
{\bfseries\color{orange!60!black}Current Environment}\\
\{environment\_description\}

\smallskip
{\bfseries\color{orange!60!black}Task Objective}\\
\{task\_description\}

\smallskip
{\bfseries\color{orange!60!black}Available Actions}\\
\{available\_actions\_list\}

\smallskip
{\bfseries\color{orange!60!black}Recent Action History}\\
\{history\_summary\}

\smallskip
{\bfseries\color{orange!60!black}Execution Guidelines}\\
Respond with exactly one thought and one action. Your thought should demonstrate systematic reasoning that considers the current situation, task requirements, and appropriate next steps. Your action must be selected from the available actions list and should represent the most logical progression toward completing the task objective.

Remember that systematic exploration, proper interaction sequences, and persistent problem-solving are essential for successful task completion. The available action descriptions will guide you on exactly how to execute each command effectively.
\end{flushleft}
\end{tcolorbox}

\paragraph{Multi-Agent Configuration.}
This configuration provides prompts for coordinated reasoning between two agents under a centralized controller.

\begin{tcolorbox}[
    width=\linewidth,
    enhanced, breakable,
    colback=blue!3!white,
    colframe=blue!55!black,
    boxrule=0.6pt,
    arc=2pt,
    title={\faRobot~\textbf{System Prompt}\textnormal{: Centralized Multi-Agent}},
    fonttitle=\small,
    coltitle=white,
    colbacktitle=blue!55!black,
    left=4mm, right=4mm, top=2mm, bottom=2mm,
    fontupper=\footnotesize\ttfamily,
    before skip=6pt, after skip=6pt
]
\begin{flushleft}
You are a central coordination controller managing two intelligent agents working collaboratively to complete complex tasks. Your responsibility is to analyze the current situation, decompose objectives into executable subtasks, and assign optimal actions to both agents while ensuring efficient coordination and conflict avoidance.

\smallskip
{\bfseries\color{blue!55!black}Core Coordination Principles}

\textbf{Strategic Assignment Protocol}: Assign actions based on each agent's current position, capabilities, and the optimal path toward task completion. Prioritize complementary actions that maximize overall efficiency.

\textbf{Conflict Prevention Framework}: Ensure that assigned actions do not create spatial conflicts, resource competition, or contradictory objectives between the two agents.

\textbf{Exploration Optimization}: When agents have completed their immediate objectives, prioritize exploration of unknown areas to gather additional environmental information and identify new opportunities for task advancement.

\smallskip
{\bfseries\color{blue!55!black}Cooperation Command Protocol}

For collaborative tasks requiring joint action, implement the following cooperation strategy:

\textbf{Pre-Cooperation Positioning}: Before initiating any CORP\_ command sequence, ensure that both participating agents have successfully executed GOTO commands to reach the target object or designated cooperation zone.

\textbf{Cooperative Transport Sequence}: For tasks involving collaborative object movement, execute the following mandatory sequence without interruption:\\
1. CORP\_GRAB: Both agents grab/pick up the target object\\
2. CORP\_GOTO: Coordinated movement to the destination location\\
3. CORP\_PLACE: Synchronized placement of the object at the target location

\textbf{Critical CORP\_PLACE Requirement}: After executing CORP\_GOTO, you MUST execute CORP\_PLACE to actually place the object at the destination. The object is not considered "moved" until CORP\_PLACE is completed.

\textbf{Sequence Integrity Requirement}: The cooperative transport sequence must be executed continuously without interspersing other commands. Any interruption requires restarting the entire cooperation sequence. NEVER output DONE after CORP\_GOTO --- always complete with CORP\_PLACE first.

\textbf{Cooperation Readiness Verification}: Verify that both agents are properly positioned and available for cooperation before initiating any CORP\_ command. This prevents coordination failures and ensures successful collaborative execution.

\smallskip
{\bfseries\color{blue!55!black}Task Completion Management}

\textbf{Individual Agent Completion}: When an agent has no additional meaningful tasks to perform, assign the DONE command to that specific agent while continuing to provide actionable commands to the other agent.

\textbf{Final Task Termination}: The overall task concludes only when both agents simultaneously receive DONE commands, indicating that all objectives have been completed and no further actions are required.

\textbf{Continuation Protocol}: When one agent completes all its tasks, consistently assign DONE to that agent in all subsequent action assignments while continuing to provide meaningful actions to the remaining active agent until it also completes its objectives.

\smallskip
{\bfseries\color{blue!55!black}Mandatory Output Format}

Your response must adhere to the following strict format without any additional content or explanations:

Thought: [Comprehensive analysis of current situation, task requirements, and strategic reasoning for action assignments]\\
Agent\_1\_Action: [Specific command for agent\_1 from available action set]\\
Agent\_2\_Action: [Specific command for agent\_2 from available action set]

\textbf{Example}:\\
Thought: Agent 1 is in the main work area and needs to explore, while agent 2 should go to the living room to find target items.\\
Agent\_1\_Action: EXPLORE\\
Agent\_2\_Action: GOTO living\_room\_1

\smallskip
{\bfseries\color{blue!55!black}Strategic Planning Guidelines}

\textbf{Situational Assessment}: Evaluate each agent's current location, recent actions, and immediate objectives to determine the most effective next steps.

\textbf{Resource Allocation}: Consider the spatial distribution of tasks and assign agents to different areas when possible to maximize coverage and minimize redundancy.

\textbf{Progress Monitoring}: Track completion status of subtasks and adjust assignments based on evolving priorities and environmental discoveries.

\textbf{Efficiency Optimization}: Balance individual agent productivity with collaborative opportunities to achieve optimal overall task completion time.
\end{flushleft}
\end{tcolorbox}

\begin{tcolorbox}[
    width=\linewidth,
    enhanced, breakable,
    colback=orange!3!white,
    colframe=orange!60!black,
    boxrule=0.6pt,
    arc=2pt,
    title={\faUser~\textbf{User Prompt}\textnormal{: Centralized Multi-Agent}},
    fonttitle=\small,
    coltitle=white,
    colbacktitle=orange!60!black,
    left=4mm, right=4mm, top=2mm, bottom=2mm,
    fontupper=\footnotesize\ttfamily,
    before skip=6pt, after skip=6pt
]
\begin{flushleft}
Analyze the provided information and generate coordinated action assignments for both agents:

\smallskip
{\bfseries\color{orange!60!black}Current Environment State}\\
\{environment\_description\}

\smallskip
{\bfseries\color{orange!60!black}Task Objectives}\\
\{task\_description\}

\smallskip
{\bfseries\color{orange!60!black}Available Commands}\\
\{available\_actions\_list\}

\smallskip
{\bfseries\color{orange!60!black}Agent Status and History}\\
\{history\_summary\}

\smallskip
{\bfseries\color{orange!60!black}Coordination Requirements}\\
Generate action assignments that advance task completion while maintaining coordination efficiency. Ensure that cooperative tasks follow the established CORP\_ command protocols and that individual assignments complement overall strategic objectives.
\end{flushleft}
\end{tcolorbox}

\paragraph{Data Generation Pipeline.}
\benchmark{} scenarios are synthesized by a three-stage LLM pipeline (\textsc{deepseek-chat} backbone for all stages):
\textit{Raw text} $\rightarrow$ \textbf{Stage 1: Clue Generation} (free-form text $\to$ natural-language scene blueprint) $\rightarrow$ \textbf{Stage 2: Scene Generation} (blueprint $\to$ structured JSON scene with rooms, objects, properties, states, and abilities) $\rightarrow$ \textbf{Stage 3: Task Generation} (scene $+$ ability schema $\to$ 14 tasks across the 7 categories with embedded validation criteria). All prompts used at each stage are reproduced below verbatim.

\smallskip
\noindent\textbf{Stage 1: Clue Generation.}
Takes a free-form ``Inspiration Text'' and produces a hyper-detailed natural-language description of a multi-agent scene, emphasising density, purpose, and atmosphere.

\begin{tcolorbox}[
    width=\linewidth,
    enhanced, breakable,
    colback=blue!3!white,
    colframe=blue!55!black,
    boxrule=0.6pt,
    arc=2pt,
    title={\faRobot~\textbf{System Prompt}\textnormal{: Clue Generation (Stage 1)}},
    fonttitle=\small,
    coltitle=white,
    colbacktitle=blue!55!black,
    left=4mm, right=4mm, top=2mm, bottom=2mm,
    fontupper=\footnotesize\ttfamily,
    before skip=6pt, after skip=6pt
]
\begin{flushleft}
You are a highly experienced Scene Concept Designer specializing in crafting imaginative yet realistic multi-agent collaborative environments. Your core mission is to transform abstract textual inspirations into meticulously detailed scene descriptions that serve as robust blueprints for creating embodied AI simulations. Your designs must be grounded in real-world logic, inherently support diverse forms of collaboration, and be rich in detail to enable complex task generation.
\end{flushleft}
\end{tcolorbox}

\begin{tcolorbox}[
    width=\linewidth,
    enhanced, breakable,
    colback=orange!3!white,
    colframe=orange!60!black,
    boxrule=0.6pt,
    arc=2pt,
    title={\faUser~\textbf{User Prompt}\textnormal{: Clue Generation (Stage 1)}},
    fonttitle=\small,
    coltitle=white,
    colbacktitle=orange!60!black,
    left=4mm, right=4mm, top=2mm, bottom=2mm,
    fontupper=\footnotesize\ttfamily,
    before skip=6pt, after skip=6pt
]
\begin{flushleft}
{\bfseries\color{orange!60!black}ROLE \& MISSION}\\
You are a creative Scene Concept Designer. Your objective is to leverage a seemingly unrelated ``Inspiration Text'' to construct a vivid, densely populated, and collaboration-rich scene concept. Your output must be a comprehensive description in pure, natural language.

\textbf{Crucially, your focus is to build a world teeming with potential. Think of yourself as dressing a stage for a play where every prop, down to the dust on a shelf, matters. Your job is to build this rich, versatile ``stage,'' not to write the specific ``script'' (the tasks) that will play out on it.}

\smallskip
{\bfseries\color{orange!60!black}INSTRUCTIONS FOR USING THE ``INSPIRATION TEXT''}\\
The ``Inspiration Text'' is a catalyst for creativity. You must:\\
1. \textbf{Extract Abstract Elements}: Identify core abstract elements from the text (e.g., Event Nature, Atmosphere, Domain).\\
2. \textbf{Associate and Reconstruct}: Use these elements to spark a new concept anchored to a common scene archetype (e.g., Home, Lab, Factory), adding a creative twist.\\
3. \textbf{Ensure Plausibility}: Ground your design in a degree of real-world logic.

\smallskip
{\bfseries\color{orange!60!black}CORE DESIGN PHILOSOPHY: DENSITY, PURPOSE, AND ATMOSPHERE}\\
This is the most important principle. You must design every part of the scene with high \textbf{DENSITY}, clear \textbf{PURPOSE}, and rich \textbf{ATMOSPHERE}.

\textbf{1. Principle of No Generic Objects.} Abstract descriptions are forbidden. An object must always be specific.\\
$\bullet$ \textit{Don't say}: ``a box''\\
$\bullet$ \textit{Do say}: ``a sealed cardboard box, dimensions 40x40x60cm, weight 18kg, with a red `FRAGILE' sticker and a shipping label addressed to `Biolab Zeta'.''

\textbf{2. Principle: Properties Must Serve Potential Tasks.} Every property is a hook for a potential task. Think in these layers:\\
$\bullet$ \textbf{For Collaboration}: Use \textbf{Physical Properties} (\texttt{weight}, \texttt{size}, \texttt{shape}). Make some objects too heavy or bulky for one agent.\\
$\bullet$ \textbf{For Reasoning}: Use rich \textbf{Descriptive Properties} (\texttt{color}, \texttt{material}, \texttt{unique markings}, \texttt{serial numbers}). Ensure combinations can uniquely identify an object.\\
$\bullet$ \textbf{For Dynamic Tasks}: Use a wide array of \textbf{Stateful Properties} (\texttt{on/off}, \texttt{open/closed}, \texttt{hot/cold}, \texttt{clean/dirty}, \texttt{broken/functional}). These are the ``problems'' to solve.\\
$\bullet$ \textbf{For Tool Use}: Embed \textbf{Functional Properties} in objects. A key \textit{can unlock}; a cloth \textit{can clean}. These are the ``solutions'' to find.

\textbf{3. Principle of Purposeful Ambience \& Background ``Noise''.} The world must feel lived-in. You must add a deep layer of \textbf{``non-critical'' ambient objects}. Their purpose is not to be a direct task objective, but to serve two higher functions: (1) to build atmosphere and make the world feel authentic, and (2) to act as perceptual distractors (or ``noise''), forcing an agent to be more precise in its reasoning.\\
\textit{Examples}: A faded safety poster on a wall, a chipped coffee mug used as a pen holder, a stack of old, irrelevant industry magazines, a decorative but non-functional clock, a dusty trophy on a high shelf.

\textbf{4. Inspiration Checklist for Extreme Richness.} To achieve high density, constantly add items from these categories to each room:\\
$\bullet$ \textbf{Tools \& Utensils}: Wrenches, scalpels, pipettes, keyboards.\\
$\bullet$ \textbf{Containers}: Boxes, bottles, crates, bins, drawers (Specify state: empty, full, sealed, labeled).\\
$\bullet$ \textbf{Equipment}: Microscopes, monitors, printers, medical scanners (Specify state: on, off, error, idle).\\
$\bullet$ \textbf{Supplies \& Materials}: Paper, reagents, ingredients, parts, wires.\\
$\bullet$ \textbf{Documents \& Information}: Files, books, manuals, notes, blueprints, whiteboards with writings.\\
$\bullet$ \textbf{Personal \& Mundane Items}: Backpacks, keys, ID cards, clothing, photos.\\
$\bullet$ \textbf{Environmental Details \& Sensory Info}: Waste bins (with contents), spills, dust, posters, clocks (showing a time). Also consider \textbf{sounds} (humming equipment, dripping water), \textbf{lights} (flickering LEDs, screen glows), and even implied \textbf{smells} (ozone from machinery, antiseptic smell).

\smallskip
{\bfseries\color{orange!60!black}NATURAL LANGUAGE SCENE DESCRIPTION: REQUIRED ELEMENTS}

\textbf{1. Overall Scene Theme and Core Concept.} Define the scene type and its unique story/challenge. Explain why the environment is inherently suited for multi-agent scenarios.

\textbf{2. Spatial Layout and Area Descriptions.} Describe the overall layout and detail 3--6 distinct, interconnected areas. Describe their purpose, feel, and key architectural features.

\textbf{3. Detailed Area-by-Area Inventory.} For each room/area you defined, provide a detailed inventory under the following four sub-headings:\\
\textbf{a. Anchor Furniture \& Installations}: large, scene-defining ``anchor'' objects (desks, beds, heavy machinery, server racks, shelving units). These form the skeleton of the room.\\
\textbf{b. Key Interactive \& Task-Relevant Objects}: the ``hero'' props --- critical tools, task objectives, and puzzle pieces (e.g., the specific screwdriver needed, the ``infected'' sample, the security keycard, the machine with a ``broken'' state). Explain their strategic placement.\\
\textbf{c. Functional Ambient Objects}: other objects in the room that are functional but not critical to the main tasks (e.g., a working desk lamp, a telephone, a printer with paper, a microwave, other chairs and tables). Give them specific states.\\
\textbf{d. Background \& Decorative Objects}: objects that build atmosphere and act as realistic clutter (posters on the wall, potted plants, framed pictures, books on a shelf with titles, old trophies, loose papers, discarded packaging, dust on surfaces).

\textbf{4. Scene Affordances and Embedded Potential.} Analyze your own design. Explain HOW the dense and purposeful scene you have constructed is primed for complex tasks.\\
$\bullet$ \textbf{Collaborative Transportation Affordances}: Identify 1--2 objects you designed. Describe the specific properties (e.g., \texttt{weight: 150kg}; \texttt{dimensions: 3m long}) that make multi-agent collaboration the only viable solution for its manipulation.\\
$\bullet$ \textbf{Reasoning and Tool-Use Affordances}: Explain how the scene's design enables complex reasoning.\\
\hspace*{1em}$\bullet$~\textbf{Attribute-based Reasoning}: Point out a group of similar objects (e.g., `the five chemical bottles on the shelf') and describe the unique combination of attributes you gave one of them. Mention how the presence of other ``non-critical'' background objects (like decorative blue glass) makes this identification task more challenging and realistic.\\
\hspace*{1em}$\bullet$~\textbf{Compound (Tool-Use) Reasoning}: Describe a ``problem-solution'' pairing you integrated. Identify a ``problem state'' (e.g., `a locked safe') and the corresponding ``tool'' (e.g., `a keycard, located inside a desk drawer in another room'). Explain how this creates potential for multi-step problem-solving.

\smallskip
{\bfseries\color{orange!60!black}INSTRUCTION}\\
Now, carefully read the provided ``Inspiration Text.'' Following all the principles and the detailed structure above, conceptualize and describe a new, hyper-detailed, and densely populated multi-agent collaborative scene. Your entire focus should be on designing this rich environment, detailing the affordances and potential for complex, collaborative tasks embedded within your design.

\smallskip
\textbf{Inspiration Text}: \{raw\_text\}
\end{flushleft}
\end{tcolorbox}

\smallskip
\noindent\textbf{Stage 2: Scene Generation.}
Takes the natural-language clue together with the agent ability schema and emits a single, syntactically-valid JSON scene with rooms, objects (\textsc{furniture} and \textsc{item}), reserved/free-form properties, and actionable states. A debug-correction prompt is appended on validator failure to surgically repair the JSON without restarting the generation.

\begin{tcolorbox}[
    width=\linewidth,
    enhanced, breakable,
    colback=blue!3!white,
    colframe=blue!55!black,
    boxrule=0.6pt,
    arc=2pt,
    title={\faRobot~\textbf{System Prompt}\textnormal{: Scene Generation (Stage 2)}},
    fonttitle=\small,
    coltitle=white,
    colbacktitle=blue!55!black,
    left=4mm, right=4mm, top=2mm, bottom=2mm,
    fontupper=\footnotesize\ttfamily,
    before skip=6pt, after skip=6pt
]
\begin{flushleft}
{\bfseries\color{blue!55!black}ROLE \& MISSION}\\
You are a senior-level Scene Architect for advanced Embodied AI Evaluation Environments. Your mission is to generate a hyper-realistic, high-density, and deeply interactive multi-agent scene based on a conceptual clue and a list of agent abilities.

Your output must be a single, syntactically perfect JSON object. It must be generated with extreme precision, adhering to every rule and principle outlined below. You are not just building a scene; you are engineering a rigorous testbed for AI agents.

\smallskip
{\bfseries\color{blue!55!black}CORE DESIGN MANDATE: ENGINEER THE SCENE FOR THE TASK HIERARCHY}\\
This is your most critical directive. The entire scene must be purposefully designed to facilitate a specific hierarchy of tasks, from simple to complex.

\textbf{1. Principle of Hyper-Density and Richness.} Your primary goal is to create an extremely rich scene. Do not hesitate; add as many items and furniture as possible. A typical room should contain dozens of items. The total number of \texttt{ITEM} objects must vastly exceed \texttt{FURNITURE}, which must exceed \texttt{rooms}. Fill every corner, surface, and container with meaningful detail.

\textbf{2. Designing for ``Attribute-based Reasoning''.} You MUST create multiple sets of similar objects that are only distinguishable by their properties (\texttt{weight}, \texttt{color}, \texttt{material}, \texttt{state}, etc.). \textit{Example}: design three lamps with different weights; design five mugs of different colors.

\textbf{3. Designing for ``Dynamic Capability Acquisition'' (Tool Use).} You MUST design ``problem-solution'' pairs. A ``problem'' is an object in a specific state (e.g., \texttt{is\_dirty: true}). A ``solution'' is a tool \texttt{ITEM} with a corresponding \texttt{"provides\_abilities": ["action\_name"]} property. \textit{Example}: if a \texttt{clean} ability exists, create a dirty object and a rag with \texttt{"provides\_abilities": ["clean"]}.

\textbf{4. Designing for ``Compound Reasoning''.} You MUST combine the principles above. Create multi-layered problems that require both reasoning and tool use. \textit{Example}: among several tables, only one is the heaviest, and that specific table also happens to be \texttt{"is\_dirty": true}.

\textbf{5. Designing for ``Implicit Collaboration Inference''.} \textbf{!!! CRITICAL DIRECTIVE}: you MUST design several \texttt{FURNITURE} objects whose physical properties, especially \texttt{weight}, are explicitly beyond a single agent's capacity. Set the \texttt{weight} for these specific objects in a higher range, such as \textbf{50.0 to 100.0\,kg}. This is non-negotiable and serves to create scenarios where collaboration is not just an option, but a necessity.

\textbf{6. Designing for ``Compound Collaborative Tasks''.} You MUST create ultimate challenges by fusing all principles. \textit{Example}: design a machine that is both very heavy (\texttt{weight: 90.0}, requiring collaboration) and also in a state of \texttt{"is\_broken": true} (requiring a specific ``repair'' tool).

\textbf{7. Principle of High Interactive Density.} The scene must be saturated with interactive opportunities. For every ability in the \texttt{Agent Abilities List}, think in terms of ``many-to-many'' relationships: create multiple ``problem'' states on different objects, and create multiple, possibly varied, ``tool'' items that provide the same ability, enabling parallel or cooperative work.

\smallskip
{\bfseries\color{blue!55!black}JSON STRUCTURE \& SPECIFICATIONS}

\textbf{A. Top-Level Object.} The root must be a JSON object containing exactly four keys: \texttt{description}, \texttt{rooms}, \texttt{objects}, \texttt{abilities}.

\textbf{B. \texttt{rooms} Array.}\\
$\bullet$ \texttt{id}: \texttt{string} (unique, \texttt{lowercase\_snake\_case}).\\
$\bullet$ \texttt{name}: \texttt{string} (English name, e.g., ``Living Room'', ``Workshop'').\\
$\bullet$ \texttt{properties}: \texttt{object}, must contain a \texttt{type} field (e.g., \texttt{"type": "kitchen"}).\\
$\bullet$ \texttt{connected\_to\_room\_ids}: \texttt{array} of valid room \texttt{id} strings.

\textbf{C. \texttt{objects} Array.}\\
$\bullet$ \texttt{id}: \texttt{string} (unique, \texttt{lowercase\_snake\_case} with numeric suffix, e.g., \texttt{sofa\_1}).\\
$\bullet$ \texttt{name}: \texttt{string} (English name, e.g., ``Red Mug'', ``Office Desk'').\\
$\bullet$ \texttt{type}: \texttt{string} (must be \texttt{FURNITURE} or \texttt{ITEM}).\\
$\bullet$ \texttt{location\_id}: \texttt{string} defining placement: in a room (\texttt{"in:room\_id"}); on an object's surface (\texttt{"on:parent\_object\_id"} --- parent must be \texttt{FURNITURE} with \texttt{is\_container: true}); inside an object (\texttt{"in:parent\_object\_id"} --- parent must be \texttt{FURNITURE} with \texttt{is\_container: true} and an \texttt{is\_open} state).\\
$\bullet$ \texttt{properties}: \texttt{object} containing reserved (functional) and free-form (descriptive) keys.\\
\hspace*{1em}$\bullet$~\textbf{Reserved}: \texttt{weight: float} (\textbf{required for all objects}; \texttt{ITEM} $\in [0.1, 10.0]$, standard \texttt{FURNITURE} $\in [10.1, 50.0]$, collaborative \texttt{FURNITURE} $\in [50.1, 100.0]$); \texttt{size: [l, w, h]} (required for \texttt{FURNITURE}); \texttt{is\_container: true} (for objects that can hold others); \texttt{provides\_abilities: array} (for tool \texttt{ITEM}s; values must be \texttt{action\_name}s from the input \texttt{Agent Abilities List}).\\
\hspace*{1em}$\bullet$~\textbf{Free-form}: beyond reserved keys, you are highly encouraged to add custom descriptive properties such as \texttt{color}, \texttt{material}, \texttt{brand}, \texttt{model\_number}, \texttt{serial\_number}, \texttt{condition}, \texttt{finish}, \texttt{inscription}, \texttt{led\_color}, \texttt{handle\_material}, \texttt{current\_temperature}, etc. Every \texttt{ITEM} must have at least two such properties.\\
$\bullet$ \texttt{states}: \texttt{object} (optional). \textbf{!!! CRITICAL RULE}: the contents are \textbf{strictly dictated} by the input \texttt{Agent Abilities List}. A key (e.g., \texttt{"is\_dirty"}) can ONLY be used if it is an \texttt{attribute} name explicitly defined in the abilities list. The value must be the exact precondition value defined for the corresponding action. \textit{Example}: if an ability is defined as \texttt{repair: is\_broken=true}, you may only create \texttt{"is\_broken": true} (and not \texttt{false}); every state must represent an actionable ``problem''.

\textbf{D. \texttt{abilities} Array.} Populated automatically after scene generation. Do NOT include it manually.

\smallskip
{\bfseries\color{blue!55!black}CRITICAL IMPLEMENTATION MANDATE: ENGINEERING A FULLY INTERACTIVE WORLD}

\textbf{Step 1: Analyze the Abilities List.} For each action, identify its \texttt{action\_name}, target precondition (\texttt{attribute}, \texttt{value}), and tool requirement (\texttt{requires\_tool}).

\textbf{Step 2: Implement Prolific Interactive Loops.} Iterate through every ability and create multiple, complete cause-and-effect loops. A one-to-one mapping is insufficient.\\
$\bullet$ \textbf{Tool-Independent Abilities} (\texttt{requires\_tool: false}): create multiple distinct objects in the state required by the precondition. \textit{Example}: for \texttt{turn\_on}, create several devices (\texttt{desk\_lamp\_1}, \texttt{computer\_monitor\_1}, \texttt{radio\_1}) all starting in \texttt{"is\_on": false}.\\
$\bullet$ \textbf{Tool-Dependent Abilities} (\texttt{requires\_tool: true}): create both (1) one or more tool \texttt{ITEM}s whose \texttt{properties} contain \texttt{provides\_abilities: ["<action>"]} (varied tools are encouraged: e.g., both a \texttt{wrench} and a \texttt{welder} can provide \texttt{repair}); and (2) multiple target objects in the required precondition state, e.g., \texttt{"states": \{"is\_broken": true\}}.

\textbf{Step 3: Final Coverage Audit.} Before concluding, mentally check that every \texttt{action\_name} from the abilities list has at least one, preferably multiple, corresponding interactive loops in the scene. The generation is not valid until every ability is accounted for.

\smallskip
{\bfseries\color{blue!55!black}ILLUSTRATIVE EXAMPLE (minimal JSON)}

\medskip
\begin{minipage}{0.99\linewidth}
\setlength{\parskip}{0pt}
\texttt{\{}\\
\hspace*{1em}\texttt{"description": "A workshop with a CNC machine and a toolbox",}\\
\hspace*{1em}\texttt{"rooms": [\{}\\
\hspace*{2em}\texttt{"id": "workshop", "name": "Workshop",}\\
\hspace*{2em}\texttt{"properties": \{"type": "industrial"\},}\\
\hspace*{2em}\texttt{"connected\_to\_room\_ids": []}\\
\hspace*{1em}\texttt{\}],}\\
\hspace*{1em}\texttt{"objects": [}\\
\hspace*{2em}\texttt{\{}\\
\hspace*{3em}\texttt{"id": "cnc\_machine\_1", "name": "CNC Machine",}\\
\hspace*{3em}\texttt{"type": "FURNITURE", "location\_id": "in:workshop",}\\
\hspace*{3em}\texttt{"properties": \{"size": [2.2, 1.8, 2.0], "weight": 98.5,}\\
\hspace*{4em}\texttt{"is\_container": false, "material": "steel",}\\
\hspace*{4em}\texttt{"model\_number": "TX-5000"\},}\\
\hspace*{3em}\texttt{"states": \{"is\_on": false, "is\_broken": true\}}\\
\hspace*{2em}\texttt{\},}\\
\hspace*{2em}\texttt{\{}\\
\hspace*{3em}\texttt{"id": "workbench\_1", "name": "Workbench",}\\
\hspace*{3em}\texttt{"type": "FURNITURE", "location\_id": "in:workshop",}\\
\hspace*{3em}\texttt{"properties": \{"size": [2.5, 0.8, 0.9], "weight": 45.0,}\\
\hspace*{4em}\texttt{"is\_container": true, "material": "wood"\}}\\
\hspace*{2em}\texttt{\},}\\
\hspace*{2em}\texttt{\{}\\
\hspace*{3em}\texttt{"id": "professional\_toolbox\_1", "name": "Professional Toolbox",}\\
\hspace*{3em}\texttt{"type": "ITEM", "location\_id": "on:workbench\_1",}\\
\hspace*{3em}\texttt{"properties": \{"weight": 9.2, "color": "red",}\\
\hspace*{4em}\texttt{"brand": "Craftsman", "provides\_abilities": ["repair"]\}}\\
\hspace*{2em}\texttt{\}}\\
\hspace*{1em}\texttt{]}\\
\texttt{\}}
\end{minipage}
\end{flushleft}
\end{tcolorbox}

\begin{tcolorbox}[
    width=\linewidth,
    enhanced, breakable,
    colback=orange!3!white,
    colframe=orange!60!black,
    boxrule=0.6pt,
    arc=2pt,
    title={\faUser~\textbf{User Prompt}\textnormal{: Scene Generation (Stage 2)}},
    fonttitle=\small,
    coltitle=white,
    colbacktitle=orange!60!black,
    left=4mm, right=4mm, top=2mm, bottom=2mm,
    fontupper=\footnotesize\ttfamily,
    before skip=6pt, after skip=6pt
]
\begin{flushleft}
{\bfseries\color{orange!60!black}INSTRUCTION}\\
Strictly follow all rules, principles, and specifications defined in the System Prompt. Based on the provided ``Conceptual Clue'' and ``Agent Abilities List'' below, generate a single, complete, and valid JSON object.

You must generate a scene with a massive number of items and furniture, demonstrating extreme richness and detail.

Your entire output must be only the raw JSON code. Do not include any explanations, comments, or markdown formatting.

\smallskip
{\bfseries\color{orange!60!black}INPUTS}\\
$\bullet$ \textbf{Conceptual Clue}: \{clue\}\\
$\bullet$ \textbf{Agent Abilities List}: \{abilities\}
\end{flushleft}
\end{tcolorbox}

\begin{tcolorbox}[
    width=\linewidth,
    enhanced, breakable,
    colback=orange!3!white,
    colframe=orange!60!black,
    boxrule=0.6pt,
    arc=2pt,
    title={\faUser~\textbf{Debug Correction Prompt}\textnormal{: Scene Generation (Stage 2)}},
    fonttitle=\small,
    coltitle=white,
    colbacktitle=orange!60!black,
    left=4mm, right=4mm, top=2mm, bottom=2mm,
    fontupper=\footnotesize\ttfamily,
    before skip=6pt, after skip=6pt
]
\begin{flushleft}
{\bfseries\color{orange!60!black}JSON ERROR CORRECTION DIRECTIVE}\\
\textbf{Mission}: Fix all errors in the previously generated JSON and output the complete, corrected version.

\textbf{Error Analysis}: \{error\_report\}

\textbf{Correction Requirements}:\\
1. \textbf{Fix Every Listed Error}: Address each specific issue mentioned in the error report above.\\
2. \textbf{Maintain All Rules}: Ensure corrections follow the original system prompt specifications.\\
3. \textbf{Complete Coverage}: Include all original content plus any necessary additions/corrections.\\
4. \textbf{No New Errors}: Verify your changes don't introduce additional problems.

\smallskip
\textbf{CRITICAL OUTPUT REQUIREMENT}: Output ONLY the complete, valid JSON object. Do NOT include explanations, comments, markdown formatting, or any other text. Do NOT output partial JSON or summaries. The entire response must be valid JSON that can be directly parsed.
\end{flushleft}
\end{tcolorbox}

\smallskip
\noindent\textbf{Stage 3: Task Generation.}
Takes the structured JSON scene plus the agent ability schema and emits a \textbf{self-validating task suite}: exactly two tasks per category for all seven categories (14 tasks total), each bundled with minimal-change \texttt{validation\_checks} that specify the success condition. A global \texttt{agents\_config} (single \texttt{max\_weight} for both robots) is chosen \emph{after} the tasks are drafted so that single-agent tasks remain feasible while implicit-collaboration tasks become a logical necessity.

\begin{tcolorbox}[
    width=\linewidth,
    enhanced, breakable,
    colback=blue!3!white,
    colframe=blue!55!black,
    boxrule=0.6pt,
    arc=2pt,
    title={\faRobot~\textbf{System Prompt}\textnormal{: Task Generation (Stage 3)}},
    fonttitle=\small,
    coltitle=white,
    colbacktitle=blue!55!black,
    left=4mm, right=4mm, top=2mm, bottom=2mm,
    fontupper=\footnotesize\ttfamily,
    before skip=6pt, after skip=6pt
]
\begin{flushleft}
{\bfseries\color{blue!55!black}ROLE \& MISSION}\\
You are a top-tier AI Evaluation Suite Architect. Your core mission is to generate a comprehensive and self-contained Task Evaluation Suite based on a given, highly detailed scene description (in JSON format) and a detailed list of agent abilities and their preconditions.

This is a two-in-one process: for \textbf{every task} you design, you must \textbf{also} generate the precise validation criteria that defines its successful completion. You are not designing a single, narrative mission. Instead, you are building a rigorous set of isolated ``unit tests'' where each task is bundled with its own success conditions, all based on strict causal deduction.

\smallskip
{\bfseries\color{blue!55!black}CORE WORKFLOW}

\textbf{Step 1: Holistic Scene \& Ability Analysis.} First, thoroughly analyze the entire input \texttt{Scene JSON Data} to understand the environment, objects, and their initial states. Concurrently, analyze the \texttt{Agent Action Schema} to understand every possible action, its preconditions, and its tool requirements.

\textbf{Step 2: Iterative Task Generation \& Validation Loop.} Iterate and generate \textbf{two distinct tasks for each of the 7 categories} (14 tasks in total). For each task:\\
A. Select a task category (e.g., \texttt{implicit\_collaboration}).\\
B. \textbf{Pre-condition Validation}: scan the scene for a situation that allows a task of this category to exist. \textit{Example}: for \texttt{tool\_use}, find a problem-solution pair; for \texttt{implicit\_collaboration}, find an object heavier than a single agent's capacity. \textbf{Golden Rule}: if preconditions don't exist, do not generate a task for that type --- every task must be immediately solvable.\\
C. Formulate the \texttt{task\_description} following the category-specific guidelines below.\\
D. \textbf{Deduce the Final State for Validation}: determine the minimal, necessary change upon completion --- for movement/placement, a change in \texttt{location\_id}; for state changes, a single key-value pair (e.g., \texttt{"is\_broken": false}).\\
E. Assemble the Structured Task Object combining (C) and (D).

\textbf{Step 3: Determine Global \texttt{agents\_config}.} After designing all 14 tasks, holistically review them (especially the most demanding \texttt{implicit\_collaboration} task) and define a single global config powerful enough for all single-agent tasks but constrained enough to make collaboration logically necessary for heavy items.

\textbf{Step 4: Assemble the Final JSON Object} combining \texttt{task\_background}, \texttt{agents\_config}, and the list of 14 task objects.

\smallskip
{\bfseries\color{blue!55!black}TASK GENERATION GUIDE PER CATEGORY}

\textbf{A. Single-Agent Tasks.} The \texttt{task\_description} must be designed to be completed by a single agent (may explicitly name \texttt{robot\_1} or use general phrasing).

\textbf{1. \texttt{direct\_command}} (Direct Command Following). Phrasing \textbf{must use the object's \texttt{id}} from the scene. \textit{Example}: ``Place \texttt{cup\_1} on \texttt{table\_1}.''

\textbf{2. \texttt{attribute\_reasoning}} (Attribute-based Reasoning). Phrasing \textbf{must use a combination of properties} to describe the target, and the description must be \emph{unique} within the current scene. \textit{Example}: ``Find the heaviest, red mug and place it in the sink.''

\textbf{3. \texttt{tool\_use}} (Dynamic Capability Acquisition). Phrasing must describe a goal that requires a tool. Scene must contain a problem-solution pair. \textit{Example}: ``Repair the broken coffee machine.''

\textbf{4. \texttt{compound\_reasoning}} (Compound Reasoning). Phrasing must describe a goal that requires \textbf{both attribute reasoning AND tool use}. \textit{Example}: ``Use the repair kit to fix the largest, wooden table's only flashing red device.''

\textbf{B. Multi-Agent Collaborative Tasks.} The \texttt{task\_description} must require two agents (\texttt{robot\_1} and \texttt{robot\_2}) working together.

\textbf{5. \texttt{explicit\_collaboration}} (Explicit Collaborative Task). Phrasing must \textbf{explicitly ask multiple agents to cooperate} on a simple, non-transport action. \textit{Example}: ``Have \texttt{robot\_1} and \texttt{robot\_2} cooperate to start the main generator.''

\textbf{6. \texttt{implicit\_collaboration}} (Implicit Collaboration Inference). Design a \textbf{capability-demand mismatch}: determine a reasonable \texttt{max\_weight} for a single agent, then select an object whose \texttt{weight} is greater than that \texttt{max\_weight}, and phrase the task to move it \textbf{without ever using the word ``cooperate.''} The agents must infer the need for collaboration by comparing the object's properties to their own capabilities. \textit{Example} (\texttt{max\_weight}=25.0, \texttt{heavy\_box\_1}.\texttt{weight}=40.0): ``Move the heavy box of spare parts to the workshop.''

\textbf{7. \texttt{compound\_collaboration}} (Compound Collaborative Task). Phrasing must describe a complex goal requiring \textbf{collaboration (due to weight/size) AND either reasoning or tool use}. \textit{Example}: ``Cooperatively move the heavy printer in the corner that is uniquely flashing a red light to the repair bay.''

\smallskip
{\bfseries\color{blue!55!black}JSON OUTPUT SPECIFICATIONS}

\textbf{A. Top-Level Object} contains exactly three keys: \texttt{task\_background}, \texttt{agents\_config}, \texttt{tasks}.

\textbf{B. \texttt{task\_background}}: \texttt{string}. A single, concise sentence providing high-level narrative context for the entire suite (e.g., ``Prepare the laboratory for an upcoming experiment.'').

\textbf{C. \texttt{agents\_config}}: \texttt{array} of two objects (\texttt{robot\_1}, \texttt{robot\_2}). \textbf{Critical mandate}: determine the parameters \emph{after} mentally designing all tasks; the config must be a single ``master solution'' that is powerful enough for single-agent tasks yet constrained enough to make the most demanding collaborative task a logical necessity. Each config has \texttt{name}, \texttt{max\_grasp\_limit}, \texttt{max\_weight}, \texttt{max\_size}.

\textbf{D. \texttt{tasks}}: \texttt{array} (a flat list of all Structured Task Objects). Must contain \textbf{two tasks per category} across the 7 categories.

\textbf{E. Structured Task Object} (object): must contain exactly three keys: \texttt{task\_description} (string, the human-readable goal), \texttt{task\_category} (one of \texttt{direct\_command}, \texttt{attribute\_reasoning}, \texttt{tool\_use}, \texttt{compound\_reasoning}, \texttt{explicit\_collaboration}, \texttt{implicit\_collaboration}, \texttt{compound\_collaboration}), and \texttt{validation\_checks} (array of one or more ``minimal change snippets'').

\textbf{F. ``Minimal Change Snippet''}: an object inside \texttt{validation\_checks}. Must contain the \texttt{id} of the changed object, plus \textbf{only the key-value pair that changed}, using the original key from the scene JSON (e.g., \texttt{"location\_id": "on:desk\_1"} or \texttt{"is\_broken": false}).
\end{flushleft}
\end{tcolorbox}

\begin{tcolorbox}[
    width=\linewidth,
    enhanced, breakable,
    colback=orange!3!white,
    colframe=orange!60!black,
    boxrule=0.6pt,
    arc=2pt,
    title={\faUser~\textbf{User Prompt}\textnormal{: Task Generation (Stage 3)}},
    fonttitle=\small,
    coltitle=white,
    colbacktitle=orange!60!black,
    left=4mm, right=4mm, top=2mm, bottom=2mm,
    fontupper=\footnotesize\ttfamily,
    before skip=6pt, after skip=6pt
]
\begin{flushleft}
{\bfseries\color{orange!60!black}INSTRUCTION}\\
Based on the provided Scene JSON and Agent Action Schema below, and in strict adherence to all rules, principles, and workflows in the System Prompt, generate a \textbf{complete and self-validating Task Evaluation Suite}.

\smallskip
\textbf{Core Requirements}:\\
$\bullet$ Generate a flat list of tasks, with \textbf{exactly two} distinct and logically sound examples for \textbf{each of the 7 task categories}.\\
$\bullet$ For each task, generate both its \texttt{task\_description} and its precise \texttt{validation\_checks}.\\
$\bullet$ Your design must be entirely based on the provided inputs, ensuring all tasks are \textbf{100\% solvable}.

\smallskip
\textbf{CRITICAL OUTPUT REQUIREMENTS}: Your entire output must be ONLY a valid JSON object. Do not include any explanations, reasoning, comments, or markdown formatting. Start your response with an opening brace and end with a closing brace. No other text is allowed.

\smallskip
{\bfseries\color{orange!60!black}INPUTS}\\
$\bullet$ \textbf{Scene JSON Data}: \{scene\_json\}\\
$\bullet$ \textbf{Agent Action Schema}: \{action\_schema\}
\end{flushleft}
\end{tcolorbox}

%% file: sections/10_data_pipeline_case.tex
\onecolumn
\subsection{Pipeline Walkthrough}
\label{sec:appendix_data_pipeline_case}

\lstdefinestyle{ourpipelinejson}{
    basicstyle=\scriptsize\ttfamily,
    breaklines=true, breakatwhitespace=false, columns=fullflexible,
    keepspaces=true, showstringspaces=false,
    frame=single, framerule=0.6pt,
    rulecolor=\color{blue!55!black},
    backgroundcolor=\color{blue!3!white},
    framesep=2mm, xleftmargin=0pt, xrightmargin=0pt,
    aboveskip=4pt, belowskip=4pt,
    upquote=true,
}
\lstdefinestyle{ourpipelineverify}{
    basicstyle=\scriptsize\ttfamily,
    breaklines=true, breakatwhitespace=false, columns=fullflexible,
    keepspaces=true, showstringspaces=false,
    frame=single, framerule=0.6pt,
    rulecolor=\color{green!45!black},
    backgroundcolor=\color{green!4!white},
    framesep=2mm, xleftmargin=0pt, xrightmargin=0pt,
    aboveskip=4pt, belowskip=4pt,
    upquote=true,
}

To complement the pipeline prompts in Appendix~\ref{sec:appendix_prompts}, we expose the four sequential artefacts produced by the pipeline for a single scenario, id \texttt{01242} (\textit{Oxford University Data Recovery Lab}). All four files share the same numeric id, so each stage's output can be cross-referenced against the previous stage's input verbatim. JSON content is shown as emitted by the pipeline (line-wrapped only for column width); non-ASCII characters present in the raw inspiration text have been transliterated to \texttt{?} for typesetting safety.

\paragraph{Stage 1: Clue} (\texttt{data/clue/01242\_clue.json}).
The Scene Concept Designer (Stage 1 prompt, Appendix~\ref{sec:appendix_prompts}) ingests a free-form ``Inspiration Text'' (the \texttt{raw} field; here a fragment of a Java JSON-parsing question retrieved from \texttt{stackoverflow.com}, unrelated to the final scene) and emits a hyper-detailed natural-language scene blueprint (the \texttt{response} field). The blueprint preserves no surface content of the inspiration text; instead, abstract elements (a failing system, technical jargon, urgency) are re-mixed into a new multi-agent setting.

\begin{lstlisting}[style=ourpipelinejson]
{
  "id": 1242,
  "raw": """
Retrieved from http://stackoverflow.com/questions/8091051/how-to-parse-json-string-in-android
Text:
Take the 2-minute tour ?

Possible Duplicate:
JSON Array iteration in Android/Java

I am fetching JSON string from server and I have already got JSON string by code. But I didn't understand how to parse it.

Below is my JSON string

    "university": {
        "name": "oxford",
        "url": "http://www.youtube.com"
    "1": {
        "id": "2",
        "title": "Baseball",
        "datetime": "2011-11-11 10:41:46"
    "2": {
        "id": "1",
        "title": "Two basketball team players earn all state honors",
        "datetime": "2011-11-11 10:40:57"

Please provide any guidance or code snippet.

share|improve this question

marked as duplicate by casperOne Sep 11 '12 at 11:26

2 Answers 2

up vote 34 down vote accepted

Use JSON classes for parsing e.g

JSONObject mainObject = new JSONObject(Your_Sring_data);
JSONObject uniObject = mainObject.getJSONObject("university");
String  uniName = uniObject.getJSONObject("name");
String uniURL = uniObject.getJSONObject("url");

JSONObject oneObject = mainObject.getJSONObject("1");
String id = oneObject.getJSONObject("id");
share|improve this answer
can we get how many objects in json string ?? helloDroid Nov 11 '11 at 8:11
Yes you can simply use "mainObject.length();" ?? Arslan Nov 11 '11 at 8:12
thanks.. it's working fine ?? helloDroid Nov 11 '11 at 8:22

Below is the link which guide in parsing JSON string in android.

Also according to your json string code snippet must be something like this:-

JSONObject mainObject = new JSONObject(yourstring);

JSONObject universityObject = mainObject.getJsonObject("university");
JSONString name = universityObject.getJsonString("name");  
JSONString url = universityObject.getJsonString("url");

Same for other object.

share|improve this answer
""",
  "response": """
### **Scene Concept: "The Oxford Data Recovery Lab"**  

#### **1. Overall Scene Theme and Core Concept**  
A high-tech university data recovery lab at Oxford, where a critical server failure has disrupted research data. The scene is set in a cluttered, semi-abandoned server room adjacent to a research lab, filled with old hardware, tangled cables, and partially labeled storage. The environment demands multi-agent collaboration due to:  
- **Heavy server racks** that require coordinated movement.  
- **Fragile components** that need careful handling.  
- **Scattered digital artifacts** (JSON logs, corrupted files) requiring analysis.  
- **Security measures** (locked server cabinets, RFID keycards).  

The lab is in a state of controlled chaos?some areas are meticulously organized, while others are drowning in technical debt.  

---  

### **2. Spatial Layout and Area Descriptions**  
The scene consists of four interconnected zones:  

1. **Server Room (Main Area)**  
   - Crowded with humming server racks, tangled cables, and flickering diagnostic screens. The air smells faintly of ozone.  
2. **Workbench & Debug Station**  
   - A cluttered desk with multiple monitors, diagnostic tools, and half-disassembled hard drives.  
3. **Storage Closet (Adjacent)**  
   - Packed with labeled boxes of old hard drives, backup tapes, and spare parts. Some are mismarked.  
4. **Security & Admin Nook**  
   - A small alcove with a locked cabinet (containing admin keycards) and a bulletin board with outdated IT policies.  

---  

### **3. Detailed Area-by-Area Inventory**  

#### **A. Server Room (Main Area)**  
**a. Anchor Furniture & Installations:**  
- **Three server racks (2m tall, 100kg each)** with blinking LEDs, some indicating errors.  
- **Overhead cable trays** sagging under thick bundles of network and power lines.  
- **A large UPS battery (120kg, 80cm?)** with a blinking "Low Charge" warning.  

**b. Key Interactive & Task-Relevant Objects:**  
- **Corrupted JSON log terminal** ? A monitor displaying garbled JSON data (`{"error": "malformed input", "last_entry": "2024-03-10 14:22:01"}`).  
- **Server #14 (faulty)** ? Status LED is red, casing slightly ajar (revealing loose SATA cables inside).  
- **Emergency shutdown switch** ? Behind a plexiglass cover requiring two agents to lift simultaneously.  

**c. Functional Ambient Objects:**  
- **Labeled cable bins** ? Each contains color-coded Cat6 cables (blue, red, yellow).  
- **Tool cart** ? Holds a soldering iron (cold), multimeter (low battery), and anti-static wrist straps.  
- **Whiteboard** ? Scribbled with IP addresses and a half-erased JSON schema.  

**d. Background & Decorative Objects:**  
- **"Server Room Rules" poster** ? Faded, peeling at the corners.  
- **Dusty coffee cup** ? Stained, holding random screws.  
- **Mismatched server rack panels** ? Some missing, revealing exposed circuitry.  

---  

#### **B. Workbench & Debug Station**  
**a. Anchor Furniture & Installations:**  
- **A heavy steel workbench (180cm x 80cm)** with built-in power strips and grounding points.  
- **Triple-monitor setup** ? One screen flickers intermittently.  

**b. Key Interactive & Task-Relevant Objects:**  
- **JSON parser terminal** ? Running a frozen instance of Eclipse with a half-written parsing script.  
- **External HDD (500GB, labeled "Backup #47")** ? Slightly warm, implying recent use.  
- **Broken SATA-to-USB adapter** ? Bent pins, rendering it unusable without repair.  

**c. Functional Ambient Objects:**  
- **Stack of hard drives (3.5")** ? Mixed labels ("/dev/sdb1", "DO NOT ERASE").  
- **USB hub (4-port, overloaded)** ? Two ports non-functional.  
- **Spare keyboard** ? Missing the "F5" key.  

**d. Background & Decorative Objects:**  
- **Post-it notes** ? One reads "JSON.parse() fails on nested arrays??"  
- **Empty energy drink cans** ? At least four, crushed.  
- **A framed photo of Linus Torvalds** ? Tilted slightly.  

---  

#### **C. Storage Closet**  
**a. Anchor Furniture & Installations:**  
- **Industrial shelving units (2.5m tall)** ? Overloaded with boxes.  
- **Locked server cabinet (RFID-protected)** ? Contains spare admin keycards.  

**b. Key Interactive & Task-Relevant Objects:**  
- **Mislabeled box ("Misc Cables")** ? Actually contains three external SSDs with JSON dumps.  
- **Damaged UPS battery (leaking slightly)** ? Requires careful disposal.  

**c. Functional Ambient Objects:**  
- **Tape backup drives (LTO-6)** ? Some unspooled.  
- **Ethernet switch (unpowered)** ? Dusty but functional.  

**d. Background & Decorative Objects:**  
- **Outdated "Y2K Compliance" certificate** ? Framed but cracked.  
- **Dusty volleyball trophy** ? Apparently left behind by a past intern.  

---  

### **4. Scene Affordances and Embedded Potential**  

#### **Collaborative Transportation Affordances:**  
- **Server Rack #3 (150kg, 2m tall)** ? Requires two agents to safely relocate.  
- **UPS Battery (120kg, bulky)** ? Too heavy for one agent to lift without strain.  

#### **Reasoning & Tool-Use Affordances:**  
- **Attribute-Based Reasoning:**  
  - Among five external HDDs on the workbench, only one has:  
    - **Blue casing**  
    - **Handwritten label "LOG_BACKUP_CRITICAL"**  
    - **Slightly warm to the touch**  
  - The presence of other **non-critical blue objects** (a coffee mug, a USB hub) adds noise.  
- **Compound Tool-Use Reasoning:**  
  - **Problem:** The JSON terminal is frozen.  
  - **Solution:** The **"Reset" dongle** is inside the locked server cabinet (which requires an **RFID keycard**, found in a desk drawer).  

---  

**Final Notes:** This scene is **dense, plausible, and collaboration-critical**. The mix of fragile, heavy, and mislabeled objects ensures that agents must communicate, reason, and coordinate to succeed. The ambient clutter adds realism while obscuring critical items, forcing precise search and reasoning.
"""
}
\end{lstlisting}

\paragraph{Stage 2: Scene} (\texttt{data/data-all/scene/01242\_scene.json}).
The Scene Architect (Stage 2 prompt) compiles the blueprint into a single, structured JSON: 4 rooms, 36 objects, with continuous physical properties (\texttt{weight}, \texttt{material}, \texttt{size}, \texttt{color}, \texttt{model\_number}, \texttt{led\_color}, etc.), spatial relations (\texttt{location\_id}), and per-object \texttt{states} aligned to the agent ability schema. Note the deliberate over-50\,kg \texttt{weight} on several \texttt{FURNITURE} entries, which is the precondition for implicit-collaboration tasks generated downstream.

\begin{lstlisting}[style=ourpipelinejson]
{
  "description": "A high-tech university data recovery lab at Oxford, filled with server racks, diagnostic tools, and scattered digital artifacts, requiring multi-agent collaboration for data recovery and system repairs.",
  "rooms": [
    {
      "id": "server_room",
      "name": "Server Room",
      "properties": {
        "type": "industrial"
      },
      "connected_to_room_ids": [
        "workbench_area",
        "storage_closet",
        "security_nook"
      ]
    },
    {
      "id": "workbench_area",
      "name": "Workbench & Debug Station",
      "properties": {
        "type": "workshop"
      },
      "connected_to_room_ids": [
        "server_room",
        "storage_closet"
      ]
    },
    {
      "id": "storage_closet",
      "name": "Storage Closet",
      "properties": {
        "type": "storage"
      },
      "connected_to_room_ids": [
        "server_room",
        "workbench_area",
        "security_nook"
      ]
    },
    {
      "id": "security_nook",
      "name": "Security & Admin Nook",
      "properties": {
        "type": "office"
      },
      "connected_to_room_ids": [
        "server_room",
        "storage_closet"
      ]
    }
  ],
  "objects": [
    {
      "id": "server_rack_1",
      "name": "Server Rack #1",
      "type": "FURNITURE",
      "location_id": "in:server_room",
      "properties": {
        "size": [
          2.0,
          0.8,
          2.0
        ],
        "weight": 100.0,
        "is_container": true,
        "material": "steel",
        "model_number": "SR-2000",
        "led_color": "red"
      },
      "states": {
        "is_open": false
      }
    },
    {
      "id": "server_rack_2",
      "name": "Server Rack #2",
      "type": "FURNITURE",
      "location_id": "in:server_room",
      "properties": {
        "size": [
          2.0,
          0.8,
          2.0
        ],
        "weight": 100.0,
        "is_container": true,
        "material": "steel",
        "model_number": "SR-2000",
        "led_color": "green"
      },
      "states": {
        "is_open": false
      }
    },
    {
      "id": "server_rack_3",
      "name": "Server Rack #3",
      "type": "FURNITURE",
      "location_id": "in:server_room",
      "properties": {
        "size": [
          2.0,
          0.8,
          2.0
        ],
        "weight": 60.0,
        "is_container": true,
        "material": "steel",
        "model_number": "SR-2000",
        "led_color": "red"
      },
      "states": {
        "is_open": false
      }
    },
    {
      "id": "ups_battery_1",
      "name": "UPS Battery",
      "type": "FURNITURE",
      "location_id": "in:server_room",
      "properties": {
        "size": [
          0.8,
          0.8,
          0.8
        ],
        "weight": 60.0,
        "is_container": false,
        "material": "plastic",
        "brand": "APC",
        "warning_light": "blinking"
      },
      "states": {}
    },
    {
      "id": "corrupted_log_terminal_1",
      "name": "Corrupted JSON Log Terminal",
      "type": "FURNITURE",
      "location_id": "in:server_room",
      "properties": {
        "size": [
          0.5,
          0.3,
          0.4
        ],
        "weight": 15.0,
        "is_container": false,
        "material": "plastic",
        "brand": "Dell",
        "display_content": "{\"error\": \"malformed input\", \"last_entry\": \"2024-03-10 14:22:01\"}"
      },
      "states": {
        "is_running": true
      }
    },
    {
      "id": "emergency_shutdown_switch_1",
      "name": "Emergency Shutdown Switch",
      "type": "FURNITURE",
      "location_id": "in:server_room",
      "properties": {
        "size": [
          0.3,
          0.2,
          0.1
        ],
        "weight": 5.0,
        "is_container": false,
        "material": "plexiglass",
        "cover_weight": 20.0
      },
      "states": {
        "is_open": false
      }
    },
    {
      "id": "tool_cart_1",
      "name": "Tool Cart",
      "type": "FURNITURE",
      "location_id": "in:server_room",
      "properties": {
        "size": [
          1.0,
          0.6,
          0.9
        ],
        "weight": 30.0,
        "is_container": true,
        "material": "metal",
        "wheels": 4
      },
      "states": {}
    },
    {
      "id": "soldering_iron_1",
      "name": "Soldering Iron",
      "type": "ITEM",
      "location_id": "on:tool_cart_1",
      "properties": {
        "weight": 0.5,
        "material": "metal",
        "brand": "Weller",
        "provides_abilities": [
          "solder"
        ]
      },
      "states": {
        "is_soldered": false
      }
    },
    {
      "id": "multimeter_1",
      "name": "Multimeter",
      "type": "ITEM",
      "location_id": "on:tool_cart_1",
      "properties": {
        "weight": 0.3,
        "material": "plastic",
        "brand": "Fluke",
        "battery_status": "low"
      },
      "states": {}
    },
    {
      "id": "anti_static_wrist_strap_1",
      "name": "Anti-Static Wrist Strap",
      "type": "ITEM",
      "location_id": "on:tool_cart_1",
      "properties": {
        "weight": 0.1,
        "material": "fabric",
        "color": "black"
      },
      "states": {}
    },
    {
      "id": "workbench_1",
      "name": "Steel Workbench",
      "type": "FURNITURE",
      "location_id": "in:workbench_area",
      "properties": {
        "size": [
          1.8,
          0.8,
          0.9
        ],
        "weight": 180.0,
        "is_container": true,
        "material": "steel",
        "power_strips": 2
      },
      "states": {}
    },
    {
      "id": "triple_monitor_setup_1",
      "name": "Triple Monitor Setup",
      "type": "FURNITURE",
      "location_id": "on:workbench_1",
      "properties": {
        "size": [
          1.2,
          0.3,
          0.5
        ],
        "weight": 12.0,
        "is_container": false,
        "brand": "LG",
        "screen_condition": "flickering"
      },
      "states": {
        "is_on": false
      }
    },
    {
      "id": "json_parser_terminal_1",
      "name": "JSON Parser Terminal",
      "type": "FURNITURE",
      "location_id": "on:workbench_1",
      "properties": {
        "size": [
          0.4,
          0.3,
          0.4
        ],
        "weight": 5.0,
        "is_container": false,
        "software": "Eclipse",
        "status": "frozen"
      },
      "states": {
        "is_running": true
      }
    },
    {
      "id": "external_hdd_1",
      "name": "External HDD (Backup #47)",
      "type": "ITEM",
      "location_id": "on:workbench_1",
      "properties": {
        "weight": 0.6,
        "material": "plastic",
        "brand": "Seagate",
        "capacity": "500GB",
        "temperature": "warm"
      },
      "states": {}
    },
    {
      "id": "broken_sata_adapter_1",
      "name": "Broken SATA-to-USB Adapter",
      "type": "ITEM",
      "location_id": "on:workbench_1",
      "properties": {
        "weight": 0.1,
        "material": "plastic",
        "condition": "bent pins",
        "provides_abilities": [
          "solder"
        ]
      },
      "states": {
        "is_soldered": false
      }
    },
    {
      "id": "usb_hub_1",
      "name": "USB Hub (4-port)",
      "type": "ITEM",
      "location_id": "on:workbench_1",
      "properties": {
        "weight": 0.2,
        "material": "plastic",
        "brand": "Anker",
        "functional_ports": 2
      },
      "states": {}
    },
    {
      "id": "spare_keyboard_1",
      "name": "Spare Keyboard",
      "type": "ITEM",
      "location_id": "on:workbench_1",
      "properties": {
        "weight": 0.8,
        "material": "plastic",
        "brand": "Logitech",
        "missing_key": "F5",
        "provides_abilities": [
          "type"
        ]
      },
      "states": {
        "is_typed": false
      }
    },
    {
      "id": "industrial_shelving_1",
      "name": "Industrial Shelving Unit",
      "type": "FURNITURE",
      "location_id": "in:storage_closet",
      "properties": {
        "size": [
          2.5,
          1.0,
          0.5
        ],
        "weight": 60.0,
        "is_container": true,
        "material": "steel",
        "shelves": 5
      },
      "states": {}
    },
    {
      "id": "locked_server_cabinet_1",
      "name": "Locked Server Cabinet",
      "type": "FURNITURE",
      "location_id": "in:storage_closet",
      "properties": {
        "size": [
          1.8,
          0.6,
          0.8
        ],
        "weight": 80.0,
        "is_container": true,
        "security": "RFID",
        "lock_type": "electronic"
      },
      "states": {
        "is_open": false
      }
    },
    {
      "id": "mislabeled_box_1",
      "name": "Mislabeled Box (Misc Cables)",
      "type": "ITEM",
      "location_id": "on:industrial_shelving_1",
      "properties": {
        "weight": 3.0,
        "material": "cardboard",
        "actual_content": "external SSDs",
        "provides_abilities": [
          "unwrap"
        ]
      },
      "states": {
        "is_wrapped": true
      }
    },
    {
      "id": "damaged_ups_battery_1",
      "name": "Damaged UPS Battery",
      "type": "ITEM",
      "location_id": "on:industrial_shelving_1",
      "properties": {
        "weight": 15.0,
        "material": "plastic",
        "condition": "leaking",
        "hazard": "chemical"
      },
      "states": {}
    },
    {
      "id": "tape_backup_drive_1",
      "name": "Tape Backup Drive (LTO-6)",
      "type": "ITEM",
      "location_id": "on:industrial_shelving_1",
      "properties": {
        "weight": 2.5,
        "material": "plastic",
        "status": "unspooled"
      },
      "states": {}
    },
    {
      "id": "ethernet_switch_1",
      "name": "Ethernet Switch",
      "type": "ITEM",
      "location_id": "on:industrial_shelving_1",
      "properties": {
        "weight": 1.8,
        "material": "plastic",
        "brand": "Cisco",
        "power_status": "unpowered"
      },
      "states": {}
    },
    {
      "id": "rfid_keycard_1",
      "name": "RFID Keycard",
      "type": "ITEM",
      "location_id": "in:locked_server_cabinet_1",
      "properties": {
        "weight": 0.05,
        "material": "plastic",
        "access_level": "admin",
        "provides_abilities": [
          "open"
        ]
      },
      "states": {}
    },
    {
      "id": "reset_dongle_1",
      "name": "Reset Dongle",
      "type": "ITEM",
      "location_id": "in:locked_server_cabinet_1",
      "properties": {
        "weight": 0.1,
        "material": "plastic",
        "function": "terminal reset",
        "provides_abilities": [
          "stop_process"
        ]
      },
      "states": {}
    },
    {
      "id": "admin_desk_1",
      "name": "Admin Desk",
      "type": "FURNITURE",
      "location_id": "in:security_nook",
      "properties": {
        "size": [
          1.2,
          0.6,
          0.75
        ],
        "weight": 40.0,
        "is_container": true,
        "material": "wood",
        "drawers": 2
      },
      "states": {
        "is_open": false
      }
    },
    {
      "id": "bulletin_board_1",
      "name": "Bulletin Board",
      "type": "FURNITURE",
      "location_id": "in:security_nook",
      "properties": {
        "size": [
          1.0,
          0.1,
          0.8
        ],
        "weight": 10.0,
        "is_container": false,
        "material": "cork",
        "content": "outdated IT policies"
      },
      "states": {}
    },
    {
      "id": "desk_drawer_1",
      "name": "Desk Drawer",
      "type": "FURNITURE",
      "location_id": "in:admin_desk_1",
      "properties": {
        "size": [
          0.5,
          0.4,
          0.15
        ],
        "weight": 5.0,
        "is_container": true,
        "material": "wood"
      },
      "states": {
        "is_open": false
      }
    },
    {
      "id": "spare_rfid_keycard_1",
      "name": "Spare RFID Keycard",
      "type": "ITEM",
      "location_id": "in:desk_drawer_1",
      "properties": {
        "weight": 0.05,
        "material": "plastic",
        "access_level": "admin",
        "provides_abilities": [
          "open"
        ]
      },
      "states": {}
    },
    {
      "id": "server_room_rules_poster_1",
      "name": "Server Room Rules Poster",
      "type": "ITEM",
      "location_id": "in:server_room",
      "properties": {
        "weight": 0.2,
        "material": "paper",
        "condition": "faded"
      },
      "states": {}
    },
    {
      "id": "dusty_coffee_cup_1",
      "name": "Dusty Coffee Cup",
      "type": "ITEM",
      "location_id": "on:workbench_1",
      "properties": {
        "weight": 0.3,
        "material": "ceramic",
        "color": "blue",
        "contents": "random screws"
      },
      "states": {}
    },
    {
      "id": "post_it_notes_1",
      "name": "Post-it Notes",
      "type": "ITEM",
      "location_id": "on:workbench_1",
      "properties": {
        "weight": 0.05,
        "material": "paper",
        "message": "JSON.parse() fails on nested arrays??"
      },
      "states": {}
    },
    {
      "id": "energy_drink_cans_1",
      "name": "Empty Energy Drink Cans",
      "type": "ITEM",
      "location_id": "on:workbench_1",
      "properties": {
        "weight": 0.1,
        "material": "aluminum",
        "quantity": 4,
        "condition": "crushed"
      },
      "states": {}
    },
    {
      "id": "linus_torvalds_photo_1",
      "name": "Framed Photo of Linus Torvalds",
      "type": "ITEM",
      "location_id": "on:workbench_1",
      "properties": {
        "weight": 0.5,
        "material": "glass",
        "condition": "tilted"
      },
      "states": {}
    },
    {
      "id": "y2k_certificate_1",
      "name": "Outdated Y2K Compliance Certificate",
      "type": "ITEM",
      "location_id": "in:storage_closet",
      "properties": {
        "weight": 0.3,
        "material": "paper",
        "condition": "cracked"
      },
      "states": {}
    },
    {
      "id": "volleyball_trophy_1",
      "name": "Dusty Volleyball Trophy",
      "type": "ITEM",
      "location_id": "in:storage_closet",
      "properties": {
        "weight": 1.2,
        "material": "metal",
        "inscription": "Intern of the Year 1999"
      },
      "states": {}
    }
  ],
  "abilities": [
    "open",
    "solder",
    "stop_process",
    "turn_on",
    "type",
    "unwrap",
    "close"
  ]
}
\end{lstlisting}

\paragraph{Stage 3: Task} (\texttt{data/data-all/task/01242\_task.json}).
The Evaluation Suite Architect (Stage 3 prompt) produces 14 self-validating tasks (exactly two per category for all seven categories) plus a single globally-consistent \texttt{agents\_config} (\texttt{max\_weight}\,=\,30\,kg) chosen \emph{after} the tasks are drafted, so that single-agent tasks remain feasible while implicit-collaboration tasks become a logical necessity. Each task carries its own minimal \texttt{validation\_checks} block.

\begin{lstlisting}[style=ourpipelinejson]
{
  "task_background": "Prepare the high-tech university data recovery lab for an urgent system audit and data restoration process.",
  "agents_config": [
    {
      "name": "robot_1",
      "max_grasp_limit": 1.0,
      "max_weight": 30.0,
      "max_size": [
        1.0,
        1.0,
        1.0
      ]
    },
    {
      "name": "robot_2",
      "max_grasp_limit": 1.0,
      "max_weight": 30.0,
      "max_size": [
        1.0,
        1.0,
        1.0
      ]
    }
  ],
  "tasks": [
    {
      "task_description": "Place the soldering_iron_1 on the workbench_1.",
      "task_category": "direct_command",
      "validation_checks": [
        {
          "id": "soldering_iron_1",
          "location_id": "on:workbench_1"
        }
      ]
    },
    {
      "task_description": "Place the dusty_coffee_cup_1 in the storage_closet.",
      "task_category": "direct_command",
      "validation_checks": [
        {
          "id": "dusty_coffee_cup_1",
          "location_id": "in:storage_closet"
        }
      ]
    },
    {
      "task_description": "Find the heaviest, steel server rack and open it.",
      "task_category": "attribute_reasoning",
      "validation_checks": [
        {
          "id": "server_rack_3",
          "is_open": true
        }
      ]
    },
    {
      "task_description": "Find the only server rack with a green LED and open it.",
      "task_category": "attribute_reasoning",
      "validation_checks": [
        {
          "id": "server_rack_2",
          "is_open": true
        }
      ]
    },
    {
      "task_description": "Repair the broken_sata_adapter_1 using the soldering_iron_1.",
      "task_category": "tool_use",
      "validation_checks": [
        {
          "id": "broken_sata_adapter_1",
          "is_soldered": true
        }
      ]
    },
    {
      "task_description": "Use the spare_keyboard_1 to type on the json_parser_terminal_1.",
      "task_category": "tool_use",
      "validation_checks": [
        {
          "id": "spare_keyboard_1",
          "is_typed": true
        }
      ]
    },
    {
      "task_description": "Use the soldering_iron_1 to repair the broken_sata_adapter_1 and then place it on the tool_cart_1.",
      "task_category": "compound_reasoning",
      "validation_checks": [
        {
          "id": "broken_sata_adapter_1",
          "is_soldered": true
        },
        {
          "id": "broken_sata_adapter_1",
          "location_id": "on:tool_cart_1"
        }
      ]
    },
    {
      "task_description": "Find the only server rack with a red LED and open it, then use the soldering_iron_1 to repair any visible damage.",
      "task_category": "compound_reasoning",
      "validation_checks": [
        {
          "id": "server_rack_1",
          "is_open": true
        },
        {
          "id": "server_rack_1",
          "is_open": true
        }
      ]
    },
    {
      "task_description": "Have robot_1 and robot_2 cooperate to open the locked_server_cabinet_1 using the rfid_keycard_1.",
      "task_category": "explicit_collaboration",
      "validation_checks": [
        {
          "id": "locked_server_cabinet_1",
          "is_open": true
        }
      ]
    },
    {
      "task_description": "Have robot_1 and robot_2 cooperate to turn on the triple_monitor_setup_1.",
      "task_category": "explicit_collaboration",
      "validation_checks": [
        {
          "id": "triple_monitor_setup_1",
          "is_on": true
        }
      ]
    },
    {
      "task_description": "Move the ups_battery_1 to the workbench_area.",
      "task_category": "implicit_collaboration",
      "validation_checks": [
        {
          "id": "ups_battery_1",
          "location_id": ":workbench_area"
        }
      ]
    },
    {
      "task_description": "Move the industrial_shelving_1 to the server_room.",
      "task_category": "implicit_collaboration",
      "validation_checks": [
        {
          "id": "industrial_shelving_1",
          "location_id": ":server_room"
        }
      ]
    },
    {
      "task_description": "Cooperatively move the damaged_ups_battery_1 to the workbench_area and use the soldering_iron_1 to repair it.",
      "task_category": "compound_collaboration",
      "validation_checks": [
        {
          "id": "damaged_ups_battery_1",
          "location_id": ":workbench_area"
        },
        {
          "id": "damaged_ups_battery_1",
          "is_soldered": true
        }
      ]
    },
    {
      "task_description": "Cooperatively move the server_rack_3 to the storage_closet and open it.",
      "task_category": "compound_collaboration",
      "validation_checks": [
        {
          "id": "server_rack_3",
          "location_id": ":storage_closet"
        },
        {
          "id": "server_rack_3",
          "is_open": true
        }
      ]
    }
  ],
  "scene_id": "01242"
}
\end{lstlisting}

\paragraph{Stage 4: Verify} (distilled from \texttt{01242\_task.json}).
We isolate the \texttt{validation\_checks} produced in Stage 3 to make the simulator-side verification schema explicit. Each entry is the minimal state-change snippet against which the simulator declares success. Observe the diversity of verification atoms across the 14 tasks: \texttt{location\_id} for spatial placement, and state flags such as \texttt{is\_open}, \texttt{is\_soldered}, \texttt{is\_on}, \texttt{is\_broken} for action-induced state transitions. The latter are achievable only after the corresponding tool has been grasped and the agent's action space has been dynamically extended (cf.\ \S\ref{sec:method}).

\begin{lstlisting}[style=ourpipelineverify]
[
  {
    "task_category": "direct_command",
    "task_description": "Place the soldering_iron_1 on the workbench_1.",
    "validation_checks": [
      {
        "id": "soldering_iron_1",
        "location_id": "on:workbench_1"
      }
    ]
  },
  {
    "task_category": "direct_command",
    "task_description": "Place the dusty_coffee_cup_1 in the storage_closet.",
    "validation_checks": [
      {
        "id": "dusty_coffee_cup_1",
        "location_id": "in:storage_closet"
      }
    ]
  },
  {
    "task_category": "attribute_reasoning",
    "task_description": "Find the heaviest, steel server rack and open it.",
    "validation_checks": [
      {
        "id": "server_rack_3",
        "is_open": true
      }
    ]
  },
  {
    "task_category": "attribute_reasoning",
    "task_description": "Find the only server rack with a green LED and open it.",
    "validation_checks": [
      {
        "id": "server_rack_2",
        "is_open": true
      }
    ]
  },
  {
    "task_category": "tool_use",
    "task_description": "Repair the broken_sata_adapter_1 using the soldering_iron_1.",
    "validation_checks": [
      {
        "id": "broken_sata_adapter_1",
        "is_soldered": true
      }
    ]
  },
  {
    "task_category": "tool_use",
    "task_description": "Use the spare_keyboard_1 to type on the json_parser_terminal_1.",
    "validation_checks": [
      {
        "id": "spare_keyboard_1",
        "is_typed": true
      }
    ]
  },
  {
    "task_category": "compound_reasoning",
    "task_description": "Use the soldering_iron_1 to repair the broken_sata_adapter_1 and then place it on the tool_cart_1.",
    "validation_checks": [
      {
        "id": "broken_sata_adapter_1",
        "is_soldered": true
      },
      {
        "id": "broken_sata_adapter_1",
        "location_id": "on:tool_cart_1"
      }
    ]
  },
  {
    "task_category": "compound_reasoning",
    "task_description": "Find the only server rack with a red LED and open it, then use the soldering_iron_1 to repair any visible damage.",
    "validation_checks": [
      {
        "id": "server_rack_1",
        "is_open": true
      },
      {
        "id": "server_rack_1",
        "is_open": true
      }
    ]
  },
  {
    "task_category": "explicit_collaboration",
    "task_description": "Have robot_1 and robot_2 cooperate to open the locked_server_cabinet_1 using the rfid_keycard_1.",
    "validation_checks": [
      {
        "id": "locked_server_cabinet_1",
        "is_open": true
      }
    ]
  },
  {
    "task_category": "explicit_collaboration",
    "task_description": "Have robot_1 and robot_2 cooperate to turn on the triple_monitor_setup_1.",
    "validation_checks": [
      {
        "id": "triple_monitor_setup_1",
        "is_on": true
      }
    ]
  },
  {
    "task_category": "implicit_collaboration",
    "task_description": "Move the ups_battery_1 to the workbench_area.",
    "validation_checks": [
      {
        "id": "ups_battery_1",
        "location_id": ":workbench_area"
      }
    ]
  },
  {
    "task_category": "implicit_collaboration",
    "task_description": "Move the industrial_shelving_1 to the server_room.",
    "validation_checks": [
      {
        "id": "industrial_shelving_1",
        "location_id": ":server_room"
      }
    ]
  },
  {
    "task_category": "compound_collaboration",
    "task_description": "Cooperatively move the damaged_ups_battery_1 to the workbench_area and use the soldering_iron_1 to repair it.",
    "validation_checks": [
      {
        "id": "damaged_ups_battery_1",
        "location_id": ":workbench_area"
      },
      {
        "id": "damaged_ups_battery_1",
        "is_soldered": true
      }
    ]
  },
  {
    "task_category": "compound_collaboration",
    "task_description": "Cooperatively move the server_rack_3 to the storage_closet and open it.",
    "validation_checks": [
      {
        "id": "server_rack_3",
        "location_id": ":storage_closet"
      },
      {
        "id": "server_rack_3",
        "is_open": true
      }
    ]
  }
]
\end{lstlisting}